\documentclass{article} 
\usepackage{iclr2027_conference,times}

\usepackage{amsmath,amsfonts,bm}

\def\eqref#1{equation~\ref{#1}}

\def\1{\bm{1}}

\DeclareMathAlphabet{\mathsfit}{\encodingdefault}{\sfdefault}{m}{sl}
\SetMathAlphabet{\mathsfit}{bold}{\encodingdefault}{\sfdefault}{bx}{n}

\usepackage{hyperref}
\usepackage{url}
\usepackage{amsthm}
\usepackage{graphicx}
\usepackage{float}
\usepackage{placeins}
\usepackage{algorithm} 
\usepackage{algpseudocode}
\usepackage{subcaption}
\usepackage{xspace}
\usepackage[table]{xcolor}
\usepackage{pifont}
\usepackage{xcolor}
\usepackage{etoc}
\definecolor{passgreen}{RGB}{120,200,120}
\usepackage{booktabs}
\newcommand{\cmark}{\textcolor{green!55!black}{\ding{51}}}
\newcommand{\xmark}{\textcolor{red!75!black}{\ding{55}}}
\newtheorem{proposition}{Proposition}
\newtheorem{lemma}{Lemma}
\newtheorem{theorem}{Theorem}
\newtheorem*{theorem*}{Theorem}

\usepackage{fontawesome5}
\usepackage{hyperref}
\usepackage{pifont}
\usepackage{booktabs,fontawesome5,amssymb,amsmath,longtable}
\usepackage[table]{xcolor}   
\definecolor{passgreen}{RGB}{198,231,201}
\newcommand{\Rh}{\bar{R}_h}                 
\newcommand{\Rm}{\bar{R}_m}                 

\usepackage{booktabs}
\newcommand{\alttest}{\textsc{Alt-Test}\xspace}
\newcommand{\anlt}{\textsc{AltRA-Test}\xspace}
\usepackage{wrapfig}
\usepackage{fontawesome5}
\definecolor{bestblue}{RGB}{198,219,239}
\definecolor{bestgreen}{RGB}{146,205,139}

\title{AltRA-Test: An Alternative Annotator Test for LLM Judges Without Shared Annotators}

\newcommand{\hficon}{\raisebox{-0.2em}{\includegraphics[height=1em]{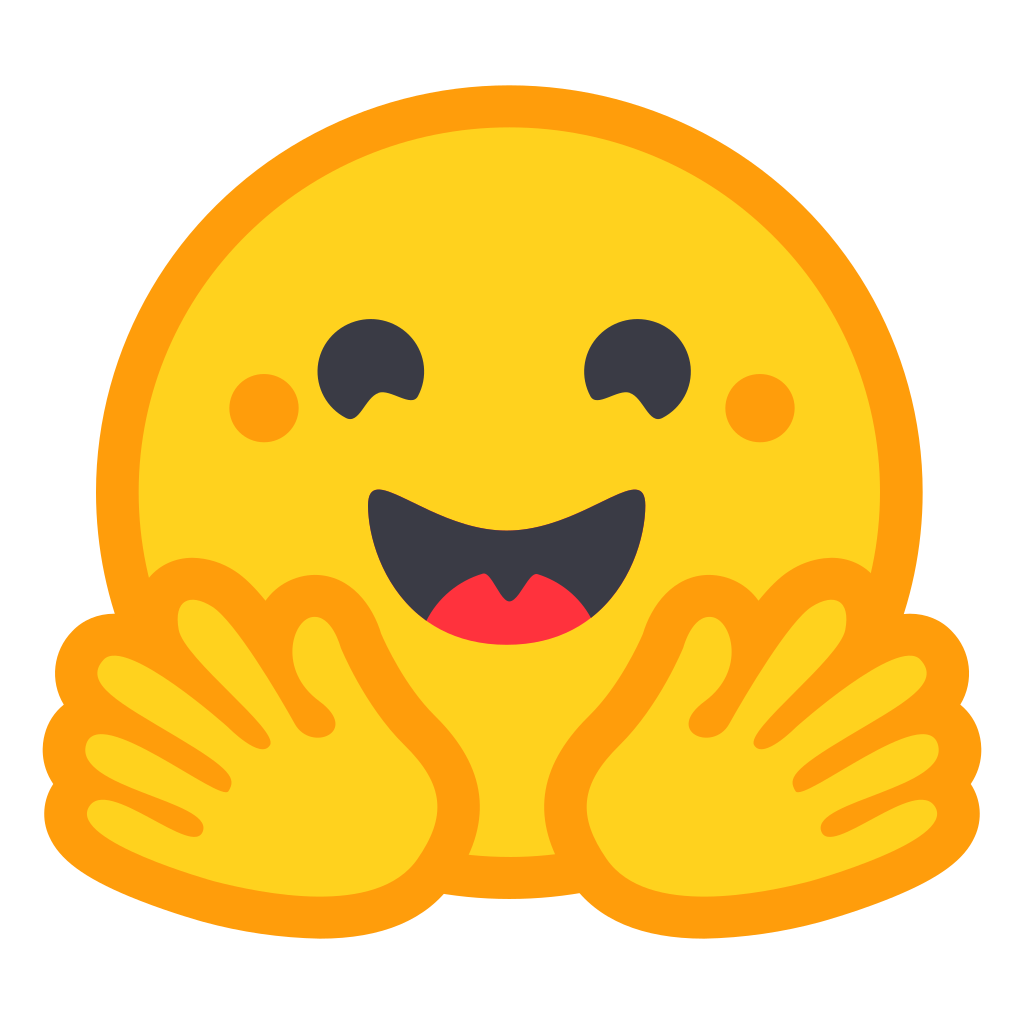}}}
\newcommand{\anonrepo}{\href{https://huggingface.co/datasets/naghamo/altra-test}{%
  \hficon\,\texttt{naghamo/altra-test}}}
\usepackage{pifont}
\definecolor{passgreen}{RGB}{198,239,206}   
\definecolor{xmarkgray}{gray}{0.55}  
\newcommand{\xmarkg}{\textcolor{xmarkgray}{\ding{55}}}

\author{Nagham Omar$^{1}$, Rotem Dror$^{2}$, Almog Gueta$^{3}$, Avigdor Gal$^{1}$, Roi Reichart$^{1}$, Nitay Calderon$^{3}$ \\
$^{1}$Faculty of Data and Decision Sciences, Technion \\
$^{2}$Faculty of Computer and Information Science, University of Haifa \\
$^{3}$Google Research \\
\texttt{nagham.omar@campus.technion.ac.il, rdror@is.haifa.ac.il,} \\
\texttt{avigal@technion.ac.il, roiri@technion.ac.il, nitayc@google.com}
}

\iclrfinalcopy 
\begin{document}

\maketitle
\lhead{} 
\etocsettocdepth.toc{none} 

\begin{abstract}
Large language models (LLMs) are increasingly used as judges and graders to replace human annotators, raising the question of when such replacement is statistically justified. To support such decisions, the Alternative Annotator Test (\alttest{}) was introduced for dense settings where all items are annotated by the same annotators. However, in many large-scale annotation settings, items are annotated by different subsets of annotators. We therefore propose the Alternative Annotator Test with Resampled Annotations (\anlt{}), a statistical test that does not require that all items are annotated by the same annotators.
\anlt{} uses a bootstrap that resamples items, draws $k$ human annotations for each item, and substitutes one of them with an LLM annotation. This produces paired all-human and mixed human–LLM panels. A replacement is supported when the inter-annotator agreement (IAA) of the mixed panel is statistically non-inferior to the IAA of the all-human panel, i.e., when the LLM preserves the signal that the human annotators share.
We establish that \anlt{} is a consistent test of asymptotic level $\alpha$. We then evaluate it through simulations and real-world experiments in which we test 18 LLMs and 5 prompting techniques on 20 datasets across text, image, audio, and video.
Across these evaluations, about 100 items often suffice for the test. 
In simulations, where we know whether the LLM is a better or a worse annotator than the humans, \anlt{} makes more correct replacement decisions than existing baselines.
In real-world data experiments, LLMs pass the test in about half of the setups, depending mainly on the task. Since replacement is not always justified, we encourage researchers to test it for their task\footnote{\faGlobe~\href{https://naghamo.github.io/altra-test.io/}{AltRA-Test paper page}: run the test on your own data online.}.
\end{abstract}

\section{Introduction}
\label{sec:intro}

Creating labeled datasets has long depended on human annotation, one of the slowest and most expensive stages of the AI pipeline~\citep{DBLP:conf/acl/DingQLCLJB23, tan2024large}. 
Researchers use human annotations to train models and to evaluate them, and they draw empirical conclusions from both~\citep{DBLP:conf/acl/MeiHCCXWBW26, omar2026train}, so annotation quality directly affects the validity of the resulting models and findings. 
Large language models (LLMs) offer an appealing alternative: they can annotate quickly, cheaply, and at a scale that is difficult to achieve with experts or even with crowd workers \citep{ DBLP:conf/acl/ChiangL23,gilardi2023chatgpt,tornberg2023chatgpt, nahum2025llms}. Indeed, LLMs are increasingly used as annotators, judges (LLM-as-a-judge), and raters~\citep{zheng2023judging, wei2024systematic,li2025generation,wu2025sailing,DBLP:conf/acl/MeiHCCXWBW26}.

In many AI applications, such as response quality evaluation, there is no ground truth to check labels against (unlike, e.g., checking whether code runs or a math answer is correct), and human judgment defines the correct label~\citep{DBLP:journals/jmlr/RaykarYZVFBM10, DBLP:journals/tacl/PaunCCHKP18}. Human annotations can serve two different goals. In most cases, they are used to train, evaluate, or judge models. This requires one target per item, so what matters is the signal that the annotators share, and not the variation between individual annotators. In other studies, the disagreement is itself the object of study, and the goal is to capture the different perspectives of the annotators~\citep{aroyo2015truth,basile2021we,uma2021learning,plank2022problem}. In this paper, we focus on the first goal, although we also propose adaptations of our test for the second. Replacing human annotators with LLMs is therefore useful only if the replacement preserves this shared signal~\citep{DBLP:conf/nips/EgamiHSW23,DBLP:conf/naacl/GligoricZLCJ25,DBLP:journals/natmi/WangMD25}. However, it is not guaranteed that LLMs preserve it. They have been shown to disagree with experts even when their responses appear fluent or confident~\citep{DBLP:conf/acl/ZhouH0S24, griot2025large, szymanski2025limitations, tseng2025evaluating, DBLP:conf/coling/WuA25,DBLP:journals/corr/abs-2609-37863}, and to systematically diverge from annotators in subjective tasks ~\citep{lissak2024colorful,DBLP:conf/iclr/Elangovan0KELBR25, DBLP:conf/icml/ZhangWHDD0C0P25}.
Accordingly, using an LLM as an alternative to humans requires statistical justification for the replacement~\citep{bavaresco2025llms,calderon2025alternative}.

\begin{figure}[t]
\centering
\includegraphics[width=\textwidth]{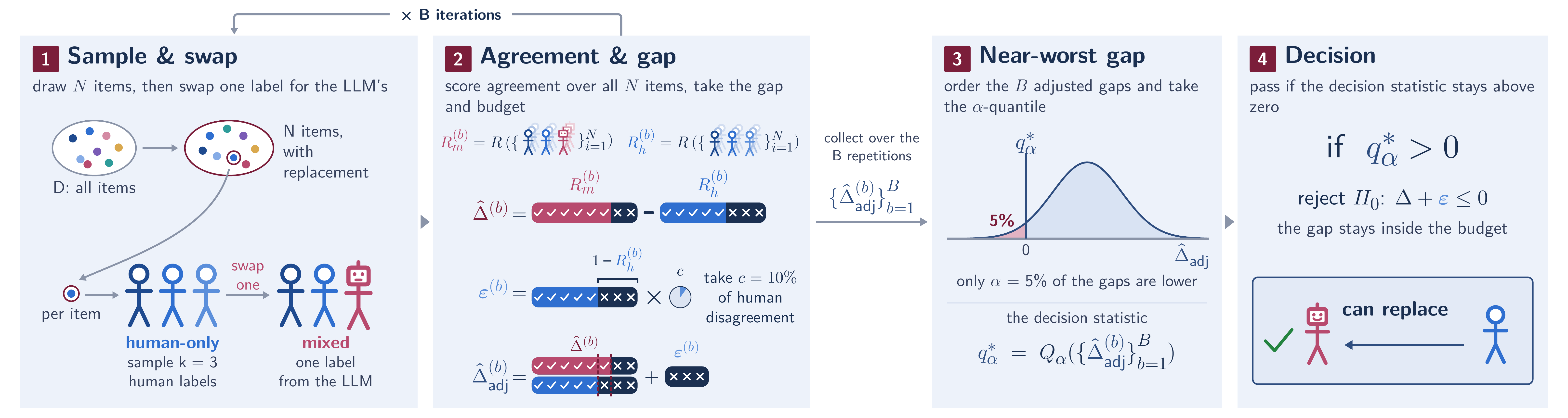}
\caption{An illustration of the \anlt{}. The two left boxes present the paired-bootstrap procedure: for each of $B$ iterations, we sample $N$ items and $k\!=\!3$ annotations per item, replace one annotation with the LLM annotation, and compute IAA for the human-only and human--LLM panels. The substitution gap $\Delta^{(b)}$ and adaptive margin $\varepsilon^{(b)}$ are then computed. The two right boxes present the decision stage, based on a non-inferiority test. Here, $q_\alpha^*$ is the $\alpha$-quantile statistic; if it is positive, we conclude that the LLM is a good alternative to a human annotator.}
\label{fig:overview}
\vspace{-0.8em}
\end{figure}

Since there is no ground truth to compare with, researchers assess human annotation quality using inter-annotator agreement (IAA)~\citep{hayes2007answering, DBLP:journals/coling/KlieCG24}: if independent annotators consistently agree above chance, the task is well-defined, and the labels capture a meaningful shared signal rather than random noise~\citep{DBLP:journals/coling/ArtsteinP08,krippendorff2011computing}. Even when IAA is modest, as in subjective tasks where some disagreement is inherent~\citep{basile2021we, DBLP:journals/tacl/DavaniDP22, plank2022problem, DBLP:conf/naacl/SapSVZCS22, DBLP:conf/emnlp/FleisigAK23}, agreement clearly above chance still indicates a shared, if noisy, signal. LLM annotations are typically evaluated using accuracy, $F_1$, or correlation with a majority-vote label~\citep{ziems2024can, bavaresco2025llms, DBLP:conf/iclr/Elangovan0KELBR25, DBLP:conf/nips/GuerdanBHWWC25}. However, since some disagreement is inherent, individual human annotators also disagree with the majority-vote label, so even a human would not score perfectly on these metrics. Their ceiling depends on the task, and without knowing it, it is unclear whether a given score is good or poor. As a result, such measures cannot provide a statistically grounded replacement decision. A replacement test should therefore mirror how we evaluate human annotation quality, and test whether the LLM keeps the IAA over all human annotations statistically non-inferior to that of the humans.

Two recent works formulate human replacement by LLMs as a statistical procedure, the Alternative Annotator Test (\alttest{};~\citet{calderon2025alternative}) and the Statistical Turing Test (STT;~\citet{he2025can}). Both assume a representative set of items, which we call the evaluation set, each labeled by multiple human annotators. Crucially, they also require the same group of annotators to label every item in this set.
However, this requirement is restrictive, and in many common annotation settings it does not hold. In crowdsourcing, workers sometimes choose their assignments and may stop after a few items. In preference annotation, a large pool of participants contributes only a few judgments each, and in demanding tasks, such as verifying the factuality of a long document, an annotator can complete only a few items. Finally, many existing datasets omit the identities of the annotators~\citep{siththaranjan2024distributional}. In all of these settings, researchers observe multiple human labels for each item, but the items are not all labeled by the same annotators, making existing tests inapplicable. Indeed, $11$ of our $20$ datasets do not satisfy their requirements.

In this work, we introduce the \emph{\textbf{Alt}ernative annotator test with \textbf{R}esampled \textbf{A}nnotations} (\anlt{}\footnote{\anonrepo{}}; Figure~\ref{fig:overview}), which does not require that the same annotators label all the items in the evaluation set. The test uses a paired percentile bootstrap: in each bootstrap sample, it resamples items with replacement, draws $k$ human annotations per item, and constructs a mixed panel by replacing one human annotation with the LLM annotation. Unlike existing tests, it resamples the annotations within each item, and therefore never uses the identities of the annotators. It then tests whether the mixed-panel IAA is non-inferior to the human-only IAA. An adaptive margin permits a small IAA loss proportional to the observed human disagreement, so the same test applies to objective tasks, in which annotators agree strongly, and to subjective tasks, in which some disagreement is inherent. E.g., our calibrated default of $c=0.1$ allows at most a $10\%$ increase in disagreement. We say that an LLM \emph{passes} the test when we reject the null hypothesis that the mixed-panel IAA is inferior.

We theoretically establish that \anlt{} is a consistent test of asymptotic level $\alpha$, i.e., as the number of items grows, it accepts valid replacements with probability approaching $1$ and falsely accepts invalid ones with probability at most $\alpha$.
We then use simulations, in which items have a latent gold label that annotators observe with noise, and we control whether the LLM is more or less noisy than the humans. This enables us to evaluate the replacement question directly, as we measure each procedure's better-model pass rate and worse-model pass rate (i.e., correctly accepting simulated LLMs that are less noisy than simulated humans versus falsely accepting noisier ones). In the setup that the existing tests require, in which the same annotators label all items, \anlt{} dominates \alttest{} and STT, with a higher better-model pass rate at any given worse-model pass rate. We also evaluate the \anlt{} in the setup that motivates it, in which the items are not all labeled by the same annotators, and there, its pass rates remain the same. The simulations also calibrate the default margin constant to $c=0.1$ and show reliable behavior with around $100$ items.

In addition, we evaluate the \anlt{} on $20$ real-world datasets across text, image, audio, and video, spanning objective and subjective tasks with a wide range of human IAA levels. We test $18$ LLM judges and $5$ prompting techniques. Under the default CoT setting, LLMs pass the test in $71$ of $201$ judge--dataset pairs ($35.3\%$); when each judge uses its best prompting technique, this increases to $95$ of $201$ ($47.3\%$). We find that replacement is strongly dataset-dependent, and in many passing cases the LLM not only passes the test but increases IAA. Consistent with the simulations, sample-size analyses show that around $100$ items often suffice for LLMs that pass with $300$ items.

We encourage researchers to apply the \anlt{} before delegating annotation to an LLM. As the test examines whether the LLM preserves the signal that the annotators share, its result is meaningful only when this signal exists, and we recommend applying it when human agreement is at least fair (IAA $\geq0.2$). We hope it supports more explicit and statistically justified decisions.
\section{Related Work}
\label{sec:related}

\paragraph{IAA as a proxy for human annotation quality.}
When ground-truth labels are unavailable, researchers commonly assess annotation quality through IAA \citep{DBLP:journals/coling/ArtsteinP08,krippendorff2011computing}. In this work, we use Krippendorff's $\alpha$, a standard chance-corrected IAA coefficient that supports multiple annotators, missing annotations, and nominal or ordinal labels \citep{hayes2007answering,krippendorff2011computing}. IAA scores should be interpreted relative to the task: objective or well-specified tasks are expected to yield high agreement, while difficult, ambiguous, or preference-based tasks may yield lower but still meaningful agreement \citep{DBLP:journals/tacl/PavlickK19,plank2022problem,lissak2024colorful}. This interpretation depends on the purpose of annotation. When disagreement is itself the object of study~\citep{aroyo2015truth,basile2021we,uma2021learning,plank2022problem}, forcing high IAA may remove the variation that the study aims to measure. Our focus is different (see Section~\ref{sec:intro}), and for such studies we propose variants of our test (see Section~\ref{subsec:variants}). Even in subjective tasks, such as response-preference annotation, where human IAA may have an inherent ceiling, positive and non-negligible IAA (e.g., 0.2 or higher) indicates that the annotations carry a shared signal that an LLM replacement should reflect. We test whether LLMs preserve it.

\paragraph{Testing whether LLMs can replace human annotators.}
Various methods propose correcting or calibrating model-generated labels~\citep{angelopoulos2023prediction, lee2025correctly, feng2026noisy}, but they do not decide whether replacement is justified. Two recent procedures address replacement as a statistical decision. \alttest{}~\citep{calderon2025alternative} holds out each annotator in turn and tests whether an LLM is non-inferior to that annotator in its agreement with the remaining annotators. STT~\citep{he2025can} tests whether human and mixed human--LLM panels are statistically equivalent. Both require the same annotators to label all the items, a requirement that often does not hold (see Section~\ref{sec:intro}). \anlt{} removes this requirement. Like \alttest{} and unlike STT, it answers a non-inferiority question. This fits the replacement decision, since LLM annotations are much cheaper than human ones, and researchers may prefer an LLM even at the cost of a small loss. An equivalence test does not accept an LLM that increases IAA, even if it captures the shared signal more reliably than a typical annotator. Like STT, it uses bootstrap resampling, but unlike both, it does not replace all annotations of a single annotator. Instead, for each sampled item, it draws $k$ human annotations, replaces one with the LLM annotation, and bases the decision on a percentile-bootstrap statistic. Its margin also differs. \alttest{} uses a fixed margin that is chosen according to the type of the annotators, and STT derives its margin from the agreement gap between two groups of humans, while the margin of \anlt{} adapts to the observed human disagreement. When all three tests can be applied, \anlt{} dominates both (see Section~\ref{sec:sim}).

\section{Method}
\label{sec:method}

\paragraph{Setup and notations.}
\label{subsec:setup}
Let $k\ge3$ be the panel size used by the test, and let $\mathcal D=\{x_1,\ldots,x_N\}$ be a dataset of $N$ items. For each item $x_i$, we observe $M_i$ human annotations $H_i=\{h_{i,1},\ldots,h_{i,M_i}\}$, with $M_i\ge k$. We do not require annotator identities, and we use each item's annotations only as a set of labels. The LLM annotation is $f(x_i)$. For each item, let $\mathbf y_{i,h}$ be a panel of $k$ annotations from $H_i$, and let $\mathbf y_{i,m}$ be formed by replacing one annotation in $\mathbf y_{i,h}$ with $f(x_i)$. Stacking these item-level panels gives two annotation matrices: the human-only matrix $\mathbf Y_h$ and the mixed matrix $\mathbf Y_m$, both of dimension $N\!\times\!k$. Let $R$ be an agreement or reliability score applied to an annotation matrix, where higher values indicate a more reliable annotation panel. We define $1\!-\!R$ as the disagreement level. In the main experiments, we use Krippendorff's $\alpha$, the standard chance-corrected IAA coefficient for nominal and ordinal labels; we give its formal definition in Appendix~\ref{sec:agreement-notation}. Finally, we define:
\[
    R_h = R(\mathbf Y_h), \qquad
    R_m = R(\mathbf Y_m), \qquad
    \Delta = R_m - R_h.
\]
The substitution gap $\Delta$ measures the effect of replacement; a positive value means that the mixed panel is more reliable than the human-only panel.

\subsection{The \anlt{}}
\label{subsec:test}

The \anlt{} tests whether an LLM can replace one human annotation per item without reducing panel reliability by more than an acceptable margin. It uses a paired bootstrap: each bootstrap iteration samples $N$ items with replacement, draws $k$ human annotations for each sampled item, replaces one of them with the LLM annotation, and compares the reliability of the human-only and mixed panels using a one-sided percentile-bootstrap non-inferiority test. Figure~\ref{fig:overview} illustrates the procedure and Appendix~\ref{app:impl} provides the full algorithm.

\paragraph{Paired bootstrap.}
In each bootstrap iteration $b=1,\!\ldots\!,B$, we sample $N$ items with replacement from $\mathcal D$, denoted $\tilde{x}_1^{(b)}\!,\ldots\!,\tilde{x}_N^{(b)}$. For each sampled item $\tilde{x}_j^{(b)}$, let $\mathbf y_{j,h}^{(b)}$ be a panel of $k$ human annotations drawn without replacement from its available annotations. Let $\mathbf y_{j,m}^{(b)}$ be formed by replacing one annotation in $\mathbf y_{j,h}^{(b)}$ with the LLM annotation for that item. Stacking the human-only and mixed panels gives $\mathbf Y_h^{(b)}$ and $\mathbf Y_m^{(b)}$. Then, given an adaptive margin $\varepsilon^{(b)}$ defined below, compute:
\[
    R_h^{(b)}=R(\mathbf Y_h^{(b)}), \qquad
    R_m^{(b)}=R(\mathbf Y_m^{(b)}), \qquad
    \widehat{\Delta}^{(b)}=R_m^{(b)}-R_h^{(b)}, \qquad
    \widehat{\Delta}_{\mathrm{adj}}^{(b)}
    =
    \widehat{\Delta}^{(b)}+\varepsilon^{(b)}.
\]

\paragraph{Adaptive margin.}
\label{subsec:margin}
Researchers who apply the \anlt{} can accept a small loss in the IAA score of the mixed panel. This reflects the practical trade-off behind replacement: LLM annotations are cheaper and easier to obtain than additional human annotations, so replacement can be useful even when it causes a small decrease in agreement. We express the largest acceptable loss as a non-inferiority margin $\varepsilon^{(b)}$. Rather than using the same fixed margin for all datasets, we set the margin adaptively. In well-defined or objective annotation tasks, human annotators are expected to agree strongly, and replacement should preserve this high reliability. In harder or more subjective tasks, human agreement is naturally lower, so the margin accommodates greater fluctuation in the mixed-panel agreement. We therefore scale the margin by the observed human disagreement, $1-R_h^{(b)}$:
\[
    \varepsilon^{(b)}
    =
    c\frac{k_c}{k}\bigl(1-R_h^{(b)}\bigr).
    \label{eq:margin}
\]
This margin has a direct interpretation. The constant $c$ is chosen by the researchers, and it is the largest proportional increase in human disagreement that they accept at the calibration panel size $k_c$. For example, with $c=0.1$, the mixed panel may increase disagreement by at most $10\%$. In Section~\ref{sec:sim}, we use simulations to calibrate a default value of $c$. The factor $k_c/k$ extends this margin to other panel sizes. This correction is needed because the effect of replacing one annotation scales as $O(1/k)$, i.e., diminishes as $k$ grows: replacing one annotation out of three changes $R_m$ more than replacing one annotation out of ten. We derive this scaling in Appendix~\ref{sec:panel-size-scaling}. Multiplying by $k_c/k$ keeps the meaning of $c$ comparable across panel sizes.

\paragraph{Statistical test.}
The \anlt{} is a non-inferiority percentile-bootstrap test \citep[Eq.~13.5, p.~171]{efron1993introduction}. The null hypothesis is that replacing one human annotation with the LLM reduces agreement by more than the margin:
\[
H_0: R_m+\varepsilon\le R_h
\quad(\text{i.e., } \Delta+\varepsilon\le0)
\quad\text{\textbf{vs.}}\quad
H_1: R_m+\varepsilon>R_h
\quad(\text{i.e., } \Delta+\varepsilon>0).
\]
Here, $R_h$ and $R_m$ are the population values of the agreement scores, and the bootstrap values $R_h^{(b)}$ and $R_m^{(b)}$ estimate them. The replaced annotation is drawn uniformly at random from the item's annotations, so the test compares the LLM with a typical annotator of the pool. Let $q_\alpha^*$ be the empirical $\alpha$-quantile of the adjusted bootstrap gaps $\{\widehat{\Delta}_{\mathrm{adj}}^{(b)}\}_{b=1}^B$. Throughout the paper we use $\alpha=0.05$. We reject $H_0$ and conclude that the LLM passes the \anlt{} when $q_\alpha^*>0$.

\paragraph{Comparing judges.}
\label{subsec:compareLLMs}
While the test gives a binary replacement decision, the same bootstrap estimates naturally support comparison between LLM judges, regardless of whether they pass the test. We suggest comparing judges by the mean mixed-panel agreement. It should be used to compare judges within the same dataset, and read it alongside the human-only agreement $R_h$.
\[
    \overline{R}_m=\frac{1}{B}\sum_{b=1}^B R_m^{(b)}.
\]
\paragraph{Consistency and validity.}
We theoretically establish that the \anlt{} is a consistent test of asymptotic level $\alpha$. Consistency means that $q_\alpha^*$ converges to $\Delta_{\mathrm{adj}}$, and that the probability of correctly accepting a valid replacement asymptotically approaches $1$. Asymptotic level $\alpha$ means that when the LLM is not a valid replacement, the probability of falsely accepting it is asymptotically bounded by $\alpha$. Under the standard regularity conditions stated in Appendix~\ref{app:theoretical-guarantees}, it follows that:
\[
q_\alpha^*\xrightarrow{p}\Delta_{\mathrm{adj}},
\qquad
\Delta_{\mathrm{adj}}\le0\Rightarrow
\limsup_{N\to\infty}\Pr(q_\alpha^*>0)\le\alpha,
\quad\text{and}\quad
\Pr(q_\alpha^*>0)\to
\begin{cases}
0, & \Delta_{\mathrm{adj}}<0,\\
1, & \Delta_{\mathrm{adj}}>0,
\end{cases}
\]
Thus, as the number of items and the number of bootstrap iterations grow, the test accepts valid replacements, and invalid replacements fail it, with probability approaching one, while false acceptances under the null remain controlled. The formal theorem and its proof are provided in Appendix~\ref{app:theoretical-guarantees}. The proof follows standard percentile-bootstrap arguments, adapted to account for two test-specific cases: the IAA-based statistic rather than a simple mean, and, within each bootstrap, within-item sampling of annotations without replacement.

\begin{algorithm}[H]
\caption{\anlt{}}
\label{alg:altra-bootstrap}
\begin{algorithmic}[1]
\State \textbf{Input:} annotated dataset $\mathcal D=\{x_1,\!\ldots\!,x_N\}$, LLM $f(x_i)$, panel size $k$, margin constant $c$ calibrated for $k_c$, agreement score $R$, bootstrap iterations $B$, significance level $\alpha$.
\For{$b=1,\ldots,B$}
    \State Sample $N$ items with replacement from $\mathcal D$, denoted $\tilde{x}_1^{(b)},\ldots,\tilde{x}_N^{(b)}$.
    \For{each sampled item $\tilde{x}_j^{(b)}$}
        \State Draw without replacement $k$ human annotations of $\tilde{x}_j^{(b)}$, forming $\mathbf y_{j,h}^{(b)}$.
        \State Form the mixed panel $\mathbf y_{j,m}^{(b)}$ by replacing one random annotation with $f(\tilde{x}_j^{(b)})$.
    \EndFor
    \State Stack the panels into $\mathbf Y_h^{(b)}$ and $\mathbf Y_m^{(b)}$, and compute:
    \[
        R_h^{(b)}=R(\mathbf Y_h^{(b)}), \qquad
        R_m^{(b)}=R(\mathbf Y_m^{(b)}).
    \]
    \State Compute the adjusted substitution gap:
    \[
        \widehat{\Delta}_{\mathrm{adj}}^{(b)}
        = \widehat{\Delta}^{(b)} + \varepsilon^{(b)}=
        \bigl(R_m^{(b)}-R_h^{(b)}\bigr)
        +
        c\frac{k_c}{k}\bigl(1-R_h^{(b)}\bigr).
    \]
\EndFor
\State Let $q_\alpha^*$ be the empirical $\alpha$-quantile of $\{\widehat{\Delta}_{\mathrm{adj}}^{(b)}\}_{b=1}^{B}$.
\State Compute $\Rm=B^{-1}\sum_{b=1}^{B}R_m^{(b)}$.
\State \textbf{Return:} binary decision $\mathbf{1}\{q_\alpha^*>0\}$ and comparison score $\Rm$.
\end{algorithmic}
\end{algorithm}

\subsection{Variants for Annotator Perspectives}
\label{subsec:variants}

The \anlt{} examines whether the LLM preserves the signal that the annotators share, which is our main focus (see Section~\ref{sec:intro}). When the goal is instead to represent the different perspectives of the annotators, the test is not sufficient, since the replaced annotation is drawn from the whole pool, and an LLM can pass while it is not aligned with a smaller group of annotators (see Section~\ref{subsec:sim-meaning}). For this goal, we propose two variants of the test.

\paragraph{Group-wise testing.}
When each annotation carries a group label, such as a demographic attribute, and the LLM should align with each group, the test can be run separately within each group, with panels drawn only from that group's annotations. This requires at least $k$ annotations per item in each tested group. Researchers can then require the LLM to pass in every group or in specific groups.

\paragraph{Equivalence.}
Without group labels, a stricter criterion can be used. An LLM that makes the annotations more consistent than those of the humans is as undesirable as one that makes them less consistent, as it hides the variation among annotators. The equivalence variant, \textsc{Eq}-\anlt{}, therefore tests whether the substitution gap is within the margin in both directions, and the LLM passes only if it does not increase agreement by more than the margin (see Appendix~\ref{app:impl-altra}).

\section{Simulation-Based Analysis}
\label{sec:sim}

Simulation studies of statistical tests typically estimate Type I error and power under the null and alternative hypotheses of the test. For the \anlt{}, these hypotheses are defined by IAA scores, which are only a proxy for the replacement question, since they do not measure whether the LLM is closer than a human annotator to the ground-truth label. In this section, we go beyond this proxy and answer the replacement question directly. We generate the ground-truth label of each item and control the noise of the humans and of the LLM. We define the better-model pass rate and the worse-model pass rate, where a better model is a simulated LLM that is less noisy than the simulated humans and a worse model is one that is noisier. A good test should accept better models, and worse models should fail it. These measures resemble power and Type I error, but they are defined by ground-truth accuracy, which also gives a common criterion for comparing tests with different hypotheses.

We design the simulations in three stages. We start with a basic setup, in which every item has one ground-truth label, all annotators make the same kind of errors, and every annotator labels every item. This setup satisfies the requirements of the existing tests, so we use it to compare the \anlt{} with them and to calibrate its margin (see Section~\ref{subsec:sim-results}). We then relax each of these assumptions and vary the parameters of the test and of the task, to examine whether the results still hold (see Section~\ref{subsec:sim-robust}). Finally, we examine what passing the test means in two cases in which a ground-truth label is less meaningful, when annotators form groups with different perspectives and when human agreement is very low (see Section~\ref{subsec:sim-meaning}).

\subsection{Experimental Setup}

\textbf{Data generation.} We simulate annotation tasks with a known ground-truth label for each item. For each simulated dataset, we first generate the true item labels and then generate human and LLM annotations by adding noise. The human annotators share a noise level, $\eta_h$, and the LLM has its own noise level, $\eta_f$. Lower noise means a higher probability of providing the ground-truth label, so $\eta_f<\eta_h$ defines a better-model case and $\eta_f>\eta_h$ defines a worse-model case. Note that nothing in this simulation is specific to humans or LLMs, as we simulate a pool of annotators, which we call the humans, and a single candidate annotator, which we call the LLM.

Following \citet{calderon2025alternative}, each simulation begins by drawing the prior probability of each label, and then drawing a ground-truth label for each item. This produces a synthetic dataset with known ground-truth labels. We then generate human and LLM annotations as follows: with probability $1\!-\!\eta$, the annotator returns the ground-truth label; with probability $\eta$, it returns a label sampled from the same label prior. In the basic setup, we draw the label prior from a Dirichlet distribution and generate a corrupted annotation as follows:
\[
\pi \sim \mathrm{Dirichlet}(\mathbf{1}_L), \qquad
y_i \sim \mathrm{Categorical}(\pi), \qquad
\tilde{y}_{i,\eta} \sim \mathrm{Categorical}\big((1-\eta)e_{y_i}+\eta\pi\big),
\]
where $L$ is the number of labels and $e_{y_i}$ is the one-hot vector for the ground-truth label. Human annotations are generated as $h_{ij}=\tilde{y}_{i,\eta_h}$, while the LLM annotation is generated as $f_i=\tilde{y}_{i,\eta_f}$.

\textbf{Parameters.} In the basic setup, we use $L=4$ labels, $m=6$ human annotators per item, and $k=3$ human annotations in the \anlt{}, which is also the calibration panel size ($k_c=3$). We vary $\eta_h$ and $\eta_f$ over $\{0.1,0.2,0.3,0.4,0.5\}$, giving cases where the LLM is better than, worse than, or equal to the human annotators, with human
agreement spanning $0.14$ to $0.87$. For each pair of noise levels, we generate multiple datasets and item subsamples, and measure agreement using Krippendorff's $\alpha$. Unless otherwise stated, we use $N\!=\!200$ items and test at $\alpha\!=\!0.05$. The full details of all the simulations are in Appendix~\ref{app:impl-sim}.

\textbf{Baselines.} We compare the \anlt{} with the two known statistical procedures closest to the replacement question. Both procedures require the same annotators to annotate all items. 
The first baseline is the \alttest{}~\citep{calderon2025alternative}. It uses a leave-one-annotator-out procedure: each annotator is held out in turn, and a statistical test compares whether the LLM is non-inferior, up to a margin $\varepsilon$, to the held-out annotator in its alignment with the remaining annotators. The second procedure is STT~\citep{he2025can}, which uses a bootstrap over items. It first computes the IAA of the all-human panel. Then, for each annotator, it replaces all their annotations with the LLM ones and computes the mean IAA of the resulting mixed panels. It uses Two One-Sided Tests (TOST; \citealp{rogers1993using}) to test whether the all-human and mixed-panel IAAs are equivalent, i.e., within a margin defined by a $\mathrm{frac}$ parameter. We evaluate STT-Alt, a non-inferiority adaptation of STT that uses only the lower one-sided test and for completeness, we also compare the \textsc{Eq}-\anlt{} (see Section~\ref{subsec:variants}) with the original STT. Appendix~\ref{app:baselines} provides full implementation details.

\begin{figure}[t]
  \centering
  \begin{subfigure}[t]{0.48\textwidth}
    \centering
    \includegraphics[width=\textwidth]{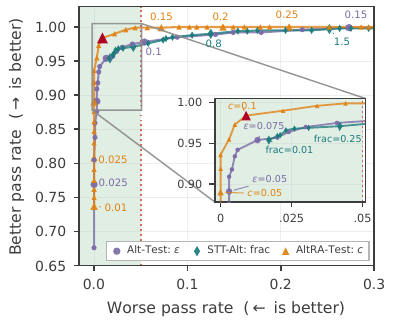}
    \phantomcaption
    \label{fig:sim-pareto}
  \end{subfigure}
  \hfill
  \begin{subfigure}[t]{0.48\textwidth}
    \centering
    \includegraphics[width=\textwidth]{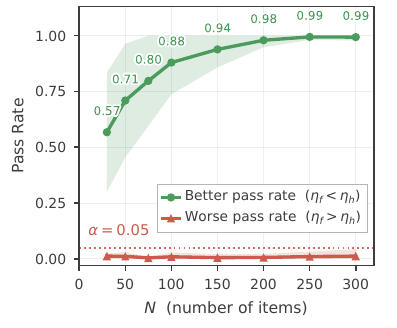}
    \phantomcaption
    \label{fig:sim-n-sweep}
  \end{subfigure}
 \caption{\textit{Left:} The \anlt{} against the baselines, simulated with $N\!=\!200$ items and $k\!=\!3$, varying each procedure's hyperparameter: $c$, $\varepsilon$, or $\mathrm{frac}$. The X-axis shows the worse-model pass rate and the Y-axis the better-model pass rate. The green region marks the valid regime, where the worse-model pass rate is below $0.05$. As shown, the \anlt{} dominates the baselines. \textit{Right:} Using the calibration in the left panel, we set $c\!=\!0.1$ and vary the number of items $N$. From $N\!=\!100$ on, the \anlt{} achieves reliable better-model pass rates ($>0.85$).}
  \label{fig:sim-main}
\end{figure}

\subsection{Simulation Results}
\label{subsec:sim-results}

\textbf{\anlt{} gives the best replacement trade-off.} We first calibrate the margin parameter for each test: $c$ for \anlt{}, $\varepsilon$ for \alttest{}, and $\mathrm{frac}$ for STT-Alt. Figure~\ref{fig:sim-pareto} compares the procedures at $N=200$ by sweeping these parameters and plotting the better-model pass rate against the worse-model pass rate. We focus on the valid region, where the worse-model pass rate is below $\alpha=0.05$. In this region, the \anlt{} gives a Pareto-dominant trade-off over the other tests: at the same worse-model pass rate, it has a higher better-model pass rate than the baselines. This means that in the simulated settings, the \anlt{} is more likely to accept LLMs that truly improve over the human annotators. Appendix~\ref{app:sim-results} shows the same pattern for sample sizes~$N\in\{50,100,150,300\}$ and for other generative families. We hypothesize that the \anlt{} achieves a better trade-off because it uses the annotations more efficiently. It pools all the annotators into a single statistic, while the existing tests split them by annotators. In line with this, the advantage is largest when there are few items (see Appendix~\ref{app:sim-baselines}). Finally, for equivalence, the \textsc{Eq}-\anlt{} performs at least as well as STT (see Appendix~\ref{app:persp-eq-sim}).

\textbf{The margin constant $c\!=\!0.1$ is conservative and effective.} As shown in Figure~\ref{fig:sim-pareto}, $c\!=\!0.1$ lies near the knee of the \anlt{} curve and remains inside the valid region; see also Figure~\ref{fig:app-calib-c}. At this value, the better-model pass rate is $0.98$, while the worse-model pass rate remains below $0.05$. Since $c\!=\!0.1$ has a direct interpretation, permitting a loss of agreement equal to $10\%$ of the observed human disagreement, we recommend it as the default and use it in our experiments.

\textbf{More items mainly improve better-model detection.} Figure~\ref{fig:sim-n-sweep} shows the behavior of the \anlt{} as the number of items $N$ increases, using $c\!=\!0.1$. Like traditional power, adding items mainly improves the better-model pass rate: it reaches $0.88$ at $N\!=\!100$, $0.94$ at $N\!=\!150$, and $0.98$ at $N\!=\!200$. The worse-model pass rate remains below $\alpha\!=\!0.05$ across the examined range.


\subsection{Robustness of the Test}
\label{subsec:sim-robust}

In this section, we examine whether the results hold beyond the basic setup. We briefly describe the main findings here, and provide the full results in Appendix~\ref{app:sim-results}.

\textbf{Test and task parameters.} Across all the examined numbers of annotators per item $m$, panel sizes $k$, and numbers of labels $L$, the worse-model pass rate remains below $0.05$ (Figure~\ref{fig:app-struct}). A larger $m$ stabilizes the better-model pass rate and a larger $L$ increases it, so tasks with few annotations per item or with binary labels require more items. A larger $k$ provides only modest gains, and $k=3$ is the most annotation-efficient choice. Among the agreement metrics, Krippendorff's $\alpha$ and Fleiss' $\kappa$ yield the highest better-model pass rates (Figure~\ref{fig:app-metric}). The comparison with the baselines also holds for other generative families (Figures~\ref{fig:app-dist-uniform} and~\ref{fig:app-dist-gaussian}).

\textbf{Annotator bias.} In the basic setup, all annotators make the same kind of errors. When every human annotator has a personal bias toward specific labels, the \anlt{} still outperforms the baselines (Figure~\ref{fig:app-bias}). We also simulate a bias that the humans and the LLM share, which is a limitation of every test that relies on human annotations. Also in this case, the worse-model pass rate of the \anlt{} remains below $0.05$ (Figure~\ref{tab:app-shared-bias}).

\textbf{Sparse annotations.} To enable the comparison with the baselines, in all the simulations above every annotator labels every item. However, the \anlt{} is designed for sparse annotations, in which every annotator labels only a few items and the baselines cannot be applied. We therefore vary the number of items that each annotator labels, from $1$ to all items. The pass rates do not change materially (Figure~\ref{fig:app-sparse}), so the \anlt{} behaves the same with sparse annotations.

\subsection{Annotator Perspectives and Low Human Agreement}
\label{subsec:sim-meaning}

Finally, we examine two cases in which the result of the test should be interpreted with care. The full results are in Appendices~\ref{app:persp-clusters} and~\ref{app:sim-behavior}.

\textbf{Annotators with different perspectives.} In subjective tasks, annotators may form groups that hold different perspectives. We simulate two such groups, a larger one and a smaller one, each with its own ground-truth label for every item, and an LLM that follows the larger group, the smaller group, or a mixture of them (see Appendix~\ref{app:persp-clusters}). As shown in Figure~\ref{fig:app-clusters}, the test behaves as expected when the goal is to preserve the shared signal. An LLM that follows the larger group passes at high rates. In contrast, an LLM that follows the smaller group rarely passes when the larger group dominates the pool. When researchers want the LLM to represent each perspective, we recommend using the variants of the test (see Section~\ref{subsec:variants}).

\textbf{Low human agreement.} Since the margin increases with human disagreement, an uninformative LLM may pass when human agreement is very low. We find that a random LLM never passes when the human IAA is above $0.1$, and starts to pass below this level (see Appendix~\ref{app:sim-behavior}). We therefore recommend applying the test only when the human IAA is at least $0.2$.

\section{Real-World Experiments}
\label{sec:exp}

\subsection{Experimental Setup}
\label{sec:setup}

We evaluate the \anlt{} on 20 datasets spanning text (12), image (5), audio (2), and video (1). The datasets cover nominal and ordinal tasks with two to ten labels, at least three annotations per item, and human agreement ranging from $\Rh=0.01$ to $0.90$. We use a subset of $300$ items per dataset and use only items with at least  three human labels. Table~\ref{tab:datasets-app} describes all $20$ datasets.


We evaluate $18$ LLMs as candidate annotators, assigning to each modality the
models that natively accept its input: Qwen3-8B (Q3-8b) and Qwen3-14B (Q3-14b),
Llama-3.2-3B (L3.2-3b) and Llama-3.1-8B (L3.1-8b), and Gemma-3-4B (G3-4b) and
Gemma-3-12B (G3-12b) for text; the vision--language models Qwen2.5-VL-3B
(Q2.5V-3b) and Qwen2.5-VL-7B (Q2.5V-7b), Llama-3.2-11B-Vision (L3.2V-11b) and
the two Gemma-3 models for image, with the Qwen2.5-VL pair also handling video
through sampled frames; and Qwen2.5-Omni-3B (Q2.5-O-3b) and Qwen2-Audio-7B
(Q2-A-7b) for audio. Alongside these $11$ open-weight models, spanning roughly
$3$B to $14$B parameters, we evaluate seven proprietary judges: Gemini-3.1 Pro
(Gem-P), Gemini-3.7 Flash (Gem-F) and Gemini-3.5 Flash-Lite (Gem-L) on all four
modalities; GPT-5.6 Terra (GPT-T) and GPT-5.6 Luna (GPT-L) on text and image;
and GPT-Audio-1.5 (GPT-A) and GPT-Audio-Mini (GPT-A-M) on audio. This gives $11$
text, $10$ image, $7$ audio and $5$ video judges. Dataset details, references and links for all models are provided in Appendix~\ref{app:datasets}.

Each judge is evaluated using up to five prompting techniques \citep{mizrahi2024state, li2025generation}: zero-shot, few-shot, chain-of-thought~\citep{wei2022chain}, an ensemble of these three~\citep{nahum2025llms}, and GEPA prompt optimization~\citep{agrawal2026gepa}. GEPA is applied only to open-weight models because of its computational cost.
We run the \anlt{} using panels of $k=3$ annotations (with $k_c=3$), Krippendorff's $\alpha$ as the agreement coefficient, $B=300$ bootstrap iterations, a significance level of $\alpha=0.05$, and the simulation-calibrated margin constant $c=0.1$. Since real-world data have no gold labels with which $c$ can be chosen, we use the value calibrated in the simulations and do not vary it.

\begin{table}[t]\centering\scriptsize
\setlength{\tabcolsep}{1pt}\renewcommand{\arraystretch}{1.12}
\caption{\textbf{Main results.} Rows are datasets with their human-only agreement $\Rh$; columns are LLM judges, grouped by family. Each cell reports the mean mixed-panel agreement $\Rm$. Cells in \colorbox{passgreen}{green} mean the judge passes the \anlt{}; \textbf{bold} marks the highest $\Rm$ for that dataset.}
\label{tab:main-grid-rmixed}
\resizebox{\textwidth}{!}{%
\begin{tabular}{@{}p{3.5cm} rccccccccccc@{}}
\toprule
& & \multicolumn{3}{c}{\textbf{Gemini}} & \multicolumn{2}{c}{\textbf{GPT}} & \multicolumn{2}{c}{\textbf{Qwen}} & \multicolumn{2}{c}{\textbf{Gemma}} & \multicolumn{2}{c}{\textbf{Llama}} \\
\textbf{\faFile*\ Text} & $\boldsymbol{\bar R_h}$ & \textbf{Gem-P} & \textbf{Gem-F} & \textbf{Gem-L} & \textbf{GPT-T} & \textbf{GPT-L} & \textbf{Q3-8b} & \textbf{Q3-14b} & \textbf{G3-4b} & \textbf{G3-12b} & \textbf{L3.2-3b} & \textbf{L3.1-8b} \\
\midrule
DICES-350 \citep{aroyo2023dicesdatasetdiversityconversational} & 0.20 & $0.14$ & $0.06$ & $0.10$ & $0.16$ & $0.16$ & $0.09$ & $0.16$ & $0.12$ & \cellcolor{passgreen}$\mathbf{0.22}$ & $0.14$ & $0.15$ \\
EPIC \citep{frenda2023epic} & 0.20 & \cellcolor{passgreen}$0.22$ & \cellcolor{passgreen}$0.19$ & \cellcolor{passgreen}$\mathbf{0.24}$ & \cellcolor{passgreen}$0.22$ & \cellcolor{passgreen}$0.21$ & \cellcolor{passgreen}$0.20$ & \cellcolor{passgreen}$0.22$ & $-0.06$ & \cellcolor{passgreen}$0.19$ & $0.01$ & $0.02$ \\
MultiPref \citep{miranda2025hybridpreferenceslearningroute} & 0.28 & \cellcolor{passgreen}$\mathbf{0.33}$ & \cellcolor{passgreen}$0.30$ & \cellcolor{passgreen}$0.28$ & $0.24$ & $0.24$ & $0.19$ & \cellcolor{passgreen}$0.26$ & $0.14$ & $0.25$ & $0.20$ & $0.12$ \\
ChaosNLI \citep{nie2020learncollectivehumanopinions} & 0.28 & \cellcolor{passgreen}$0.31$ & \cellcolor{passgreen}$0.30$ & \cellcolor{passgreen}$0.29$ & \cellcolor{passgreen}$0.31$ & \cellcolor{passgreen}$0.26$ & \cellcolor{passgreen}$0.26$ & \cellcolor{passgreen}$\mathbf{0.32}$ & \cellcolor{passgreen}$0.28$ & \cellcolor{passgreen}$0.29$ & $0.23$ & \cellcolor{passgreen}$0.28$ \\
\href{https://huggingface.co/datasets/Anthropic/persuasion}{Persuasion} & 0.31 & $0.06$ & $0.07$ & $0.01$ & $0.15$ & $0.10$ & $0.12$ & $\mathbf{0.18}$ & $\mathbf{0.18}$ & $0.17$ & $0.10$ & $0.09$ \\
\href{https://huggingface.co/datasets/data-is-better-together/10k_prompts_ranked}{10k-Prompts} & 0.37 & \cellcolor{passgreen}$0.40$ & $0.36$ & \cellcolor{passgreen}$\mathbf{0.43}$ & $0.35$ & $0.35$ & \cellcolor{passgreen}$0.39$ & $0.15$ & $0.33$ & \cellcolor{passgreen}$0.39$ & $0.23$ & $0.28$ \\
Scruples \citep{lourie2021scruples} & 0.38 & \cellcolor{passgreen}$\mathbf{0.45}$ & \cellcolor{passgreen}$0.44$ & \cellcolor{passgreen}$0.43$ & \cellcolor{passgreen}$0.39$ & $0.35$ & $0.18$ & $0.10$ & $0.14$ & $0.27$ & $0.08$ & $0.23$ \\
SummEval \citep{fabbri2021summeval} & 0.60 & \cellcolor{passgreen}$\mathbf{0.65}$ & \cellcolor{passgreen}$0.61$ & $0.58$ & $0.54$ & $0.52$ & $0.51$ & $0.42$ & $0.42$ & $0.45$ & $0.40$ & $0.28$ \\
ConvAbuse \citep{curry2021convabusedataanalysisbenchmarks} & 0.72 & $\mathbf{0.73}$ & $0.71$ & $\mathbf{0.73}$ & $0.68$ & $0.71$ & $0.70$ & $0.65$ & $0.41$ & $0.61$ & $0.48$ & $0.58$ \\
CEBaB \citep{abraham2022cebab} & 0.72 & \cellcolor{passgreen}$\mathbf{0.79}$ & \cellcolor{passgreen}$\mathbf{0.79}$ & \cellcolor{passgreen}$\mathbf{0.79}$ & \cellcolor{passgreen}$\mathbf{0.79}$ & \cellcolor{passgreen}$\mathbf{0.79}$ & \cellcolor{passgreen}$0.78$ & \cellcolor{passgreen}$0.77$ & \cellcolor{passgreen}$0.78$ & \cellcolor{passgreen}$\mathbf{0.79}$ & \cellcolor{passgreen}$0.78$ & \cellcolor{passgreen}$0.78$ \\
\href{https://huggingface.co/datasets/yael-katsman/Loneliness-Causes-and-Intensity}{Loneliness} & 0.90 & $0.57$ & $\mathbf{0.61}$ & $0.43$ & $0.38$ & $0.44$ & $0.18$ & $0.39$ & $0.22$ & $0.38$ & $0.33$ & $0.22$ \\
\addlinespace[4pt]
\textbf{\faImage\,\faVideo\ Image \& Video} &  &  &  &  &  &  & \textbf{Q2.5V-3b} & \textbf{Q2.5V-7b} &  &  & \textbf{L3.2V-11b} &  \\
\midrule
T2I-DPO \citep{cai2025diffusion} & 0.12 & \cellcolor{passgreen}$0.14$ & \cellcolor{passgreen}$\mathbf{0.15}$ & \cellcolor{passgreen}$0.13$ & \cellcolor{passgreen}$0.14$ & \cellcolor{passgreen}$0.12$ & $0.00$ & $0.07$ & $0.01$ & $0.04$ & $0.02$ &  \\
\href{https://huggingface.co/datasets/datapointai/vibe-landing-page-arena}{Vibe-Arena} & 0.12 & \cellcolor{passgreen}$0.13$ & \cellcolor{passgreen}$0.11$ & \cellcolor{passgreen}$0.12$ & \cellcolor{passgreen}$\mathbf{0.14}$ & \cellcolor{passgreen}$0.11$ & $-0.05$ & $0.05$ & $-0.01$ & $-0.00$ & $-0.01$ &  \\
FER+ \citep{barsoum2016training} & 0.51 & \cellcolor{passgreen}$\mathbf{0.54}$ & \cellcolor{passgreen}$0.52$ & $0.46$ & $0.49$ & $0.49$ & $0.28$ & $0.31$ & $0.36$ & $0.38$ & $0.42$ &  \\
Flickr8k \citep{hodosh2013framing} & 0.87 & $0.82$ & $\mathbf{0.85}$ & $0.82$ & $0.83$ & $0.83$ & $0.65$ & $0.66$ & $0.80$ & $0.76$ & $0.64$ &  \\
CIFAR-10H \citep{peterson2019human} & 0.88 & $0.89$ & \cellcolor{passgreen}$\mathbf{0.90}$ & $0.88$ & $0.74$ & $0.65$ & $0.76$ & $0.79$ & $0.81$ & $0.82$ & $0.80$ &  \\
\midrule
T2V-Preference \citep{he2509videoscore2} & 0.04 & \cellcolor{passgreen}$\mathbf{0.08}$ & \cellcolor{passgreen}$\mathbf{0.08}$ & \cellcolor{passgreen}$0.06$ &  &  & $-0.01$ & \cellcolor{passgreen}$0.07$ &  &  &  &  \\
\addlinespace[4pt]
\textbf{\faVolumeUp\ Audio} &  &  &  &  & \textbf{GPT-A} & \textbf{GPT-A-M} & \textbf{Q2.5-O-3b} & \textbf{Q2-A-7b} &  &  &  &  \\
\midrule
\href{https://huggingface.co/datasets/datapointai/tts-human-preferences-medium}{TTS-Preference} & 0.01 & \cellcolor{passgreen}$\mathbf{0.02}$ & \cellcolor{passgreen}$0.01$ & \cellcolor{passgreen}$-0.00$ & $-0.03$ & \cellcolor{passgreen}$-0.00$ & \cellcolor{passgreen}$0.01$ & $-0.12$ &  &  &  &  \\
SOMOS \citep{maniati2022somos} & 0.05 & $-0.12$ & \cellcolor{passgreen}$0.04$ & \cellcolor{passgreen}$\mathbf{0.05}$ & \cellcolor{passgreen}$\mathbf{0.05}$ & \cellcolor{passgreen}$0.04$ & \cellcolor{passgreen}$0.03$ & $-0.00$ &  &  &  &  \\
\bottomrule
\end{tabular}
}
\end{table}

\subsection{Results}
\label{sec:results}

\textbf{In almost half of the setups LLMs pass the test.} Table~\ref{tab:main-grid-rmixed} reports the default CoT setting for 18 LLMs and 19 of the 20 datasets (the results for Measuring-Hate-Speech are in Appendix~\ref{app:real-full}). It presents the post-substitution agreement scores, $\Rm$, and highlights in green the cases where the LLM passes the \anlt{}. Under this setting, LLM judges pass in 71 of the 201 judge--dataset pairs (35.3\%). When allowing each judge to use its best prompting technique, this number increases to 95 of 201 (47.3\%; Appendix~\ref{app:real-results}). The outcome is strongly dataset-dependent, for example, all judges pass on CEBaB, whereas none pass on Persuasion, Measuring-Hate-Speech, Flickr8k-Expert, or Loneliness. Proprietary judges pass most often. With at least one prompting technique, Gemini-3.1-Pro passes 16/20 datasets (80\%), while Gemini-3.7-Flash and Gemini-3.5-Flash-Lite pass 14/20 each (70\%). GPT-5.6-Luna and GPT-5.6-Terra pass 9/17 (52.9\%) and 8/17 (47.1\%), respectively. The strongest open-weight judge, Gemma-3-12B, also passes 8/17 (47.1\%).

\textbf{Passing often means higher agreement.} Under CoT, 49 of the 71 passing cells have $\Rm>\Rh$, and 59 have $\Rm\ge\Rh$. In these cases, replacing one human annotation with the LLM preserves or increases the agreement of the annotation panel, suggesting that the LLM is not merely an acceptable substitute but a better alternative to recruiting another annotator from the same pool. However, a higher $\Rm$ alone is not a replacement decision. In four CoT cases, $\Rm\ge\Rh$ but the judge still fails the test because the uncertainty is too large. Conversely, 12 passing cells have $\Rm<\Rh$, meaning that a small loss can still be statistically acceptable under the margin. The test is therefore needed to separate apparent changes in agreement from statistically supported replacement decisions.

\textbf{Passing agrees with accuracy when gold labels exist.} Real-world datasets usually have no gold labels, so we cannot tell whether an LLM that passes is a better annotator than the humans. CIFAR-10H~\citep{peterson2019human} is an exception, since it also includes the original CIFAR-10 labels~\citep{krizhevsky2009learning}. On its evaluated items, the average accuracy of a human annotator is $0.94$. Of the $9$ combinations of LLM and prompting technique that are at least as accurate as the humans, $8$ pass the test, and none of the $36$ less accurate combinations passes (see Appendix~\ref{app:real-full}). These results show that, also in real-world data, the test accepts LLMs that are at least as accurate as the humans and does not accept less accurate ones, as we found in the simulations.

\textbf{Low human agreement limits what passing the test means.} The datasets span a wide range of human agreement. Most of our generated-output preference datasets have $\Rh$ below 0.2, consistent with prior work showing that preferences are heterogeneous and weakly agreed upon~\citep{calderon2025alternative,calderon2025multidomain,nussbaumhoffer2026counterfactual}. We report the results on these datasets for completeness. Many subjective tasks, such as irony, safety, persuasiveness, and moral judgment, fall in a moderate range of $\Rh=0.2$--$0.6$ \citep{lissak2024colorful}. Notably, $\Rh$ does not determine the result of the \anlt{}: LLMs can pass or fail at both low and high agreement levels. 
The \anlt{} is relative to the observed human pool. When $\Rh$ is around 0.2 or higher, a passing result indicates that the LLM is aligned with a non-trivial signal in the human pool. When $\Rh$ is near zero, a pass supports cost-equivalent substitution, and using an LLM will not change much. This is the key nuance: the \anlt{} evaluates replacement, not whether the annotations contain a useful signal. If the human pool does not exhibit such a signal, replacement, although justified, may not be meaningful. In such cases, we recommend first obtaining annotations with at least fair agreement ($\Rh \ge 0.2$) before testing for replacement. The simulations support this threshold, since a random LLM starts to pass only when human agreement is below $0.1$ (see Section~\ref{subsec:sim-meaning}).

\begin{table}[t]\centering\scriptsize
\setlength{\tabcolsep}{2pt}\renewcommand{\arraystretch}{1.15}
\caption{\textbf{Prompting techniques and minimum item count.} For each judge, the table reports the number of datasets on which it passes under \textit{at least one} prompting technique, the pass rate (in \%) of each technique, and the average minimum number of items needed to pass among datasets where it passes at $N\!=\!300$. \textbf{Bold} marks the judge's best prompting technique.}
\label{tab:method-passrate}
\resizebox{\textwidth}{!}{%
\begin{tabular}{@{}l!{\color{black!25}\vrule width 0.4pt}ccc!{\color{black!25}\vrule width 0.4pt}cc!{\color{black!25}\vrule width 0.4pt}cccc!{\color{black!25}\vrule width 0.4pt}cc!{\color{black!25}\vrule width 0.4pt}cc@{}}
\toprule
& \multicolumn{3}{c!{\color{black!25}\vrule width 0.4pt}}{\textbf{Gemini}} & \multicolumn{2}{c!{\color{black!25}\vrule width 0.4pt}}{\textbf{GPT}} & \multicolumn{4}{c!{\color{black!25}\vrule width 0.4pt}}{\textbf{Qwen}} & \multicolumn{2}{c!{\color{black!25}\vrule width 0.4pt}}{\textbf{Gemma}} & \multicolumn{2}{c}{\textbf{Llama}} \\
 & \textbf{Gem-P} & \textbf{Gem-F} & \textbf{Gem-L} & \textbf{GPT-T} & \textbf{GPT-L} & \textbf{Q3-8b} & \textbf{Q3-14b} & \textbf{Q2.5V-3b} & \textbf{Q2.5V-7b} & \textbf{G3-4b} & \textbf{G3-12b} & \textbf{L3.2-3b} & \textbf{L3.1-8b} \\
\midrule
\#\,passed & $16/20$ & $14/20$ & $14/20$ & $8/17$ & $9/17$ & $4/12$ & $5/12$ & $1/6$ & $3/6$ & $2/17$ & $8/17$ & $2/12$ & $2/12$ \\
\addlinespace[2pt]
Zero-shot & $65\%$ & $55\%$ & $50\%$ & $\mathbf{41\%}$ & $41\%$ & $33\%$ & $\mathbf{42\%}$ & $\mathbf{17\%}$ & $\mathbf{33\%}$ & $12\%$ & $29\%$ & $8\%$ & $17\%$ \\
Few-shot & $\mathbf{80\%}$ & $50\%$ & $50\%$ & $\mathbf{41\%}$ & $29\%$ & $33\%$ & $33\%$ & $0\%$ & $\mathbf{33\%}$ & $12\%$ & $24\%$ & $8\%$ & $17\%$ \\
CoT & $60\%$ & $\mathbf{65\%}$ & $\mathbf{55\%}$ & $35\%$ & $29\%$ & $33\%$ & $33\%$ & $0\%$ & $17\%$ & $12\%$ & $29\%$ & $8\%$ & $17\%$ \\
Ensemble & $70\%$ & $\mathbf{65\%}$ & $50\%$ & $\mathbf{41\%}$ & $\mathbf{53\%}$ & $33\%$ & $\mathbf{42\%}$ & $\mathbf{17\%}$ & $17\%$ & $12\%$ & $\mathbf{41\%}$ & $\mathbf{17\%}$ & $17\%$ \\
GEPA & \xmarkg & \xmarkg & \xmarkg & \xmarkg & \xmarkg & $33\%$ & $33\%$ & $0\%$ & $0\%$ & $12\%$ & $29\%$ & $8\%$ & $17\%$ \\
\midrule
min $N$ & $100$ & $111$ & $96$ & $122$ & $133$ & $75$ & $70$ & $225$ & $117$ & $75$ & $147$ & $138$ & $88$ \\
\bottomrule
\end{tabular}
}
\end{table}

\textbf{Prompting helps, but rarely changes the decision.} Prompting changes some verdicts, and its effect depends on both the model and the dataset. Overall, ensemble prompting, i.e., majority vote over different prompts, gives the highest aggregate pass count: 83 of 201 (41.3\%), compared with 37.3\% for zero-shot and 35.3\% for both few-shot and CoT. GEPA, evaluated only for open-weight models, does not increase replacement decisions (18.8\% for GEPA compared with 19.8\% for CoT when considering the same models). Furthermore, in Appendix~\ref{app:real-results}, we report additional experiments varying the number of few-shot examples and the CoT reasoning effort. We find that more examples or more reasoning can help, but the trends are mixed and dataset-dependent.

\textbf{About 100 items often suffice to pass.} We next ask how many items are required for reliable \anlt{} results in real-world settings. Unlike simulations, real data do not provide ground truth about whether an LLM should pass. We therefore use the 300-item experiments as the reference verdict and ask, for LLMs that pass with 300 items, what was the minimum number of items for which they pass. Table~\ref{tab:method-passrate} reports these minimum passing sizes. Across judges, the median is 111 items and the mean is 115, with all values between 70 and 150 except for Qwen2.5-VL-3B, which reaches 225. Appendix~\ref{app:real-results} shows that decisions often stabilize around 100 items.

\subsection{Annotator Perspectives in Real Data}
\label{subsec:real-perspectives}

In Section~\ref{subsec:sim-meaning}, we show with simulations that passing the test means that the LLM preserves the signal that the pool of annotators shares, and that an LLM can pass while it is not aligned with a smaller group of annotators. However, when researchers want the LLM to represent the perspectives of different groups of annotators, and not only the shared signal, passing the test with the whole pool is not sufficient. For such cases, we propose two variants of the test (see Section~\ref{subsec:variants}). In group-wise testing, the test is applied separately within each group, and in the equivalence variant, an LLM that increases the agreement also fails. We next apply group-wise testing to real-world data, and examine whether LLMs that pass the test are also aligned with each group of annotators. We evaluate the equivalence variant with simulations in Appendix~\ref{app:persp-eq-sim}.

We use DICES-350, a conversational-safety dataset in which the demographics of the annotators define $13$ groups, by age, education, gender, and race. When panels are drawn from the whole pool and every LLM uses its best prompting technique, four LLMs pass the test, Gemma-3-12B, Gemini-3.1-Pro, GPT-5.6-Terra, and GPT-5.6-Luna. When we require an LLM to pass in every group, only Gemma-3-12B passes, and the other three fail in the Latinx group (see Appendix~\ref{app:persp-dices}). Group-wise testing therefore exposes a misalignment with a specific group, which pooled testing averages out.

\section{Conclusion}
\label{sec:conclusion}


As LLMs are increasingly used in place of human annotators, their efficiency must be accompanied by statistical evidence that the replacement preserves annotation quality. We propose \anlt{}, the Alternative Annotator Test with Resampled Annotations, which formulates this as a non-inferiority test: it asks whether the inter-annotator agreement (IAA) of a mixed human-LLM panel is statistically non-inferior to that of an all-human panel. Unlike prior replacement tests, \anlt{} can be applied to sparse annotation settings, where different items are labeled by different subsets of annotators and no shared panel is required. 

In simulations, \anlt{} achieves a stronger trade-off than existing procedures: for the same rate of false replacements, it accepts more valid ones. Around 100 annotated items often suffice for the test. Replacement is possible in many settings, but the outcome depends more on the dataset than on the judge or prompting technique. The main implication is that LLM annotators should be validated per task rather than adopted as general practice.

Because the test is relative to the observed human pool, its verdict should always be reported alongside human IAA: when human agreement is itself very low, a passing result does not show that the labels contain a useful signal. For the same reason, the test cannot detect errors that the LLM shares with the human annotators. In addition, passing the test means that the LLM preserves the signal that the annotators share, and when the goal is to represent the perspectives of different groups of annotators, the group-wise and equivalence variants of the test should be used. Finally, in this paper, the LLM provides a single annotation for each item. We leave for future work an extension of the test that considers the distribution of the LLM annotations for each item, and examines whether it is aligned with the distribution of the human annotations. Such an extension is also suitable for the goal of representing the perspectives of the annotators. We hope \anlt{} helps researchers make more rigorous and transparent decisions about when LLMs can replace human annotators.

\bibliography{iclr2027_conference}
\bibliographystyle{iclr2027_conference}

\appendix


\etocsettocdepth.toc{subsection} 
\etocsettocstyle{\section*{\textbf{Appendix Contents}}}{}
\tableofcontents

\section{Additional Implementation Details}
\label{app:impl}

In this section, we provide additional implementation details. We start with the notation (Table~\ref{tab:notation}) and the full algorithm of the \anlt{} (Algorithm~\ref{alg:altra-bootstrap}). Then, in Appendix~\ref{app:baselines}, we describe the existing tests and how we implement them. In Appendix~\ref{app:impl-sim}, we provide the details of the simulations, and finally, in Appendix~\ref{app:datasets}, we describe the datasets, the LLMs, and the prompting techniques.




{\small
\setlength{\tabcolsep}{6pt}
\renewcommand{\arraystretch}{1.1}
\begin{longtable}{p{0.26\linewidth}p{0.66\linewidth}}
\caption{Main notation used in the paper.}
\label{tab:notation}\\
\toprule
\textbf{Notation} & \textbf{Meaning} \\
\midrule
\endfirsthead

\multicolumn{2}{@{}l}{\emph{Table~\ref{tab:notation}, continued}}\\
\toprule
\textbf{Notation} & \textbf{Meaning} \\
\midrule
\endhead

\midrule
\multicolumn{2}{r@{}}{\emph{continued on next page}}\\
\endfoot

\bottomrule
\endlastfoot

\(\mathcal D=\{x_1,\ldots,x_N\}\) & Dataset of \(N\) items. \\
\(H_i=\{h_{i,1},\ldots,h_{i,M_i}\}\) & The \(M_i\) human annotations observed for item \(x_i\). \\
\(f(x_i)\) & LLM annotation for item \(x_i\). \\
\(k,\;k_c\) & Panel size used by the test, with \(k\ge3\) and \(M_i\ge k\), and the panel size at which \(c\) is calibrated. \\
\(c\) & Margin constant; \(c=0.1\) by default. \\
\(\mathbf y_{i,h},\;\mathbf y_{i,m}\) & Human-only panel of \(k\) annotations for item \(x_i\), and the mixed panel formed by replacing one of them with \(f(x_i)\); stacking them over items gives the matrices \(\mathbf Y_h\) and \(\mathbf Y_m\). \\
\(R(\cdot)\) & Chance-corrected agreement score; higher values indicate more reliable panels. \\
\(R_h=R(\mathbf Y_h)\), \(R_m=R(\mathbf Y_m)\) & Human-only and mixed-panel agreement. \\
\(\Delta=R_m-R_h\) & Substitution gap. \\
\(\varepsilon=c\frac{k_c}{k}(1-R_h)\) & Adaptive non-inferiority margin. \\
\(\Delta_{\mathrm{adj}}=\Delta+\varepsilon\) & Adjusted substitution gap; the null hypothesis is \(\Delta_{\mathrm{adj}}\le0\). \\
\(B\) & Number of bootstrap iterations. \\
\(R_h^{(b)},\;R_m^{(b)}\) & Human-only and mixed agreement in bootstrap iteration \(b\); \(\varepsilon^{(b)}\), \(\widehat{\Delta}^{(b)}\) and \(\widehat{\Delta}^{(b)}_{\mathrm{adj}}\) are defined analogously. \\
\(q_\alpha^*\) & Empirical \(\alpha\)-quantile of the adjusted bootstrap gaps; the LLM passes when \(q_\alpha^*>0\). \\
\(H_i^g\) & The annotations of item \(x_i\) that come from group \(g\), used in group-wise testing. \\
\(\tilde{q}^{*}_{1-\alpha}\) & Empirical \((1-\alpha)\)-quantile of \(\widehat{\Delta}^{(b)}-\varepsilon^{(b)}\), used in the \textsc{Eq}-\anlt{}. \\
\(\Rh,\;\Rm\) & Mean human-only and mixed-panel agreement across bootstrap iterations; \(\Rm\) is used to compare judges. \\
\(L\) & Number of labels. \\
\(m,\;\eta_h,\;\eta_f\) & In the simulations: human annotators per item, and the noise levels of the humans and of the LLM. \\
\end{longtable}
}

\subsection{\alttest{} and STT}
\label{app:baselines}

In this section, we describe the two existing tests with which we compare the \anlt{}, the \alttest{}~\citep{calderon2025alternative} and the Statistical Turing Test (STT; \citealp{he2025can}), and how we implement them. Table~\ref{tab:test-comparison} summarizes the differences between the three tests.

To make the comparison fair, all tests receive the same input whenever their requirements allow it. In the simulations, we generate every dataset and every subsample of items once, and apply all tests to it. In the real-world experiments, all tests receive the same LLM annotations, and for each dataset, we use the items that satisfy the requirements of all the compared tests. Differences between the results therefore come from the tests and not from the data.

\begin{table}[t]
\small
\caption{Comparison of the three procedures evaluated in the paper. \cmark{} marks a property the test has; \xmark{} marks a limitation or requirement.}
\label{tab:test-comparison}
\setlength{\tabcolsep}{4pt}
\renewcommand{\arraystretch}{1.15}
\centering
\begin{tabular}{@{}p{2.55cm}p{3.2cm}p{3.3cm}p{3.5cm}@{}}
\toprule
& \textbf{\alttest{}} \citep{calderon2025alternative}
& \textbf{STT} \citep{he2025can}
& \textbf{\anlt{} (ours)} \\
\midrule
Question & can the LLM replace a human annotator? & is the LLM indistinguishable from a human annotator? & can the LLM replace a human annotator? \\
Hypothesis & non-inferiority, one-sided & equivalence, two-sided & non-inferiority, one-sided \\
Compared against & the excluded annotator, head to head & the all-human mean & the annotator pool \\
\midrule
Annotator identities & \xmark{} required & \xmark{} required & \cmark{} not required \\
Annotator structure & \xmark{} repeated annotators across items & \xmark{} two disjoint groups, same items & \cmark{} a single pool, no overlap needed \\
Annotators per item & $\ge 3$ & $\ge 4$ ($\ge 2$ per group) & $\ge 3$ \\
Recommended items & $50$--$100$ & $\ge 80$ & $\ge 100$ \\
\midrule
Agreement metric & task-specific: \textsc{acc}, $-$\textsc{rmse}, \textsc{sim} & Krippendorff's $\alpha$ & \cmark{} any chance-corrected IAA \\
Test & paired $t$-test per annotator & TOST (two one-sided $t$-tests) & bootstrap quantile of $\Delta+\varepsilon$ \\
Decision rule & $\omega \ge 0.5$ & both one-sided $p < \alpha$ & $q^{*}_{\alpha} > 0$ \\
Margin & \xmark{} fixed $\varepsilon \in \{0.1, 0.15, 0.2\}$, set by annotator type & \cmark{} $(\alpha_a - \alpha_b)\cdot\mathrm{frac}$, factor unspecified & \cmark{} $c\,(1 - R_h^{(b)})\,k_c/k$, per replicate, calibrated \\
Ranks several LLMs & \cmark{} $\rho$ (adv.\ probability) & \xmark{} & \cmark{} $\Rm$ \\
\bottomrule
\end{tabular}

\end{table}

\paragraph{\alttest{}.}
The \alttest{} compares the LLM with each human annotator separately. Each annotator is held out in turn, and the test examines whether the LLM is aligned with the remaining annotators at least as well as the held-out annotator, up to a margin $\varepsilon$. The tests of the different annotators are corrected for multiple comparisons, and the LLM passes when it wins against at least half of the annotators. We use the implementation of \citet{calderon2025alternative}.

The margin of the \alttest{} is chosen according to the type of the annotators, with $\varepsilon=0.10$ for crowd workers, $0.15$ for skilled annotators, and $0.20$ for experts. In the simulations, we therefore examine a range of margins and mark these three values. In real-world data, the test can be applied only when the identities of the annotators are available and every annotator labels enough items, for which we require at least $30$ of the evaluated items. We therefore apply it only to the datasets that satisfy these conditions.

\paragraph{Statistical Turing Test (STT).}
The STT examines whether the LLM is indistinguishable from a human annotator. Each annotator of an all-human panel is replaced by the LLM in turn, and the agreement of the resulting mixed panels is compared with the agreement of the all-human panel using a paired bootstrap. Since the STT is an equivalence test, the LLM passes only when both one-sided tests reject. We implement the test following \citet{he2025can}.

The margin of the STT is estimated from two disjoint groups of human annotators who label the same items, and it is multiplied by a scaling factor $\mathrm{frac}$. The original paper presents several values of $\mathrm{frac}$ and does not recommend one. We therefore examine a range of values in the simulations and use $\mathrm{frac}=0.5$ in the real-world experiments. The requirement of two groups of annotators holds in our simulations, but among our real-world datasets, it holds only for DICES-350. We therefore evaluate the STT in the simulations and on DICES-350.

\paragraph{\textsc{STT-Alt}.}
The STT tests equivalence, while the \alttest{} and the \anlt{} test non-inferiority, so their results cannot be compared directly. When the LLM is better than the humans, a non-inferiority test should accept it, but an equivalence test may correctly not accept it, since the LLM differs from the humans. We therefore introduce \textsc{STT-Alt}, which adapts the STT to the alternative annotator question.

\textsc{STT-Alt} is identical to the STT, with the same sampling, agreement score, bootstrap, and margin, but it applies only the lower one-sided test. It therefore tests whether the mixed panel is not worse than the all-human panel by more than the margin. Since both tests use the same bootstrap statistics, every LLM that passes the STT also passes \textsc{STT-Alt}.

\subsection{Simulations}
\label{app:impl-sim}

In this section, we provide the full details of the simulations. We start with the basic simulation of Section~\ref{sec:sim}, and then describe the additional settings, which vary the label distribution, the behavior of the annotators, and the level of human agreement. The simulations with groups of annotators are described in Appendix~\ref{app:persp-clusters}.

\paragraph{Basic simulation.}
Each simulated dataset contains $600$ items, generated as described in Section~\ref{sec:sim}. We examine $25$ conditions, one for each pair of noise levels $(\eta_h,\eta_f)$. In $10$ conditions the LLM is noisier than the humans (worse-model), in $10$ it is less noisy (better-model), and in $5$ the noise levels are equal. For each condition, we generate $10$ independent datasets and draw from each of them $10$ independent subsamples of $N$ items, without replacement. Subsamples from the same dataset may overlap, but no item is repeated within a subsample. This gives $100$ test decisions per condition and $2500$ decisions overall. Every run uses $B=500$ bootstrap iterations. Since $k=k_c$ in the default setting, the margin is $\varepsilon^{(b)}=c(1-R_h^{(b)})$, and when we vary $k$, we apply the full $k_c/k$ adjustment.

\paragraph{Generative families.}
In the basic simulation, we draw the label prior from a Dirichlet distribution and sample the noisy labels from this prior, which gives a skewed nominal distribution. We repeat the full noise grid with two additional distributions, which change both the gold labels and the errors.
\begin{itemize}\itemsep2pt
\item \textbf{Uniform.} The gold labels are uniform over the $L$ labels, and a noisy label is sampled uniformly from the remaining $L-1$ labels. This represents balanced nominal tasks.
\item \textbf{Gaussian.} The gold labels follow a discretized normal distribution, and a noisy label is $\mathrm{round}(y+\mathcal{N}(0,\sigma))$, reflected at the boundaries, where $\sigma$ is chosen to obtain the target error rate $\eta$. This represents ordinal tasks, in which errors fall on nearby labels.
\end{itemize}

\paragraph{Annotator behavior.}
In the basic simulation, all annotators err in the same way, and every annotator labels every item. We relax both assumptions.

\emph{Personal bias.} In practice, each annotator may have a personal bias and tend toward specific labels when making an error. We therefore give every human annotator $j$ a personal bias $d_j$, which is a distribution over the labels. When annotator $j$ errs, the label is sampled from $(\pi+\lambda d_j)/(1+\lambda)$ with $\lambda=2$, and not from $\pi$. The basic simulation is the special case $\lambda=0$. To generate the biases, we draw for each dataset a shared direction $b_0\sim\mathrm{Dirichlet}(\mathbf 1_L)$, and then draw $d_j\sim\mathrm{Dirichlet}(\kappa b_0)$ with $\kappa=4$, so the biases of different annotators are related but not identical. The LLM has no personal bias and errs according to $\pi$, as in the basic simulation.

\emph{Shared bias.} We also simulate a bias that is shared by different annotators. For each item, one label other than the gold label is chosen in advance as a distractor. When an annotator errs, the label is the distractor with probability $p=0.5$, and otherwise it is sampled from $\pi$, as in the basic simulation. We examine two cases. In the first case, both the humans and the LLM err toward the distractor, so their errors are correlated. In the second case, only the humans do.

\emph{Sparse annotations.} Finally, we simulate the setting for which the \anlt{} is designed, in which each annotator labels only a few items. Each annotator labels $t$ items, and we examine twenty values of $t$ from $1$ to $200$. Every item still receives $m=6$ labels from six different annotators, so the pool contains $1200/t$ annotators, from $1200$ at $t=1$ to $6$ at $t=200$. The annotators have a personal bias, as described above. At $t=200$, every annotator labels every item, which is the dense setting. For $t<200$, the \alttest{} and the STT cannot be applied, so we evaluate only the \anlt{}.

\paragraph{Low human agreement.}
To examine the test when human agreement is very low, we use the basic simulation with higher human noise levels, so that human IAA falls in three low ranges, $0.1$ to $0.2$, $0.05$ to $0.1$, and below $0.05$. We pair these humans with an LLM that carries no information about the items. We examine an LLM that assigns random labels, and an LLM that always assigns the most frequent label of the dataset.

\subsection{Datasets, Models, and Prompting}
\label{app:datasets}

\paragraph{Datasets.}
Table~\ref{tab:datasets-app} describes the $20$ datasets. For each dataset, it reports the task, the type of the labels, the number of labels $L$, the number of items with at least three human annotations $N$, the number of human annotations per item $M_i$, and the human-only agreement $\Rh$.

The datasets differ in their annotation setup, which is the last column of the table. In four datasets (Loneliness, SummEval, Flickr8k-Expert, and DICES-350), the same annotators label all the evaluated items. In nine datasets, the identities of the annotators are available, but different annotators label different items. For example, Measuring-Hate-Speech has $7{,}912$ annotators, who label about six of the evaluated items each, and in Persuasion, no annotator labels more than two of them. In the remaining seven datasets, only the labels of each item are available. The \anlt{} is applied in the same way in all three cases, since it uses only the $M_i$ annotations of each item.

\small
\setlength{\tabcolsep}{2.5pt}
\begin{longtable}{@{}p{5.25cm}p{3.05cm}llrrll@{}}
\caption{\textbf{Dataset descriptions.} The table reports the task, label type, number of labels, number of usable items, number of human annotations per item, and human-only agreement $\Rh$. Here, ord.\ means ordinal, nom.\ means nominal, $L$ is the number of labels, $N$ is the number of items with at least three labels, and $M_i$ is the number of human annotations for item $x_i$. Setup is the annotation setup, in which the same annotators label all items (same), different annotators label different items (varying), or the annotators are not identified (no IDs).}
\label{tab:datasets-app}\\
\toprule
\textbf{Dataset} & \textbf{Task} & \textbf{Type} & $\boldsymbol{L}$
& $\boldsymbol{N}$ & $\boldsymbol{M_i}$
& $\boldsymbol{\Rh}$ & \textbf{Setup} \\
\midrule
\endfirsthead

\multicolumn{8}{@{}l}{\emph{Table~\ref{tab:datasets-app}, continued}}\\
\toprule
\textbf{Dataset} & \textbf{Task} & \textbf{Type} & $\boldsymbol{L}$
& $\boldsymbol{N}$ & $\boldsymbol{M_i}$
& $\boldsymbol{\Rh}$ & \textbf{Setup} \\
\midrule
\endhead

\bottomrule
\endfoot

\multicolumn{8}{@{}l}{\textit{\textbf{\faFile*\ Text}}}\\
EPIC \citep{frenda2023epic}                                    & Irony detection          & ord. & 2  & 1500  & 5--8   & 0.20 & varying \\
DICES-350 \citep{aroyo2023dicesdatasetdiversityconversational} & Conversational safety    & ord. & 3  & 350   & 123    & 0.20 & same \\
ChaosNLI \citep{nie2020learncollectivehumanopinions}           & NLI entailment           & nom. & 3  & 1500  & 100    & 0.28 & no IDs \\
Measuring-Hate-Speech \citep{sachdeva2022measuring}            & Hate-speech severity     & ord. & 3  & 1500  & 4--815 & 0.53 & varying \\
Loneliness\footnote{\scriptsize\url{https://huggingface.co/datasets/yael-katsman/Loneliness-Causes-and-Intensity}} & Loneliness intensity & ord. & 5 & 497 & 3 & 0.90 & same \\
CEBaB \citep{abraham2022cebab}                                 & Restaurant star rating   & ord. & 5  & 711   & 3--4   & 0.72 & varying \\
MultiPref \citep{miranda2025hybridpreferenceslearningroute}    & Preference strength      & ord. & 5  & 1500  & 4      & 0.28 & varying \\
Scruples \citep{lourie2021scruples}                            & Moral judgment           & nom. & 5  & 1500  & 30     & 0.38 & no IDs \\
SummEval \citep{fabbri2021summeval}                            & Summary quality          & ord. & 5  & 1600  & 3      & 0.60 & same \\
10k-Prompts\footnote{\scriptsize\url{https://huggingface.co/datasets/data-is-better-together/10k_prompts_ranked}} & Prompt quality rank & ord. & 5 & 391 & 3--5 & 0.37 & varying \\
ConvAbuse \citep{curry2021convabusedataanalysisbenchmarks}     & Abuse severity           & ord. & 5  & 2225  & 3--8   & 0.72 & varying \\
Persuasion\footnote{\scriptsize\url{https://huggingface.co/datasets/Anthropic/persuasion}} & Persuasiveness & ord. & 7 & 1313 & 3 & 0.31 & varying \\
\midrule
\multicolumn{8}{@{}l}{\textit{\textbf{\faImage\ Image}}}\\
T2I-DPO \citep{cai2025diffusion}                               & Text-to-image pref. & nom. & 2  & 400   & 8      & 0.12 & no IDs \\
Vibe-Arena\footnote{\scriptsize\url{https://huggingface.co/datasets/datapointai/vibe-landing-page-arena}} & Landing-page pref. & nom. & 2 & 400 & 15 & 0.12 & no IDs \\
Flickr8k-Expert \citep{hodosh2013framing}                      & Caption quality          & ord. & 4  & 400   & 3      & 0.87 & same \\
FER+ \citep{barsoum2016training}                               & Facial expression        & nom. & 8  & 600   & 10     & 0.51 & no IDs \\
CIFAR-10H \citep{peterson2019human}                            & Image classification     & nom. & 10 & 600   & 30     & 0.88 & varying \\
\midrule
\multicolumn{8}{@{}l}{\textit{\textbf{\faVolumeUp\ Audio}}}\\
TTS-Preference\footnote{\scriptsize\url{https://huggingface.co/datasets/datapointai/tts-human-preferences-medium}} & TTS naturalness pref. & nom. & 2 & 1000 & 15 & 0.01 & no IDs \\
SOMOS \citep{maniati2022somos}                                 & Speech quality (MOS) & ord. & 5  & 20100 & 16--27 & 0.05 & varying \\
\midrule
\multicolumn{8}{@{}l}{\textit{\textbf{\faVideo\ Video}}}\\
T2V-Preference \citep{he2509videoscore2}                       & Text-to-video pref. & nom. & 2  & 300   & 74--90 & 0.04 & no IDs \\
\end{longtable}

\paragraph{Models.}
We evaluate $11$ open-weight LLMs, which are Qwen3-8B and Qwen3-14B \citep{yang2025qwen3}, Llama-3.2-3B, Llama-3.1-8B, and Llama-3.2-11B-Vision \citep{grattafiori2024llama}, Gemma-3-4B and Gemma-3-12B \citep{team2025gemma}, Qwen2.5-VL-3B and Qwen2.5-VL-7B \citep{bai2025qwen25vltechnicalreport}, Qwen2.5-Omni-3B \citep{xu2025qwen25omnitechnicalreport}, and Qwen2-Audio-7B \citep{chu2024qwen2audiotechnicalreport}. We also evaluate seven proprietary LLMs, which are Gemini-3.1 Pro, Gemini-3.7 Flash, and Gemini-3.5 Flash-Lite \citep{gemini2026models}, GPT-5.6 Terra and GPT-5.6 Luna \citep{openai2026gpt56}, GPT-Audio-1.5 \citep{openai2026gptaudio}, and GPT-Audio-Mini \citep{openai2026gptaudiomini}. For video, the two Qwen2.5-VL models receive frames that are sampled from the video, so their results apply to this full pipeline and not only to the model.

\paragraph{Prompting.}
Every LLM receives the definition of the task and the annotation guidelines of the dataset. In few-shot prompting, we add five examples that are sampled at random. In chain-of-thought prompting, we use a mild reasoning effort. The ensemble is the majority vote over the labels of the zero-shot, few-shot, and chain-of-thought prompts. In Appendix~\ref{app:real-prompting}, we examine how the results change with the number of few-shot examples and with the reasoning effort.

\section{Annotator Perspectives}
\label{app:perspectives}

The \anlt{} asks whether an LLM preserves the signal that the annotators share. This is the relevant question when human labels are used to train, evaluate, or judge models. In these uses, a model is trained toward, or scored against, one target per item, so what matters is the judgment on which the annotators converge. In some studies, however, annotators hold distinct perspectives, for example across demographic groups, and the study aims to preserve these perspectives~\citep{aroyo2015truth,basile2021we,uma2021learning,plank2022problem}.

The standard test is not designed for this second goal, for two reasons. First, the replaced annotation is drawn from the whole pool of an item's annotations. The test therefore measures the agreement of the pooled annotations, to which a larger group contributes more annotations than a smaller one. Second, the test is one-sided, so it accepts an LLM that raises agreement, although such an LLM may remove the variation that the study aims to capture.

In this section, we address both points. In Appendix~\ref{app:impl-altra}, we start by defining two variants of the test, group-wise testing and an equivalence test. Then, in Appendix~\ref{app:persp-clusters}, we use simulations with groups of annotators to show how the standard test behaves when perspectives differ. In Appendix~\ref{app:persp-dices}, we apply group-wise testing to a real dataset, and finally, in Appendix~\ref{app:persp-eq-sim}, we evaluate the equivalence variant in simulation. We leave for future work an extension of the test that considers the distribution of the LLM annotations for each item, and examines whether it is aligned with the distribution of the human annotations.

\subsection{Variants of the Test}
\label{app:impl-altra}

\paragraph{Group-wise testing.}
Suppose that every annotation carries a group label, such as a demographic attribute of its annotator. For a group $g$, let $H_i^g\subseteq H_i$ denote the annotations of item $x_i$ that come from this group. The group-wise test applies the \anlt{} without any change to the items with at least $k$ such annotations, and it draws the panels only from $H_i^g$. The human-only agreement, the margin, and the decision statistic are then all computed within the group, so the test compares the LLM with a typical annotator of that group.

Researchers can require the LLM to pass in every group, or only in the groups of interest. Since this involves several tests, researchers should account for multiple comparisons when drawing conclusions about individual groups. In addition, the chance that an LLM passes in every group decreases as the number of groups grows. Each tested group also needs enough items with at least $k$ annotations from that group. In Section~\ref{sec:sim}, we show that about $100$ items are enough for a high better-model pass rate.

\paragraph{Equivalence.}
The \anlt{} is a non-inferiority test, so an LLM that increases the agreement of the panel passes it. In some settings, we instead want to test whether the LLM is indistinguishable from a human annotator. This is the case when the annotations should reflect human judgment and its variation, since labels that are more consistent than the human ones misrepresent the annotators. In such settings, an LLM that increases agreement by more than the margin should fail the test, like an LLM that decreases it.

We therefore define the \textsc{Eq}-\anlt{}, which tests whether the substitution gap is within the margin in both directions:
\begin{equation}
    H_0:\; |\Delta| \ge \varepsilon
    \qquad \text{\textbf{vs.}} \qquad
    H_1:\; |\Delta| < \varepsilon .
\end{equation}
The test uses the same bootstrap iterations as the \anlt{}, and the LLM passes when
\begin{equation}
    q^*_\alpha > 0
    \quad \text{and} \quad
    \tilde{q}^{*}_{1-\alpha} < 0 ,
\end{equation}
where $\tilde{q}^{*}_{1-\alpha}$ is the $(1-\alpha)$-quantile of $\widehat{\Delta}^{(b)}-\varepsilon^{(b)}$. The first condition is the non-inferiority test of Section~\ref{subsec:test}. The second condition tests that the mixed-panel agreement does not exceed the human-only agreement by more than the margin. Together, they form the Two One-Sided Tests procedure (TOST; \citealp{rogers1993using}), applied to bootstrap quantiles.

Two points should be noted when using the \textsc{Eq}-\anlt{}. First, the constant $c$ must be recalibrated. We calibrated $c=0.1$ for the non-inferiority test, and with the second condition, this value reduces the better-model pass rate from $0.984$ to $0.030$ (see Appendix~\ref{app:persp-eq-sim}). Researchers should choose $c$ according to the tolerance that their setting requires; in Appendix~\ref{app:persp-eq-sim}, we use $c=0.125$. Second, a failure has a different meaning. We cannot conclude that the LLM is worse than the humans, only that it is distinguishable from them, possibly because it is better.

\subsection{Simulations with Annotator Groups}
\label{app:persp-clusters}

\paragraph{Setup.}
We simulate a task in which the annotators form two groups with different perspectives, a larger group and a smaller group. Each group has its own label prior and its own gold label for every item, and the annotators of a group are noisy around the gold label of their group, as in Section~\ref{sec:sim}. We vary the sizes of the two groups from $9{+}1$ to $5{+}5$ annotators. The simulated LLM follows one of four profiles. It labels like an annotator of the larger group, like an annotator of the smaller group, like an equal mixture of the two, or like a mixture in proportion to the group sizes. Since no profile is correct by construction, we report the pass rate of each profile, with $c=0.1$ and $N=200$. We also repeat the simulation when every human annotator has a personal bias $b_j\sim\mathrm{Dirichlet}(\mathbf 1_L)$ and errs according to $(\pi_g+b_j)/2$, where $\pi_g$ is the label prior of the annotator's group.

\begin{figure}[!tb]
  \centering
  \includegraphics[trim=0 24bp 0 0,clip,width=0.72\textwidth]{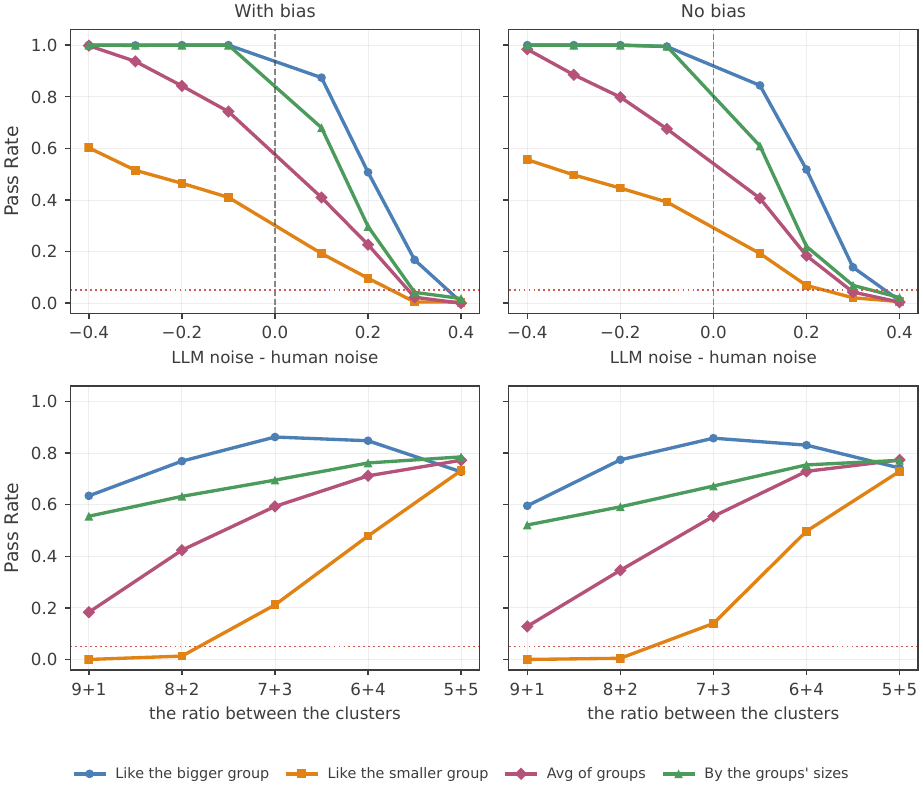}\\[2pt]
  \includegraphics[trim=0 0 0 356bp,clip,width=\textwidth]{figures/sim_final_clusters.pdf}
 \caption{Simulations with two groups of annotators, with (left) and without (right) a personal bias of the annotators, at $c\!=\!0.1$ and $N\!=\!200$. Top: the pass rate of each LLM profile against the noise difference $\eta_f-\eta_h$, averaged over all group sizes from $9{+}1$ to $5{+}5$. Bottom: the pass rate against the group sizes, averaged over every noise pair with $\eta_f\neq\eta_h$. As shown, the test favors an LLM that labels like the larger group.}
  \label{fig:app-clusters}
\end{figure}

\paragraph{The test behaves as expected for a shared signal.}
Figure~\ref{fig:app-clusters} presents the results. When annotators form groups, the signal that the pool shares is mainly the judgment of the larger group. Accordingly, an LLM that follows the larger group passes the test at high rates, while an LLM that follows the smaller group rarely passes when the larger group dominates the pool, even when it is less noisy than the humans. Such an LLM disagrees with most of the annotators, so replacing a human annotation with its annotation reduces the agreement of the pool. As the two groups approach equal size, neither group dominates the pool, and the gap between the profiles closes.

The two mixtures lead to the same conclusion. An LLM that mixes the groups in proportion to their sizes behaves like an annotator drawn at random from the pool, and its pass rates are close to those of the LLM that follows the larger group. The equal mixture represents the smaller group more than the pool does, and it passes less often when the groups are unequal.

For every profile, the pass rate decreases as the LLM becomes noisier, so the test remains sensitive to the quality of the LLM. However, noise alone does not determine the result. An LLM that follows the larger group can pass even when it is somewhat noisier than the humans, since it agrees with most annotators more than a randomly drawn annotator does. Finally, the results with and without a personal bias are nearly identical.

\paragraph{When perspectives should be represented.}
The same behavior is not desired when the goal is to represent each perspective. In this simulation, an LLM that follows the smaller group faithfully rarely passes, and an LLM that ignores the smaller group passes. For this goal, we therefore recommend the variants of Appendix~\ref{app:impl-altra}. In Appendix~\ref{app:persp-dices}, we apply group-wise testing to a real dataset, and in Appendix~\ref{app:persp-eq-sim}, we evaluate the equivalence variant in simulation.

\subsection{Group-wise Testing on DICES-350}
\label{app:persp-dices}

DICES-350 reports the demographics of its annotators. We use them to define $13$ groups, according to age, education, gender, and race (Table~\ref{tab:app-dices-groups}). Every annotator labels every item, so each group has between $7$ and $75$ annotations per item. The human agreement is similar in all groups, between $0.14$ and $0.25$, compared with $0.20$ for the whole pool. Every LLM uses its best prompting technique.

When panels are drawn from the whole pool, four LLMs pass the \anlt{}. When panels are drawn within each group and an LLM must pass in all $13$ groups, only G3-12b passes. Gem-P and GPT-T fail only in the Latinx group, and GPT-L fails in three groups, including the Latinx group. In all these cases, the decision statistic is only slightly below zero. Group-wise testing therefore exposes a misalignment with one group, which pooled testing averages out. The choice between the two depends on whether the goal is to match the pooled signal or to represent each group.

Selecting panels with diverse labels cannot replace group labels, because label diversity alone does not distinguish a minority perspective from annotation noise.

\begin{table}[!tb]
\centering\small
\caption{Group-wise testing on DICES-350. For each group, the table reports the number of annotators, the human-only agreement $\Rh$, and whether each of the four LLMs that pass with pooled panels also passes within the group.}
\label{tab:app-dices-groups}
\setlength{\tabcolsep}{5pt}
\begin{tabular}{@{}llrccccc@{}}
\toprule
\textbf{Attribute} & \textbf{Group} & \textbf{Annotators} & $\boldsymbol{\Rh}$ & \textbf{G3-12b} & \textbf{Gem-P} & \textbf{GPT-T} & \textbf{GPT-L} \\
\midrule
All & Whole pool & $123$ & $0.20$ & \cmark & \cmark & \cmark & \cmark \\
\midrule
Age & Gen X+ & $31$ & $0.17$ & \cmark & \cmark & \cmark & \xmark \\
 & Gen Z & $56$ & $0.19$ & \cmark & \cmark & \cmark & \cmark \\
 & Millennial & $36$ & $0.22$ & \cmark & \cmark & \cmark & \cmark \\
\addlinespace[2pt]
Education & College degree or higher & $75$ & $0.19$ & \cmark & \cmark & \cmark & \cmark \\
 & High school or below & $41$ & $0.20$ & \cmark & \cmark & \cmark & \cmark \\
 & Other & $7$ & $0.19$ & \cmark & \cmark & \cmark & \cmark \\
\addlinespace[2pt]
Gender & Man & $61$ & $0.21$ & \cmark & \cmark & \cmark & \cmark \\
 & Woman & $62$ & $0.18$ & \cmark & \cmark & \cmark & \cmark \\
\addlinespace[2pt]
Race & Asian & $26$ & $0.17$ & \cmark & \cmark & \cmark & \cmark \\
 & Black & $29$ & $0.24$ & \cmark & \cmark & \cmark & \xmark \\
 & Latinx & $22$ & $0.25$ & \cmark & \xmark & \xmark & \xmark \\
 & Multiracial & $16$ & $0.18$ & \cmark & \cmark & \cmark & \cmark \\
 & White & $30$ & $0.14$ & \cmark & \cmark & \cmark & \cmark \\
\bottomrule
\end{tabular}
\end{table}

\subsection{Equivalence in Simulation}
\label{app:persp-eq-sim}

We evaluate the \textsc{Eq}-\anlt{} with the simulation setup of Section~\ref{sec:sim}. We first replace only the decision rule of the \anlt{} with the equivalence rule of Appendix~\ref{app:impl-altra} and keep everything else fixed. At $c=0.1$, the better-model pass rate drops from $0.984$ to $0.030$. This is expected, since an equivalence test should not accept an LLM that differs from the humans, also when it is better. The better-model pass rate is therefore not a suitable measure for an equivalence test. Instead, we report the pass rate when equal, over the five conditions with $\eta_f=\eta_h$, and the pass rate when not equal, over the twelve conditions with $|\eta_f-\eta_h|\ge0.2$. We also note that the percentile form that we use and equivalent $t$- and $z$-based implementations disagree on only $0.68\%$ of the $2500$ runs.

\begin{figure}[t]
  \centering
  \includegraphics[width=0.45\textwidth]{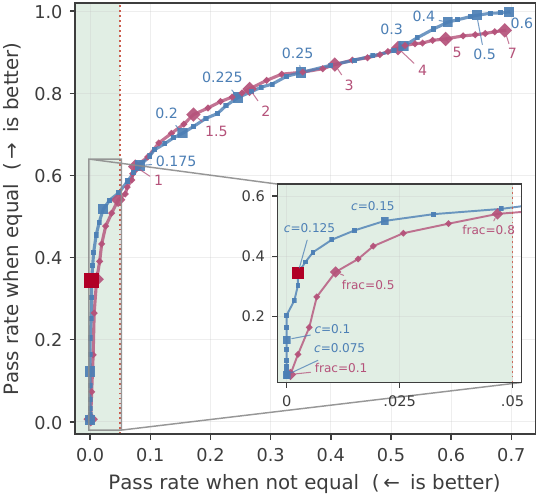}\hfill
  \includegraphics[width=0.51\textwidth]{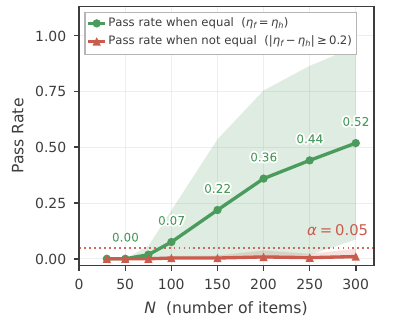}

  \vspace{0.4em}
  \includegraphics[width=0.8\textwidth]{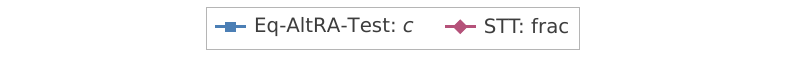}

  \caption{\textit{Left:} Comparison of the \textsc{Eq}-\anlt{} (see Appendix~\ref{app:impl-altra}) against the STT for the equivalence question. Both apply two one-sided tests: the \textsc{Eq}-\anlt{} to bootstrap quantiles, and the STT as $t$-tests. Simulations use $N\!=\!200$ items and $k\!=\!3$, while varying $c$ for the \textsc{Eq}-\anlt{} and $\mathrm{frac}$ for the STT. The X-axis is the pass rate when not equal, over conditions with $|\eta_f-\eta_h|\ge0.2$, and the Y-axis is the pass rate when equal, over conditions with $\eta_f\!=\!\eta_h$. \textit{Right:} Using the calibration in the left panel, we set $c\!=\!0.125$ and vary the number of items $N$.}
  \label{fig:app-tost}
\end{figure}

Figure~\ref{fig:app-tost} (left) compares the \textsc{Eq}-\anlt{} with the STT using these measures. Within the valid region, the \textsc{Eq}-\anlt{} attains higher pass rates when equal than the STT, and the two tests are nearly indistinguishable elsewhere. Figure~\ref{fig:app-tost} (right) fixes $c=0.125$ and varies $N$. The pass rate when not equal stays below $\alpha=0.05$ at every sample size, while the pass rate when equal increases from $0.07$ at $N=100$ to $0.36$ at $N=200$ and $0.52$ at $N=300$. The equivalence variant therefore requires substantially more items than the non-inferiority test.

\FloatBarrier

\section{Theoretical Guarantees}
\label{app:theoretical-guarantees}

This section establishes two properties of the \anlt{}. First, the substitution gap decreases at rate $1/k$ as the panel size grows, supporting the panel-size adjustment in the adaptive margin. Second, under standard regularity conditions, the lower bootstrap quantile consistently estimates the adjusted substitution gap and yields a consistent test of asymptotic level $\alpha$.

\subsection{Population Quantities}
\label{sec:agreement-notation}

Let $Z_i=(X_i,H_i,F_i)$, $i=1,\ldots,N$, denote the observed item clusters, where $H_i$ is the multiset of available human labels and $F_i=f(X_i)$ is the label produced by a fixed model and prompt. We assume that every item has at least $k\ge3$ human labels.

For each item, a panel draw $U$ samples $k$ distinct human annotations uniformly without replacement and selects one of them uniformly for replacement by $F_i$. The same draw therefore produces a paired human-only panel containing $k$ human labels and a mixed panel containing $k-1$ human labels and one model label.

Let $\mathcal L$ be a finite label space and let $\delta(a,b)$ be a fixed, bounded, symmetric disagreement function satisfying $\delta(a,a)=0$ and $\delta(a,b)=\delta(b,a)$. 
For panel type $t\in\{h,m\}$, where $h$ stands for human-only and $m$ stands for mixed, let $Y_{t,1},\ldots,Y_{t,k}$ denote its labels and define
$$
d_t(Z,U)
=
\binom{k}{2}^{-1}
\sum_{r<s}\delta(Y_{t,r},Y_{t,s}),
\qquad
p_t(a;Z,U)
=
\frac{1}{k}\sum_{r=1}^k\mathbf 1\{Y_{t,r}=a\}.
$$

Thus, $d_t(Z,U)$ is the average pairwise disagreement within the panel and $p_t(a;Z,U)$ is the proportion of labels assigned to category $a$.

The corresponding population quantities are
$$
D_{o,t}=\mathbb E[d_t(Z,U)],
\qquad
P_t(a)=\mathbb E[p_t(a;Z,U)],
\qquad
D_{e,t}
=
\sum_{a,b\in\mathcal L}
P_t(a)P_t(b)\delta(a,b),
\qquad
R_t=1-\frac{D_{o,t}}{D_{e,t}},
$$
where the expectations are over both the item and panel draws. 

The population substitution gap is
$$
\Delta=R_m-R_h.
$$
When panel size varies, we write $R_{m,k}$ and $\Delta_k$ to denote the panel size.

For $N$ sampled panels, the empirical agreement coefficient is obtained by replacing $D_{o,t}$ and $P_t$ with their pooled sample averages and applying the usual finite-sample correction to $D_{e,t}$. This is Krippendorff's coefficient for complete panels of equal size and the specified disagreement function~\citep{krippendorff2011computing}.

For a fixed panel size $k$, define
$$
\gamma=c\frac{k_c}{k},
\qquad
\varepsilon=\gamma(1-R_h),
\qquad
\Delta_{\mathrm{adj}}=\Delta+\varepsilon.
$$
The non-inferiority hypotheses are
$$
H_0:\Delta_{\mathrm{adj}}\le0
\qquad\text{versus}\qquad
H_1:\Delta_{\mathrm{adj}}>0.
$$

In each bootstrap replicate, $N$ items are sampled with replacement from the observed dataset. A fresh panel draw is generated independently for every sampled occurrence, and the corresponding human-only and mixed panels are constructed jointly. The bootstrap statistic is
$$
\widehat\Delta_{\mathrm{adj}}^{(b)}
=
\widehat R_m^{(b)}
-
\widehat R_h^{(b)}
+
\gamma\bigl(1-\widehat R_h^{(b)}\bigr).
$$
Thus, both agreement coefficients and the adaptive margin are re-estimated within every replicate.

Let $B_N$ denote the number of bootstrap replicates and define
$$
q_\alpha^*
=
Q_\alpha\!\left(
\left\{
\widehat\Delta_{\mathrm{adj}}^{(b)}
\right\}_{b=1}^{B_N}
\right),
$$

where $Q_\alpha$ is the empirical lower $\alpha$-quantile. The LLM passes when $q_\alpha^*>0$.

\subsection{Panel-Size Scaling}
\label{sec:panel-size-scaling}

Let $P_F(a)=\mathbb P(F=a)$ denote the marginal distribution of the model labels. We further define the expected within-item model--human disagreement
$$
D_{FH}
=
\mathbb E\!\left[
\frac{1}{|H|}
\sum_{y\in H}\delta(F,y)
\right],
$$
and the chance disagreements
$$
G_{HF}
=
\sum_{a,b\in\mathcal L}
P_h(a)P_F(b)\delta(a,b),
\qquad
G_F
=
\sum_{a,b\in\mathcal L}
P_F(a)P_F(b)\delta(a,b).
$$

Uniform sampling ensures that the marginal distribution of a selected human label and the joint distribution of a selected pair of distinct human labels do not depend on $k$. Consequently, $P_h$, $D_{o,h}$, $D_{e,h}$, and $R_h$ remain fixed across feasible panel sizes.

A mixed panel contains $\binom{k-1}{2}$ human--human pairs and $k-1$ model--human pairs. Its observed disagreement is therefore
\begin{equation}
D_{o,m,k}
=
D_{o,h}
+
\frac{2}{k}(D_{FH}-D_{o,h}).
\label{eq:mixed-observed}
\end{equation}
Its marginal label distribution is
$$
P_{m,k}(a)
=
\frac{k-1}{k}P_h(a)+\frac{1}{k}P_F(a),
$$

which yields
\begin{equation}
D_{e,m,k}
=
D_{e,h}
+
\frac{2}{k}(G_{HF}-D_{e,h})
+
\frac{1}{k^2}(D_{e,h}-2G_{HF}+G_F).
\label{eq:mixed-chance}
\end{equation}
Because every term in the equivalent expression
$$
D_{e,m,k}
=
\frac{(k-1)^2D_{e,h}+2(k-1)G_{HF}+G_F}{k^2}
$$
is nonnegative,
$$
D_{e,m,k}
\ge
\left(\frac{k-1}{k}\right)^2D_{e,h}.
$$
Thus, $D_{e,h}>0$ also guarantees $D_{e,m,k}>0$.

\begin{proposition}[First-order panel-size scaling]
\label{prop:panel-size-scaling}
Suppose that $D_{o,h}$, $D_{e,h}$, $D_{FH}$, $G_{HF}$, and $G_F$ remain fixed as $k$ varies and that $D_{e,h}>0$. Then
\begin{equation}
\Delta_k
=
\frac{2(D_{o,h}G_{HF}-D_{FH}D_{e,h})}
{D_{e,h}^{2}}
\frac{1}{k}
+
O(k^{-2}).
\label{eq:panel-size-expansion}
\end{equation}
Hence, $\Delta_k=O(k^{-1})$. If the human and model marginal label distributions coincide, then
\begin{equation}
\Delta_k
=
\frac{2(D_{o,h}-D_{FH})}{D_{e,h}}\frac{1}{k}
\label{eq:panel-size-exact}
\end{equation}
holds exactly for every admissible panel size $k$.
\end{proposition}

\begin{proof}
Equations \ref{eq:mixed-observed} and \ref{eq:mixed-chance} imply that the difference in observed disagreements between the mixed and human-only panels is
$$
D_{o,m,k}-D_{o,h}
=
\frac{2}{k}(D_{FH}-D_{o,h})
$$
and the expected difference in disagreements is
$$
D_{e,m,k}-D_{e,h}
=
\frac{2}{k}(G_{HF}-D_{e,h})+O(k^{-2}).
$$
Expanding
$$
\Delta_k=R_{m,k}-R_h
=
-\frac{D_{o,m,k}}{D_{e,m,k}}
+\frac{D_{o,h}}{D_{e,h}}
$$
around $(D_{o,h},D_{e,h})$ gives
$$
\Delta_k
=
-\frac{D_{o,m,k}-D_{o,h}}{D_{e,h}}
+
\frac{D_{o,h}}{D_{e,h}^{\,2}}
(D_{e,m,k}-D_{e,h})
+
O(k^{-2}).
$$

Substituting the two preceding expressions and collecting the terms of order $1/k$ proves Equation~\ref{eq:panel-size-expansion}.

If $P_F=P_h$, then $G_{HF}=G_F=D_{e,h}$, and Equation~\ref{eq:mixed-chance} gives $D_{e,m,k}=D_{e,h}$. Substitution of Equation~\ref{eq:mixed-observed} then yields Equation~\ref{eq:panel-size-exact}.
\end{proof}

Proposition~\ref{prop:panel-size-scaling} shows that replacing one member of a larger panel produces a proportionally smaller change in agreement. This motivates scaling the margin at the same rate. For a reference panel size $k_c\ge3$, we therefore define
$$
\varepsilon_k
=
c(1-R_h)\frac{k_c}{k}.
$$
This adjustment is exact when the human and model label marginals coincide and provides a first-order normalization otherwise. The proposition concerns variation in $k$ across a family of admissible panel constructions. The asymptotic result below instead fixes $k$ and lets the number of items $N$ increase.

\subsection{Consistency and Asymptotic level $\alpha$}
\label{sec:consistency-validity}

The proof relies on a first-order expansion of the agreement coefficient. For panel type $t$, define its influence contribution as
\begin{equation}
\begin{aligned}
\ell_t(Z,U)
={}&
-\frac{d_t(Z,U)-D_{o,t}}{D_{e,t}}
+
\frac{2D_{o,t}}{D_{e,t}^{2}}
\sum_{a,b\in\mathcal L}
P_t(b)\delta(a,b)
\bigl[p_t(a;Z,U)-P_t(a)\bigr].
\end{aligned}
\label{eq:panel-contribution}
\end{equation}
The first term represents the panel's contribution to observed disagreement, and the second represents its contribution through the marginal label distribution and chance disagreement. Equation~\ref{eq:panel-contribution} is the influence function obtained by differentiating $R_t=1-D_{o,t}/D_{e,t}$ with respect to the distribution of panels.

The adjusted substitution gap has influence contribution
$$
\ell(Z,U)
=
\ell_m(Z,U)-(1+\gamma)\ell_h(Z,U).
$$
Let
$$
\bar\ell(Z)=\mathbb E[\ell(Z,U)\mid Z],
\qquad
\sigma^2=\operatorname{Var}[\bar\ell(Z)],
\qquad
\tau^2=\operatorname{Var}[\ell(Z,U)].
$$

\begin{lemma}[Variance contributed by panel selection]
\label{lem:panel-variance}
If $D_{e,h}>0$, then $\ell$ and $\bar\ell$ are bounded and have mean zero. Moreover,
\begin{equation}
\tau^2
=
\sigma^2+
\mathbb E\left[
\operatorname{Var}[\ell(Z,U)\mid Z]
\right]
\ge\sigma^2.
\label{eq:variance-order}
\end{equation}
\end{lemma}

\begin{proof}
Boundedness follows from the bounded disagreement function, the finite label space, and the positivity of $D_{e,h}$ and $D_{e,m}$. The centered terms in Equation~\ref{eq:panel-contribution} have expectation zero, implying $\mathbb E[\ell]=\mathbb E[\bar\ell]=0$. Equation~\ref{eq:variance-order} follows directly from the law of total variance conditional on $Z$. Because the human-only and mixed contributions use the same item and panel draw, their paired dependence is retained in $\ell$.
\end{proof}

The distinction between $\sigma^2$ and $\tau^2$ is important. The sampling variance $\sigma^2$ reflects variation across item clusters after averaging over the possible panel draws for each item. The bootstrap variance $\tau^2$ additionally includes the variation introduced by drawing a fresh panel for every bootstrap occurrence. The latter is therefore at least as large as the former, making the resulting lower bound potentially conservative.

\begin{theorem}[Consistency and Asymptotic level $\alpha$]
\label{thm:consistency-validity}
Consider a fixed population distribution in which the item clusters $Z_1,\ldots,Z_N$ are independent and identically distributed. Assume that the label space, bounded disagreement function, and panel size $k\ge3$ remain fixed as $N\to\infty$, and that every item has at least $k$ human annotations. Suppose that $D_{e,h}\ge d_0$ for some constant $d_0>0$, while the model, prompt, $c$, $k_c$, and significance level $0<\alpha<1/2$ are fixed independently of the test sample. Assume further that $\sigma^2>0$, that the bootstrap replicates are conditionally independent and use fresh independent panel draws, and that $B_N\to\infty$.

Let $\Phi$ denote the standard normal cumulative distribution function, let $z_\alpha=\Phi^{-1}(\alpha)$, and let $\sigma$ and $\tau$ be the positive square roots of the variances defined above. Then
$$
q_\alpha^*\xrightarrow{p}\Delta_{\mathrm{adj}},
$$
and
$$
\mathbb P\!\left\{
\Delta_{\mathrm{adj}}\ge q_\alpha^*
\right\}
\longrightarrow
\Phi\!\left(
-z_\alpha\frac{\tau}{\sigma}
\right)
\ge1-\alpha.
$$
Consequently,
$$
\begin{aligned}
\Delta_{\mathrm{adj}}\le0
&\quad\Longrightarrow\quad
\limsup_{N\to\infty}
\mathbb P\{q_\alpha^*>0\}\le\alpha,\\
\Delta_{\mathrm{adj}}<0
&\quad\Longrightarrow\quad
\mathbb P\{q_\alpha^*>0\}\longrightarrow0,\\
\Delta_{\mathrm{adj}}>0
&\quad\Longrightarrow\quad
\mathbb P\{q_\alpha^*>0\}\longrightarrow1.
\end{aligned}
$$

All probabilities account jointly for item sampling and bootstrap randomization.
\end{theorem}

\begin{proof}
For each observed item, first average its panel disagreement and label proportions over the conditional distribution of $U$. Substituting these conditional averages into the pooled agreement formula defines an item-level estimator $\widehat\Delta_{\mathrm{adj},N}$. This quantity is used only in the proof; the test does not need to compute it.

Because its inputs are averages of bounded i.i.d.\ item-level quantities, the law of large numbers gives
$$
\widehat\Delta_{\mathrm{adj},N}
\xrightarrow{p}
\Delta_{\mathrm{adj}}.
$$

The agreement coefficient is a smooth function of these averages whenever its chance-disagreement denominator is bounded away from zero. The multivariate delta method and Equation~\ref{eq:panel-contribution} therefore yield
\begin{equation}
\sqrt N
\bigl(
\widehat\Delta_{\mathrm{adj},N}
-
\Delta_{\mathrm{adj}}
\bigr)
=
\frac{1}{\sqrt N}
\sum_{i=1}^N\bar\ell(Z_i)
+
o_p(1)
\Longrightarrow
\mathcal N(0,\sigma^2).
\label{eq:sampling-clt}
\end{equation}
The finite-sample correction in the empirical chance disagreement contributes only $O(N^{-1})$ and therefore does not affect this limit.

For a generic bootstrap replicate, let $I_j$ be the item index sampled at occurrence $j$ and let $U_j$ be its fresh panel draw. Conditional on the observed items, the pairs $(Z_{I_j},U_j)$ are i.i.d. Marking with a star the conditioning on the observed items and applying the same delta-method expansion gives
\begin{equation}
\sqrt N
\bigl(
\widehat\Delta_{\mathrm{adj}}^*
-
\widehat\Delta_{\mathrm{adj},N}
\bigr)
\Longrightarrow^*
\mathcal N(0,\tau^2)
\quad\text{in probability}.
\label{eq:bootstrap-clt}
\end{equation}
Formally, the bootstrap expansion first uses the empirical, plug-in version of Equation~\ref{eq:panel-contribution}. Its coefficients converge in probability to their population counterparts because the empirical disagreements and label proportions are consistent. Boundedness then makes the difference between the empirical and population influence contributions $o_{p^*}(1)$ on the scale in Equation~\ref{eq:bootstrap-clt}.

The conditional distribution in Equation~\ref{eq:bootstrap-clt} has asymptotic lower $\alpha$-quantile $\tau z_\alpha$. Because $B_N\to\infty$, the empirical bootstrap quantile converges to the corresponding conditional quantile. Hence,
\begin{equation}
q_\alpha^*
=
\widehat\Delta_{\mathrm{adj},N}
+
\frac{\tau z_\alpha}{\sqrt N}
+
o_p(N^{-1/2}).
\label{eq:quantile-expansion}
\end{equation}
Consistency of $q_\alpha^*$ follows immediately.

Combining Equations \ref{eq:sampling-clt} and \ref{eq:quantile-expansion} yields
$$
\sqrt N
\bigl(
q_\alpha^*-\Delta_{\mathrm{adj}}
\bigr)
\Longrightarrow
\mathcal N(\tau z_\alpha,\sigma^2).
$$
Therefore,
$$
\mathbb P\!\left\{
\Delta_{\mathrm{adj}}\ge q_\alpha^*
\right\}
\longrightarrow
\Phi\!\left(
-z_\alpha\frac{\tau}{\sigma}
\right).
$$

Since $z_\alpha<0$ and $\tau\ge\sigma$ by Lemma~\ref{lem:panel-variance}, this limit is at least $1-\alpha$.

Under $H_0$, the event $\{q_\alpha^*>0\}$ is contained in $\{q_\alpha^*>\Delta_{\mathrm{adj}}\}$, so the asymptotic false-pass probability is at most $\alpha$. If $\Delta_{\mathrm{adj}}<0$ or $\Delta_{\mathrm{adj}}>0$, the remaining conclusions follow from $q_\alpha^*\xrightarrow{p}\Delta_{\mathrm{adj}}$.
\end{proof}


\section{Simulations: Additional Results}
\label{app:sim-results}

In this section, we present additional simulation results. All results use $c=0.1$ unless stated otherwise. In Appendix~\ref{app:sim-baselines}, we start by repeating the comparison with the baselines from Section~\ref{sec:sim} in other settings, with other distributions of the labels and with other numbers of items, and we show how we calibrated $c$. Then, in Appendix~\ref{app:sim-params}, we examine how the parameters of the test and of the task affect the results. Finally, in Appendix~\ref{app:sim-behavior}, we examine other behaviors of the annotators, and the case in which human agreement is very low.

\FloatBarrier
\subsection{Comparison with the Baselines and Calibration}
\label{app:sim-baselines}

\paragraph{Other generative families.}
In Section~\ref{sec:sim}, the labels follow a skewed distribution, in which the label prior is drawn from a Dirichlet distribution. We examine whether the comparison with the baselines also holds for two other distributions, which we describe in Appendix~\ref{app:impl-sim}. The first is a uniform distribution, which represents balanced nominal tasks. The second is a Gaussian distribution, which represents ordinal tasks.

\begin{figure}[!htbp]
  \centering
  \begin{minipage}[c]{0.31\textwidth}
    \begin{subfigure}[t]{\textwidth}
      \centering
      \includegraphics[width=\textwidth]{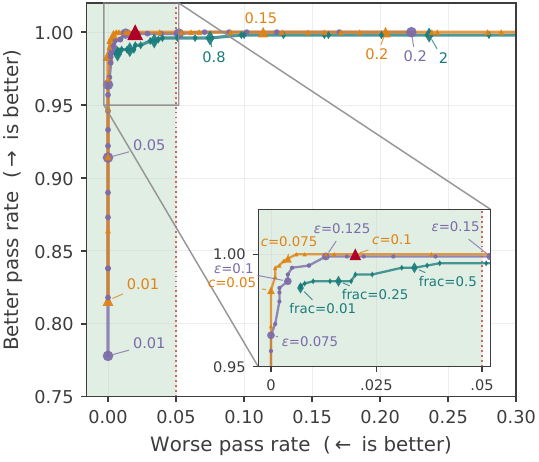}
      \caption{Uniform ($N\!=\!200$).}
      \label{fig:app-dist-uniform}
    \end{subfigure}

    \vspace{6pt}
    \begin{subfigure}[t]{\textwidth}
      \centering
      \includegraphics[width=\textwidth]{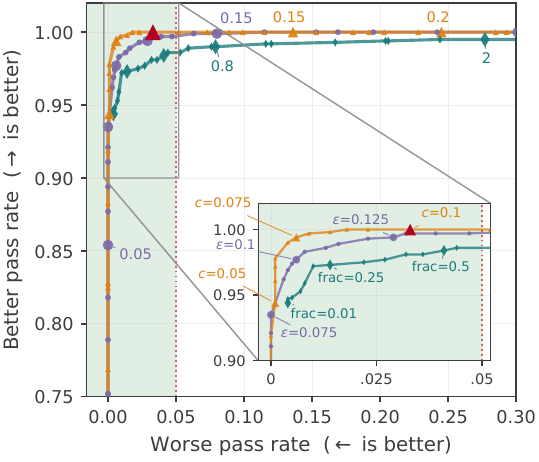}
      \caption{Gaussian ($N\!=\!200$).}
      \label{fig:app-dist-gaussian}
    \end{subfigure}
  \end{minipage}\hfill
  \begin{minipage}[c]{0.67\textwidth}
    \begin{subfigure}[t]{\textwidth}
      \centering
      \includegraphics[width=\textwidth]{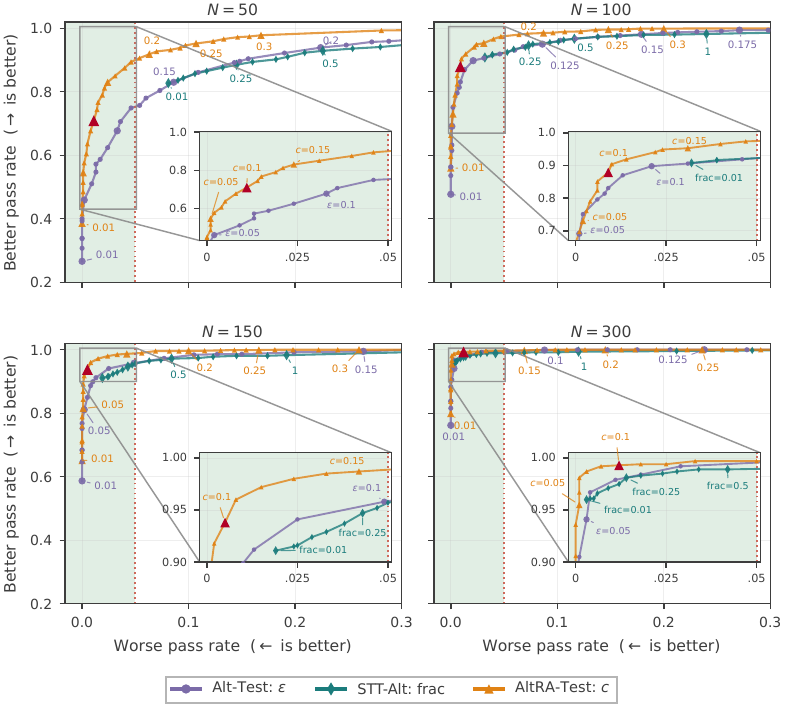}
      \caption{Other numbers of items (skewed distribution).}
      \label{fig:sim-pareto-ngrid}
    \end{subfigure}
  \end{minipage}
  \caption{Comparison with the baselines in other settings, using $k\!=\!3$ and varying the hyperparameter of each test ($c$, $\varepsilon$, or $\mathrm{frac}$). (a) and (b): other generative families. (c): $N\!=\!50$, $100$, $150$, and $300$ items. All plots share the legend of (c). The \anlt{} dominates the baselines in all settings, with the largest gains at smaller $N$.}
  \label{fig:app-dist}
\end{figure}

Figures~\ref{fig:app-dist-uniform} and~\ref{fig:app-dist-gaussian} repeat the comparison of Figure~\ref{fig:sim-pareto} for these two distributions. As in Section~\ref{sec:sim}, the \anlt{} achieves the best trade-off. Its curve is at or above the curve of the \alttest{} in nearly all of the valid region, and \textsc{STT-Alt} is below both. Both distributions are easier than the skewed one, and the \anlt{} and the \alttest{} reach a better-model pass rate of $1.00$ within the valid region. The comparison in Section~\ref{sec:sim} is therefore the most conservative of the three. We also find that the default $c=0.1$ is valid for all three distributions.

\paragraph{Number of items.}
In Section~\ref{sec:sim}, we compare the tests with $N=200$ items. Since annotated items are often scarce, we repeat the comparison with $N\in\{50,100,150,300\}$ (Figure~\ref{fig:sim-pareto-ngrid}). The \anlt{} achieves the best trade-off at every $N$, and its advantage is largest when there are few items. As $N$ grows, all tests approach a better-model pass rate of $1.00$, and the differences become small.

The default $c=0.1$ is a good choice for every number of items. At every $N$, it is within the valid region, with a worse-model pass rate of about $0.01$. With few items, this value is conservative, and a larger $c$ would increase the better-model pass rate while remaining within the valid region. Researchers who have few items can therefore calibrate $c$ for their number of items.

\begin{figure}[!htbp]
  \centering
  \begin{subfigure}[t]{0.45\textwidth}
    \centering
    \includegraphics[width=\textwidth]{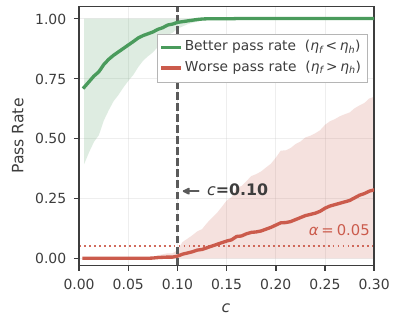}
    \caption{Better/worse-model pass rates against $c$.}
    \label{fig:app-calib-c}
  \end{subfigure}
  \hfill
  \begin{subfigure}[t]{0.45\textwidth}
    \centering
    \includegraphics[width=\textwidth]{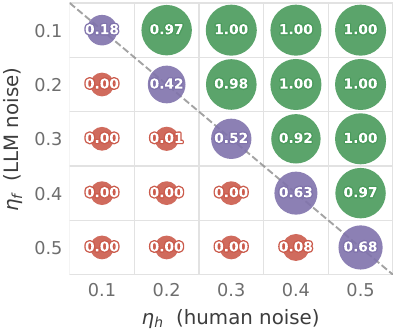}
    \caption{Pass rate over the noise grid at $c = 0.1$.}
    \label{fig:app-calib-grid}
  \end{subfigure}
    \caption{Calibration and behavior of the \anlt{}. (a) Each curve is the mean
    over the conditions of its regime and the shaded band is one standard deviation
    across them; the dashed line marks the recommended $c = 0.1$. (b) Per-condition
    pass rate at that setting; passing is correct where the LLM is less noisy, an
    error where it is noisier, and neither on the diagonal.}
  \label{fig:app-calib}
\end{figure}

\paragraph{Calibration of $c$.}
We choose the constant $c$ with the basic simulation. Figure~\ref{fig:app-calib-c} presents the better-model and worse-model pass rates for different values of $c$. The better-model pass rate increases sharply until $c=0.1$ and changes little for larger values. The worse-model pass rate is about $0.01$ up to $c=0.1$ and then increases, and it reaches $\alpha=0.05$ near $c=0.135$. We therefore choose $c=0.1$, which gives a better-model pass rate of $0.98$ and a worse-model pass rate of $0.01$.

Figure~\ref{fig:app-calib-grid} presents the pass rate for every pair of noise levels at $c=0.1$. In every better-model condition, the LLM passes in at least $0.92$ of the runs. Only one worse-model condition exceeds $\alpha$, with a pass rate of $0.08$. In this condition, $(\eta_h,\eta_f)=(0.4,0.5)$, the LLM is only slightly noisier than the humans, and the decrease in agreement is smaller than the margin.

\subsection{Test and Task Parameters}
\label{app:sim-params}

We examine three parameters of the setup, which are the panel size $k$, the number of annotators per item $m$, and the number of labels $L$. For each parameter, we repeat the full noise grid with $N=200$ and with $N=100$, and keep the other parameters at their default values, except that $m\ge k$ so that $k$ distinct annotations can be drawn for every item. In all cases, the worse-model pass rate stays below $0.05$. We also examine the agreement metric.

\paragraph{Panel size $k$.}
The margin includes the adjustment $k_c/k$, which should keep the meaning of $c$ the same for every panel size (see Appendix~\ref{sec:panel-size-scaling}). Figures~\ref{fig:app-struct-k} and~\ref{fig:app-struct-k-N100} support this. When $k$ varies from $3$ to $7$, the better-model pass rate at $N=200$ is at least $0.98$ for every $k$. At $N=100$, it is somewhat lower for $k=3$ and stable for $k\ge4$.

\paragraph{Number of annotators per item $m$.}
The test samples $k$ annotations from the $m$ annotations of each item. Figures~\ref{fig:app-struct-m} and~\ref{fig:app-struct-m-N100} show that a larger $m$ slightly improves both pass rates. At $N=200$, when $m$ increases from $3$ to $12$, the better-model pass rate increases from $0.96$ to $0.99$, and the worse-model pass rate decreases from $0.02$ to $0.01$. The effect is larger at $N=100$. When the items have few annotations, we therefore recommend using more items.

\paragraph{Number of labels $L$.}
Figures~\ref{fig:app-struct-Y} and~\ref{fig:app-struct-Y-N100} show that binary tasks are the most difficult. With $L=2$, the better-model pass rate at $N=200$ is $0.82$, compared with at least $0.98$ for $L\ge4$, and it is lower at $N=100$. With two labels, the agreement expected by chance is high, so the same error rate changes the chance-corrected agreement less, and the substitution gap is harder to detect. Binary tasks therefore require more items.

\begin{figure}[!tb]
  \centering
  \begin{subfigure}[t]{0.32\textwidth}
    \centering
    \includegraphics[width=\textwidth]{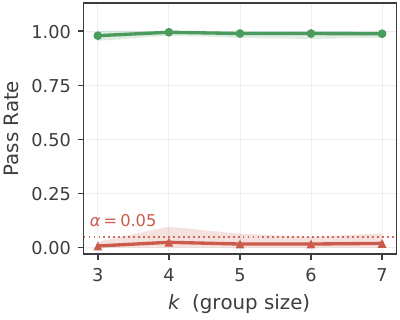}
    \caption{Panel size $k$ ($N\!=\!200$).}
    \label{fig:app-struct-k}
  \end{subfigure}
  \hfill
  \begin{subfigure}[t]{0.32\textwidth}
    \centering
    \includegraphics[width=\textwidth]{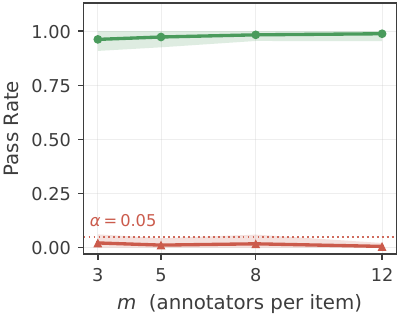}
    \caption{Annotators $m$ ($N\!=\!200$).}
    \label{fig:app-struct-m}
  \end{subfigure}
  \hfill
  \begin{subfigure}[t]{0.32\textwidth}
    \centering
    \includegraphics[width=\textwidth]{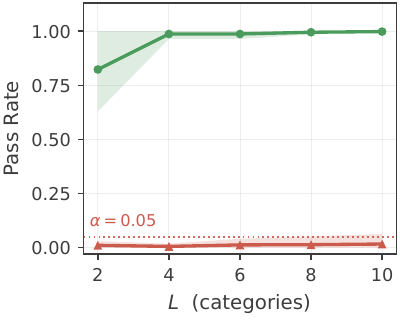}
    \caption{Labels $L$ ($N\!=\!200$).}
    \label{fig:app-struct-Y}
  \end{subfigure}

  \vspace{4pt}

  \begin{subfigure}[t]{0.32\textwidth}
    \centering
    \includegraphics[width=\textwidth]{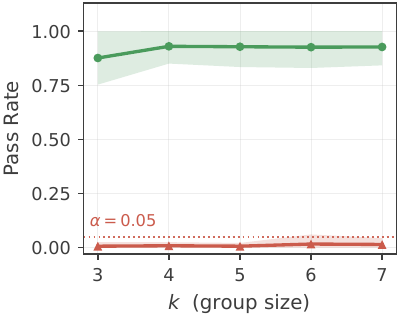}
    \caption{Panel size $k$ ($N\!=\!100$).}
    \label{fig:app-struct-k-N100}
  \end{subfigure}
  \hfill
  \begin{subfigure}[t]{0.32\textwidth}
    \centering
    \includegraphics[width=\textwidth]{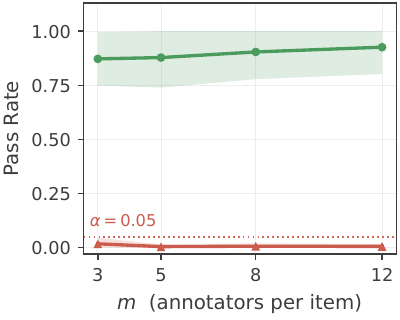}
    \caption{Annotators $m$ ($N\!=\!100$).}
    \label{fig:app-struct-m-N100}
  \end{subfigure}
  \hfill
  \begin{subfigure}[t]{0.32\textwidth}
    \centering
    \includegraphics[width=\textwidth]{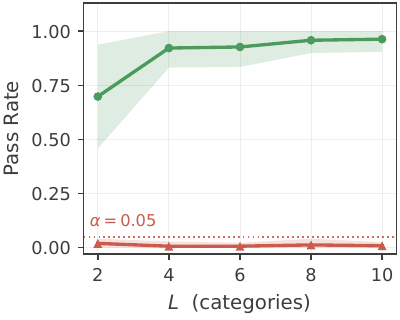}
    \caption{Labels $L$ ($N\!=\!100$).}
    \label{fig:app-struct-Y-N100}
  \end{subfigure}

  \vspace{0.4em}
  \includegraphics[width=0.6\textwidth]{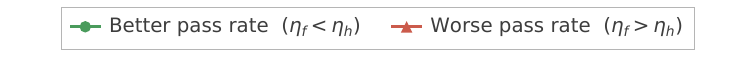}
  \caption{Simulations of the \anlt{} with $c\!=\!0.1$, varying the panel size $k$, the number of annotators per item $m$, and the number of labels $L$, with $N\!=\!200$ (top) and $N\!=\!100$ (bottom). For the panel size, the margin is scaled by $k_c/k$. In all cases, the worse-model pass rate is below $0.05$.}
  \label{fig:app-struct}
\end{figure}

\begin{figure}[!tb]
  \centering
  \begin{subfigure}[c]{0.36\textwidth}
    \centering
    \includegraphics[width=\textwidth]{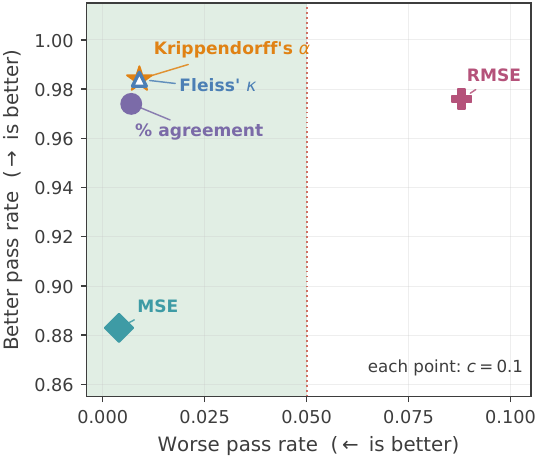}
    \caption{Agreement metrics.}
    \label{fig:app-metric}
  \end{subfigure}\hfill
  \begin{subfigure}[c]{0.6\textwidth}
    \centering\footnotesize
    \setlength{\tabcolsep}{2.5pt}
    \begin{tabular}{@{}lcccc@{}}
    \toprule
    & \multicolumn{2}{c}{\textbf{Bias in humans and LLM}} & \multicolumn{2}{c}{\textbf{Bias only in humans}} \\
    & Worse-model & Better-model & Worse-model & Better-model \\
    \midrule
    \anlt{} & $0.02$ & $1.00$ & $0.02$ & $0.91$ \\
    \alttest{} & $0.01$ & $0.98$ & $0.18$ & $0.90$ \\
    \textsc{STT-Alt} & $0.07$ & $0.97$ & $0.09$ & $0.91$ \\
    \bottomrule
    \end{tabular}
    \caption{Shared bias.}
    \label{tab:app-shared-bias}
  \end{subfigure}
  \caption{Additional simulations with $N\!=\!200$ items. (a) The \anlt{} with $c\!=\!0.1$ and different agreement metrics. (b) Worse-model and better-model pass rates when the bias is shared by the humans and the LLM, and when only the humans have it. Each test uses its default setting ($c\!=\!0.1$, $\varepsilon\!=\!0.1$, $\mathrm{frac}\!=\!0.5$).}
  \label{fig:app-metric-bias}
\end{figure}

\paragraph{Agreement metric.}
The \anlt{} can be applied with different agreement metrics. Figure~\ref{fig:app-metric} repeats the basic simulation with four other metrics. Krippendorff's $\alpha$ and Fleiss' $\kappa$ give nearly identical results, with the highest better-model pass rates and a worse-model pass rate of about $0.01$. Percent agreement, which does not correct for chance, is slightly lower. The two error-based metrics, MSE and RMSE, perform worse. MSE reduces the better-model pass rate to $0.88$, and with RMSE the worse-model pass rate is $0.09$, above $\alpha$. These metrics do not correct for chance, so $c=0.1$ is no longer a fixed proportion of the human disagreement, and a smaller value of $c$ should be used for RMSE.

\subsection{Annotator Behavior and Low Human Agreement}
\label{app:sim-behavior}

\paragraph{Personal bias of annotators.}
In the basic simulation, all annotators make the same kind of errors, since an annotator who errs samples a label from the label prior. In practice, each annotator may have a personal bias and tend toward specific labels when making an error. We therefore simulate human annotators with a personal bias. When such an annotator errs, the label is sampled from a mixture of the label prior and a distribution that is specific to this annotator (the full details are in Appendix~\ref{app:impl-sim}).

Figure~\ref{fig:app-bias} repeats the comparison of Figure~\ref{fig:sim-pareto} for this simulation. The results are similar to those without a personal bias. The \anlt{} outperforms the baselines within the valid region, and $c=0.1$ gives a better-model pass rate of $0.99$ with a worse-model pass rate of $0.01$. The \anlt{} is therefore robust to annotators with different biases.

\paragraph{Bias shared by the humans and the LLM.}
Next, we examine what happens when the humans and the LLM have the same bias, so that they tend to make the same error. Such shared errors are a limitation of every test that relies on human annotations as the reference, since an LLM that repeats the errors of the humans agrees with them. For every item, we choose in advance one incorrect label, and an annotator who errs chooses this label with probability $0.5$ (see Appendix~\ref{app:impl-sim}). We examine two cases. In the first, both the humans and the LLM have this bias. In the second, only the humans have it.

Figure~\ref{tab:app-shared-bias} presents the results for the default setting of each test. When both the humans and the LLM share the bias, the \anlt{} behaves as in the basic simulation, with a worse-model pass rate of $0.02$ and a better-model pass rate of $1.00$. When only the humans have the bias, a better LLM agrees less with the humans, and the better-model pass rate of all three tests decreases to about $0.90$. Also in this case, the worse-model pass rate of the \anlt{} remains $0.02$, while the worse-model pass rates of the \alttest{} and \textsc{STT-Alt} are above $\alpha$. To conclude, while shared errors are a limitation of every test that relies on human annotations, the \anlt{} is less affected by them than the other tests. Its worse-model pass rate stays below $\alpha$ in both cases, and its better-model pass rate is at least as high as those of the other tests.

\begin{figure}[!htbp]
  \centering
  \begin{subfigure}[t]{0.48\textwidth}
    \centering
    \includegraphics[width=\textwidth]{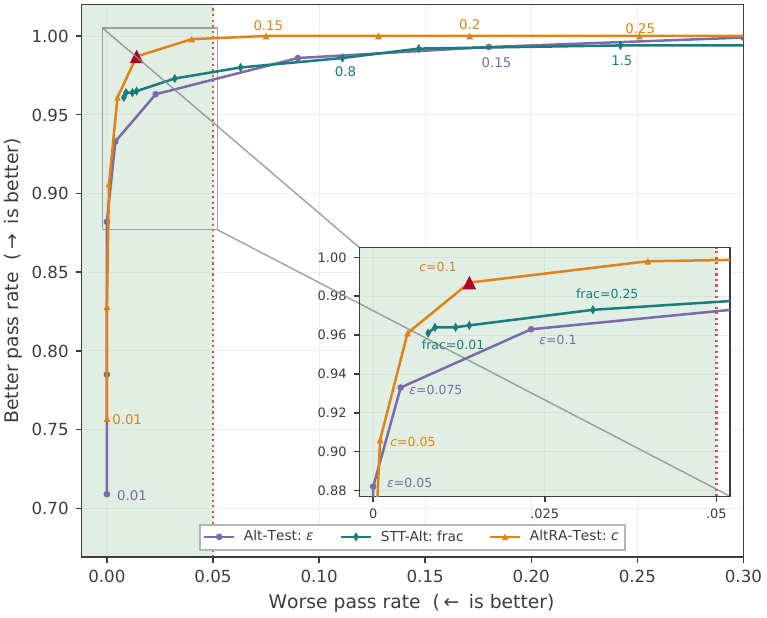}
    \caption{Personal bias.}
    \label{fig:app-bias}
  \end{subfigure}\hfill
  \begin{subfigure}[t]{0.48\textwidth}
    \centering
    \includegraphics[width=\textwidth]{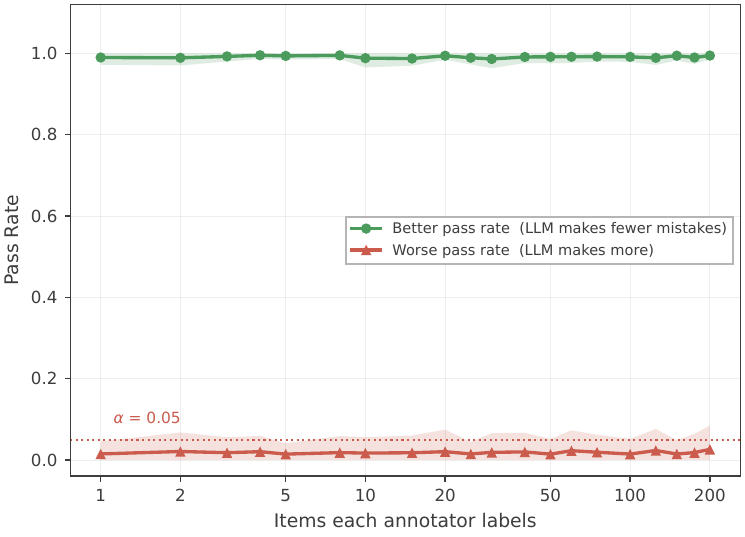}
    \caption{Sparse annotations.}
    \label{fig:app-sparse}
  \end{subfigure}
  \caption{Simulations in which every human annotator has a personal bias, using $N\!=\!200$ items and $k\!=\!3$. (a) Comparison with the baselines, while varying the hyperparameter of each test ($c$, $\varepsilon$, or $\mathrm{frac}$). The green region is the valid region, and the red triangle is $c\!=\!0.1$. The \anlt{} outperforms the baselines within the valid region. (b) The \anlt{} with $c\!=\!0.1$ and sparse annotations. The X-axis is the number of items that each annotator labels, where $200$ means that every annotator labels all items. The pass rates do not change with it.}
  \label{fig:app-bias-sparse}
\end{figure}

\paragraph{Sparse annotations.}
To support an apples-to-apples comparison with the baselines, we have so far considered only simulations with shared annotators, in which every annotator labels every item. In these simulations, we found that the \anlt{} outperforms the baselines. However, the motivation for the \anlt{} is a different and very common setup, in which the annotations are sparse and every annotator labels only a few items. In this setup, the baselines cannot be applied, so we cannot compare with them. We therefore next analyze the behavior of the \anlt{} alone. We vary the number of items that each annotator labels from $t=1$ to $t=200$, where $t=200$ means that every annotator labels all items. The annotators have a personal bias, as above.

Figure~\ref{fig:app-sparse} presents the results. The number of items that each annotator labels does not affect the test. At every $t$, the better-model pass rate is at least $0.98$ and the worse-model pass rate is at most $0.02$, including $t=1$, in which no two items share an annotator. This is expected, since the test uses only the annotations of each item and never the identity of the annotators.

\paragraph{Low human agreement.}
We designed the adaptive margin to increase with the human disagreement, since we want the test to support subjective tasks, in which some disagreement between the annotators is inherent. In such tasks, the test therefore permits a larger decrease in agreement. This raises the question of what happens when human agreement is very low. In this case, the margin is large, and an LLM that carries no information about the items may pass the test.

To examine when this happens, we simulate humans with low agreement, together with an LLM that assigns random labels (see Appendix~\ref{app:impl-sim}). When the human IAA is between $0.1$ and $0.2$, the random LLM never passes. When it is between $0.05$ and $0.1$, the random LLM passes in $6\%$ to $12\%$ of the runs, and below $0.05$, in $62\%$ to $92\%$ of the runs. This is expected, since humans with an IAA near zero are themselves close to random, and a random LLM behaves like them. We also examine an LLM that always assigns the most frequent label. This LLM passes in at most $0.32$ of the runs, and only when the human IAA is near zero.

The test should therefore not be used when human agreement is very low. Our recommendation to use it only when the human IAA is at least $0.2$ (see Section~\ref{sec:exp}) adds a safety margin above the level at which an uninformative LLM starts to pass.
\FloatBarrier

\section{Real-World Data: Additional Results}
\label{app:real-results}

In this section, we present additional results for the real-world experiments of Section~\ref{sec:exp}. In Appendix~\ref{app:real-full}, we start with the full results for every dataset, LLM, and prompting technique, and we validate the test on a dataset with gold labels. Then, in Appendix~\ref{app:real-comparison}, we compare the \anlt{} with the existing tests on the datasets to which they can be applied. In Appendix~\ref{app:real-items}, we examine how the number of items and the panel size affect the results, and finally, in Appendix~\ref{app:real-prompting}, we examine the effect of the prompting technique.

\subsection{Full Results and Validation with Gold Labels}
\label{app:real-full}

\paragraph{Full results.}
Tables~\ref{tab:full-text}--\ref{tab:full-video} present the full results of the real-world experiments, for every dataset, LLM, and prompting technique. They include Measuring-Hate-Speech and the LLMs that we omit from Tables~\ref{tab:main-grid-rmixed} and~\ref{tab:method-passrate}. Each cell reports the mixed-panel agreement $\Rm$, and green cells mark the cases in which the LLM passes the test. For every dataset, we also report the human-only agreement $\Rh$, with which $\Rm$ should be compared. Gray crosses mark configurations that we did not evaluate.

\paragraph{Validation with gold labels.}
In real-world datasets, we usually do not know whether an LLM is a better annotator than the humans, which is why we evaluate the test with simulations (see Section~\ref{sec:sim}). CIFAR-10H is an exception, since it also includes the original CIFAR-10 labels~\citep{krizhevsky2009learning}. We use these labels as gold labels and examine whether the LLMs that pass the test are also at least as accurate as the humans.

On the $300$ evaluated items, the average accuracy of a human annotator is $0.94$. Table~\ref{tab:app-cifar} compares the result of the test with the accuracy of the LLM, for $45$ combinations of LLMs and prompting techniques. The results agree in $44$ of the $45$ cases. Of the $9$ LLMs that are at least as accurate as the humans, $8$ pass the test, and none of the $36$ LLMs that are less accurate than the humans passes. These results are in line with the simulations, in which the test accepts better models and worse models fail it.

\begin{table}[H]
\centering\small
\caption{Validation on CIFAR-10H with the CIFAR-10 labels as gold labels. The table counts the $45$ combinations of LLMs and prompting techniques, according to the result of the \anlt{} and to whether the accuracy of the LLM is at least the average human accuracy ($0.94$).}
\label{tab:app-cifar}
\setlength{\tabcolsep}{8pt}
\begin{tabular}{@{}lcc@{}}
\toprule
& \textbf{Accuracy $\ge$ human} & \textbf{Accuracy $<$ human} \\
\midrule
Passes the test & $8$ & $0$ \\
Fails the test & $1$ & $36$ \\
\bottomrule
\end{tabular}
\end{table}

\footnotesize\setlength{\tabcolsep}{3.5pt}\renewcommand{\arraystretch}{1.15}
\begin{longtable}{@{}ll!{\hspace{3pt}\color{black!25}\vrule width 0.4pt\hspace{3pt}}ccccc!{\hspace{3pt}\color{black!25}\vrule width 0.4pt\hspace{3pt}}ccccc@{}}
\caption{\textbf{\faFile*\ Text.} Test outcomes and mixed-panel agreement $\bar R_m$ for every text judge and prompting technique. \colorbox{bestgreen}{Dark green} marks the judge's best technique when it passes; \colorbox{passgreen}{light green} marks a passing technique that is not the judge's best; \colorbox{gray!18}{gray} marks the judge's best technique when it fails. \textbf{Bold} marks the highest $\bar R_m$ for each dataset.}\label{tab:full-text}\\
\toprule \textbf{Family} & \textbf{Judge} & \textbf{Zero} & \textbf{Few} & \textbf{CoT} & \textbf{Ens} & \textbf{GEPA} & \textbf{Zero} & \textbf{Few} & \textbf{CoT} & \textbf{Ens} & \textbf{GEPA} \\\midrule\endfirsthead
\multicolumn{12}{@{}l}{\emph{Text (continued)}}\\
\toprule \textbf{Family} & \textbf{Judge} & \textbf{Zero} & \textbf{Few} & \textbf{CoT} & \textbf{Ens} & \textbf{GEPA} & \textbf{Zero} & \textbf{Few} & \textbf{CoT} & \textbf{Ens} & \textbf{GEPA} \\\midrule\endhead
\bottomrule\endfoot
 & & \multicolumn{5}{c!{\hspace{3pt}\color{black!25}\vrule width 0.4pt\hspace{3pt}}}{\textbf{DICES-350}\ \small($\bar R_h=0.20$)} & \multicolumn{5}{c}{\textbf{EPIC}\ \small($\bar R_h=0.20$)} \\
\cmidrule[0.4pt]{1-12}
\textit{Gemini} & Gem-P & $0.14$ & \cellcolor{bestgreen}$0.20$ & $0.14$ & $0.15$ & \xmarkg & \cellcolor{passgreen}$0.21$ & \cellcolor{bestgreen}$0.23$ & \cellcolor{passgreen}$0.22$ & \cellcolor{passgreen}$0.22$ & \xmarkg \\
 & Gem-F & $0.05$ & \cellcolor{gray!18}$0.10$ & $0.06$ & $0.05$ & \xmarkg & \cellcolor{bestgreen}$0.19$ & $0.16$ & \cellcolor{passgreen}$0.19$ & \cellcolor{passgreen}$0.19$ & \xmarkg \\
 & Gem-L & $0.10$ & \cellcolor{gray!18}$0.12$ & $0.10$ & $0.10$ & \xmarkg & \cellcolor{passgreen}$0.20$ & \cellcolor{passgreen}$0.20$ & \cellcolor{bestgreen}$0.24$ & \cellcolor{passgreen}$0.22$ & \xmarkg \\
\textit{GPT} & GPT-T & $0.14$ & \cellcolor{bestgreen}$0.19$ & $0.16$ & $0.17$ & \xmarkg & \cellcolor{passgreen}$0.22$ & \cellcolor{bestgreen}$0.23$ & \cellcolor{passgreen}$0.22$ & \cellcolor{passgreen}$0.23$ & \xmarkg \\
 & GPT-L & $0.16$ & \cellcolor{bestgreen}$0.19$ & $0.16$ & \cellcolor{passgreen}$0.18$ & \xmarkg & \cellcolor{passgreen}$0.23$ & \cellcolor{bestgreen}$\mathbf{0.28}$ & \cellcolor{passgreen}$0.21$ & \cellcolor{passgreen}$0.24$ & \xmarkg \\
\textit{Qwen} & Q3-8b & $0.06$ & $0.06$ & \cellcolor{gray!18}$0.09$ & $0.07$ & $0.06$ & \cellcolor{passgreen}$0.19$ & \cellcolor{passgreen}$0.26$ & \cellcolor{passgreen}$0.20$ & \cellcolor{bestgreen}$0.26$ & \cellcolor{passgreen}$0.24$ \\
 & Q3-14b & $0.15$ & $0.16$ & $0.16$ & \cellcolor{gray!18}$0.16$ & $0.15$ & \cellcolor{passgreen}$0.24$ & \cellcolor{passgreen}$0.25$ & \cellcolor{passgreen}$0.22$ & \cellcolor{bestgreen}$0.25$ & \cellcolor{passgreen}$0.22$ \\
\textit{Gemma} & G3-4b & $0.11$ & $0.12$ & $0.12$ & \cellcolor{gray!18}$0.12$ & $0.11$ & $-0.01$ & $0.09$ & $-0.06$ & \cellcolor{gray!18}$0.09$ & $-0.02$ \\
 & G3-12b & \cellcolor{passgreen}$0.19$ & $0.18$ & \cellcolor{bestgreen}$\mathbf{0.22}$ & \cellcolor{passgreen}$0.19$ & \cellcolor{passgreen}$0.19$ & \cellcolor{passgreen}$0.22$ & \cellcolor{passgreen}$0.24$ & \cellcolor{passgreen}$0.19$ & \cellcolor{bestgreen}$0.24$ & \cellcolor{passgreen}$0.24$ \\
\textit{Llama} & L3.2-3b & $0.09$ & $0.10$ & \cellcolor{gray!18}$0.14$ & $0.11$ & $0.09$ & $0.13$ & $0.16$ & $0.01$ & \cellcolor{bestgreen}$0.19$ & $-0.14$ \\
 & L3.1-8b & $0.09$ & $0.14$ & $0.15$ & \cellcolor{gray!18}$0.15$ & $0.09$ & \cellcolor{gray!18}$0.09$ & $-0.02$ & $0.02$ & $0.06$ & $0.09$ \\
\addlinespace[3pt]
\midrule
 & & \multicolumn{5}{c!{\hspace{3pt}\color{black!25}\vrule width 0.4pt\hspace{3pt}}}{\textbf{MultiPref}\ \small($\bar R_h=0.28$)} & \multicolumn{5}{c}{\textbf{ChaosNLI}\ \small($\bar R_h=0.28$)} \\
\cmidrule[0.4pt]{1-12}
\textit{Gemini} & Gem-P & \cellcolor{passgreen}$\mathbf{0.33}$ & \cellcolor{passgreen}$0.31$ & \cellcolor{passgreen}$\mathbf{0.33}$ & \cellcolor{bestgreen}$\mathbf{0.33}$ & \xmarkg & \cellcolor{bestgreen}$0.31$ & \cellcolor{passgreen}$0.30$ & \cellcolor{passgreen}$0.31$ & \cellcolor{passgreen}$0.31$ & \xmarkg \\
 & Gem-F & \cellcolor{bestgreen}$0.32$ & $0.27$ & \cellcolor{passgreen}$0.30$ & \cellcolor{passgreen}$0.32$ & \xmarkg & \cellcolor{passgreen}$0.28$ & \cellcolor{passgreen}$0.28$ & \cellcolor{bestgreen}$0.30$ & \cellcolor{passgreen}$0.29$ & \xmarkg \\
 & Gem-L & \cellcolor{bestgreen}$0.32$ & $0.26$ & \cellcolor{passgreen}$0.28$ & \cellcolor{passgreen}$0.31$ & \xmarkg & \cellcolor{bestgreen}$0.32$ & \cellcolor{passgreen}$0.30$ & \cellcolor{passgreen}$0.29$ & \cellcolor{passgreen}$0.31$ & \xmarkg \\
\textit{GPT} & GPT-T & $0.25$ & $0.24$ & $0.24$ & \cellcolor{gray!18}$0.25$ & \xmarkg & \cellcolor{bestgreen}$\mathbf{0.33}$ & $0.26$ & \cellcolor{passgreen}$0.31$ & \cellcolor{passgreen}$\mathbf{0.33}$ & \xmarkg \\
 & GPT-L & $0.24$ & $0.24$ & $0.24$ & \cellcolor{gray!18}$0.25$ & \xmarkg & \cellcolor{passgreen}$0.29$ & \cellcolor{bestgreen}$0.32$ & \cellcolor{passgreen}$0.26$ & \cellcolor{passgreen}$0.30$ & \xmarkg \\
\textit{Qwen} & Q3-8b & $0.18$ & $0.09$ & \cellcolor{gray!18}$0.19$ & $0.14$ & $0.18$ & \cellcolor{passgreen}$0.26$ & \cellcolor{bestgreen}$0.30$ & \cellcolor{passgreen}$0.26$ & \cellcolor{passgreen}$0.28$ & \cellcolor{passgreen}$0.26$ \\
 & Q3-14b & \cellcolor{passgreen}$0.28$ & $0.19$ & \cellcolor{passgreen}$0.26$ & \cellcolor{bestgreen}$0.30$ & $0.26$ & \cellcolor{passgreen}$0.32$ & \cellcolor{bestgreen}$0.32$ & \cellcolor{passgreen}$0.32$ & \cellcolor{passgreen}$0.32$ & \cellcolor{passgreen}$0.31$ \\
\textit{Gemma} & G3-4b & $0.02$ & $0.16$ & $0.14$ & \cellcolor{gray!18}$0.24$ & $0.02$ & \cellcolor{passgreen}$0.27$ & \cellcolor{bestgreen}$0.29$ & \cellcolor{passgreen}$0.28$ & \cellcolor{passgreen}$0.29$ & \cellcolor{passgreen}$0.27$ \\
 & G3-12b & $0.25$ & $0.21$ & $0.25$ & \cellcolor{bestgreen}$0.26$ & $0.25$ & \cellcolor{passgreen}$0.31$ & \cellcolor{bestgreen}$\mathbf{0.33}$ & \cellcolor{passgreen}$0.29$ & \cellcolor{passgreen}$0.32$ & \cellcolor{passgreen}$0.30$ \\
\textit{Llama} & L3.2-3b & $0.04$ & $0.02$ & \cellcolor{gray!18}$0.20$ & $0.03$ & $0.01$ & $0.18$ & $0.20$ & \cellcolor{gray!18}$0.23$ & $0.20$ & $0.18$ \\
 & L3.1-8b & $0.12$ & $0.16$ & $0.12$ & \cellcolor{gray!18}$0.19$ & $0.11$ & \cellcolor{passgreen}$0.28$ & \cellcolor{passgreen}$0.29$ & \cellcolor{passgreen}$0.28$ & \cellcolor{bestgreen}$0.29$ & \cellcolor{passgreen}$0.28$ \\
\addlinespace[3pt]
\midrule
 & & \multicolumn{5}{c!{\hspace{3pt}\color{black!25}\vrule width 0.4pt\hspace{3pt}}}{\textbf{Persuasion}\ \small($\bar R_h=0.31$)} & \multicolumn{5}{c}{\textbf{10k-Prompts}\ \small($\bar R_h=0.37$)} \\
\cmidrule[0.4pt]{1-12}
\textit{Gemini} & Gem-P & \cellcolor{gray!18}$0.07$ & $0.06$ & $0.06$ & $0.06$ & \xmarkg & $0.31$ & \cellcolor{bestgreen}$0.44$ & \cellcolor{passgreen}$0.40$ & \cellcolor{passgreen}$0.43$ & \xmarkg \\
 & Gem-F & $0.09$ & \cellcolor{gray!18}$0.17$ & $0.07$ & $0.08$ & \xmarkg & $0.36$ & \cellcolor{passgreen}$0.37$ & $0.36$ & \cellcolor{bestgreen}$0.42$ & \xmarkg \\
 & Gem-L & $0.01$ & \cellcolor{gray!18}$0.05$ & $0.01$ & $0.02$ & \xmarkg & \cellcolor{passgreen}$\mathbf{0.45}$ & \cellcolor{passgreen}$0.41$ & \cellcolor{passgreen}$0.43$ & \cellcolor{bestgreen}$\mathbf{0.45}$ & \xmarkg \\
\textit{GPT} & GPT-T & $0.18$ & \cellcolor{gray!18}$\mathbf{0.27}$ & $0.15$ & $0.17$ & \xmarkg & \cellcolor{bestgreen}$0.38$ & \cellcolor{passgreen}$0.38$ & $0.35$ & \cellcolor{passgreen}$0.38$ & \xmarkg \\
 & GPT-L & $0.08$ & \cellcolor{gray!18}$0.17$ & $0.10$ & $0.10$ & \xmarkg & \cellcolor{passgreen}$0.38$ & $0.35$ & $0.35$ & \cellcolor{bestgreen}$0.40$ & \xmarkg \\
\textit{Qwen} & Q3-8b & $0.24$ & $0.24$ & $0.12$ & \cellcolor{gray!18}$0.25$ & $0.24$ & \cellcolor{passgreen}$0.39$ & \cellcolor{passgreen}$0.38$ & \cellcolor{passgreen}$0.39$ & \cellcolor{bestgreen}$0.42$ & \cellcolor{passgreen}$0.40$ \\
 & Q3-14b & $0.19$ & $0.22$ & $0.18$ & \cellcolor{gray!18}$0.23$ & $0.19$ & \cellcolor{passgreen}$0.39$ & \cellcolor{passgreen}$0.39$ & $0.15$ & \cellcolor{bestgreen}$0.41$ & \cellcolor{passgreen}$0.41$ \\
\textit{Gemma} & G3-4b & $0.17$ & $-0.01$ & \cellcolor{gray!18}$0.18$ & $0.11$ & $0.17$ & \cellcolor{gray!18}$0.36$ & $0.12$ & $0.33$ & $0.23$ & $0.36$ \\
 & G3-12b & $0.15$ & \cellcolor{gray!18}$0.22$ & $0.17$ & $0.21$ & $0.15$ & $0.35$ & $0.32$ & \cellcolor{bestgreen}$0.39$ & \cellcolor{passgreen}$0.38$ & $0.35$ \\
\textit{Llama} & L3.2-3b & $0.06$ & \cellcolor{gray!18}$0.20$ & $0.10$ & $0.17$ & $0.06$ & $0.33$ & $0.23$ & $0.23$ & \cellcolor{gray!18}$0.32$ & $0.32$ \\
 & L3.1-8b & $0.06$ & $0.06$ & \cellcolor{gray!18}$0.09$ & $0.05$ & $0.06$ & $0.27$ & $0.28$ & $0.28$ & \cellcolor{gray!18}$0.32$ & $0.27$ \\
\addlinespace[3pt]
\midrule
 & & \multicolumn{5}{c!{\hspace{3pt}\color{black!25}\vrule width 0.4pt\hspace{3pt}}}{\textbf{Scruples}\ \small($\bar R_h=0.38$)} & \multicolumn{5}{c}{\textbf{Meas.-Hate-Speech}\ \small($\bar R_h=0.53$)} \\
\cmidrule[0.4pt]{1-12}
\textit{Gemini} & Gem-P & \cellcolor{passgreen}$\mathbf{0.45}$ & \cellcolor{passgreen}$0.44$ & \cellcolor{bestgreen}$\mathbf{0.45}$ & \cellcolor{passgreen}$\mathbf{0.45}$ & \xmarkg & $0.50$ & \cellcolor{gray!18}$\mathbf{0.51}$ & $0.50$ & $0.50$ & \xmarkg \\
 & Gem-F & \cellcolor{passgreen}$0.44$ & \cellcolor{passgreen}$0.43$ & \cellcolor{passgreen}$0.44$ & \cellcolor{bestgreen}$\mathbf{0.45}$ & \xmarkg & $0.49$ & $0.50$ & \cellcolor{gray!18}$0.50$ & $0.49$ & \xmarkg \\
 & Gem-L & \cellcolor{passgreen}$0.41$ & \cellcolor{passgreen}$0.40$ & \cellcolor{bestgreen}$0.43$ & \cellcolor{passgreen}$0.42$ & \xmarkg & \cellcolor{gray!18}$0.47$ & $0.47$ & $0.45$ & $0.46$ & \xmarkg \\
\textit{GPT} & GPT-T & \cellcolor{passgreen}$0.39$ & \cellcolor{passgreen}$0.37$ & \cellcolor{passgreen}$0.39$ & \cellcolor{bestgreen}$0.39$ & \xmarkg & $0.47$ & $0.44$ & \cellcolor{gray!18}$0.48$ & $0.47$ & \xmarkg \\
 & GPT-L & $0.36$ & $0.32$ & $0.35$ & \cellcolor{gray!18}$0.36$ & \xmarkg & $0.41$ & $0.41$ & \cellcolor{gray!18}$0.42$ & $0.42$ & \xmarkg \\
\textit{Qwen} & Q3-8b & $0.08$ & $0.15$ & \cellcolor{gray!18}$0.18$ & $0.10$ & $0.08$ & $0.29$ & \cellcolor{gray!18}$0.43$ & $0.31$ & $0.36$ & $0.35$ \\
 & Q3-14b & $0.04$ & \cellcolor{gray!18}$0.14$ & $0.10$ & $0.06$ & $0.08$ & $0.29$ & \cellcolor{gray!18}$0.40$ & $0.30$ & $0.34$ & $0.35$ \\
\textit{Gemma} & G3-4b & $0.06$ & $0.07$ & \cellcolor{gray!18}$0.14$ & $0.07$ & $0.06$ & $0.25$ & \cellcolor{gray!18}$0.41$ & $0.27$ & $0.30$ & $0.30$ \\
 & G3-12b & $0.24$ & $0.13$ & \cellcolor{gray!18}$0.27$ & $0.25$ & $0.22$ & $0.38$ & \cellcolor{gray!18}$0.46$ & $0.45$ & $0.45$ & $0.38$ \\
\textit{Llama} & L3.2-3b & $-0.00$ & $0.04$ & \cellcolor{gray!18}$0.08$ & $-0.02$ & $-0.01$ & $0.30$ & \cellcolor{gray!18}$0.38$ & $0.29$ & $0.34$ & $0.30$ \\
 & L3.1-8b & $-0.01$ & $-0.00$ & \cellcolor{gray!18}$0.23$ & $-0.00$ & $-0.01$ & $0.28$ & $0.36$ & \cellcolor{gray!18}$0.41$ & $0.38$ & $0.28$ \\
\addlinespace[3pt]
\midrule
 & & \multicolumn{5}{c!{\hspace{3pt}\color{black!25}\vrule width 0.4pt\hspace{3pt}}}{\textbf{SummEval}\ \small($\bar R_h=0.60$)} & \multicolumn{5}{c}{\textbf{ConvAbuse}\ \small($\bar R_h=0.72$)} \\
\cmidrule[0.4pt]{1-12}
\textit{Gemini} & Gem-P & \cellcolor{passgreen}$0.64$ & \cellcolor{passgreen}$0.62$ & \cellcolor{bestgreen}$\mathbf{0.65}$ & \cellcolor{passgreen}$\mathbf{0.65}$ & \xmarkg & \cellcolor{passgreen}$0.74$ & \cellcolor{bestgreen}$\mathbf{0.75}$ & $0.73$ & \cellcolor{passgreen}$\mathbf{0.75}$ & \xmarkg \\
 & Gem-F & $0.58$ & $0.57$ & \cellcolor{bestgreen}$0.61$ & $0.59$ & \xmarkg & $0.72$ & \cellcolor{gray!18}$0.73$ & $0.71$ & $0.71$ & \xmarkg \\
 & Gem-L & $0.57$ & $0.53$ & $0.58$ & \cellcolor{gray!18}$0.58$ & \xmarkg & \cellcolor{bestgreen}$0.74$ & $0.72$ & $0.73$ & $0.73$ & \xmarkg \\
\textit{GPT} & GPT-T & \cellcolor{gray!18}$0.56$ & $0.51$ & $0.54$ & $0.55$ & \xmarkg & \cellcolor{gray!18}$0.71$ & $0.68$ & $0.68$ & $0.70$ & \xmarkg \\
 & GPT-L & \cellcolor{gray!18}$0.53$ & $0.50$ & $0.52$ & $0.53$ & \xmarkg & $0.73$ & $0.74$ & $0.71$ & \cellcolor{bestgreen}$\mathbf{0.75}$ & \xmarkg \\
\textit{Qwen} & Q3-8b & $0.46$ & $0.50$ & $0.51$ & \cellcolor{gray!18}$0.51$ & $0.46$ & $0.71$ & $0.70$ & $0.70$ & \cellcolor{gray!18}$0.73$ & $0.71$ \\
 & Q3-14b & $0.43$ & \cellcolor{gray!18}$0.56$ & $0.42$ & $0.48$ & $0.43$ & $0.60$ & $0.68$ & $0.65$ & $0.65$ & \cellcolor{gray!18}$0.70$ \\
\textit{Gemma} & G3-4b & $0.37$ & $0.27$ & \cellcolor{gray!18}$0.42$ & $0.34$ & $0.37$ & $0.24$ & \cellcolor{gray!18}$0.44$ & $0.41$ & $0.35$ & $0.25$ \\
 & G3-12b & $0.50$ & $0.50$ & $0.45$ & \cellcolor{gray!18}$0.56$ & $0.50$ & $0.61$ & \cellcolor{gray!18}$0.64$ & $0.61$ & $0.62$ & $0.57$ \\
\textit{Llama} & L3.2-3b & $0.33$ & $0.20$ & \cellcolor{gray!18}$0.40$ & $0.30$ & $0.33$ & $0.44$ & $0.51$ & $0.48$ & \cellcolor{gray!18}$0.53$ & $0.46$ \\
 & L3.1-8b & \cellcolor{gray!18}$0.50$ & $0.38$ & $0.28$ & $0.49$ & $0.50$ & $0.43$ & \cellcolor{gray!18}$0.61$ & $0.58$ & $0.58$ & $0.09$ \\
\addlinespace[3pt]
\midrule
 & & \multicolumn{5}{c!{\hspace{3pt}\color{black!25}\vrule width 0.4pt\hspace{3pt}}}{\textbf{CEBaB}\ \small($\bar R_h=0.72$)} & \multicolumn{5}{c}{\textbf{Loneliness}\ \small($\bar R_h=0.90$)} \\
\cmidrule[0.4pt]{1-12}
\textit{Gemini} & Gem-P & \cellcolor{passgreen}$0.77$ & \cellcolor{passgreen}$0.79$ & \cellcolor{bestgreen}$0.79$ & \cellcolor{passgreen}$0.79$ & \xmarkg & $0.57$ & \cellcolor{gray!18}$\mathbf{0.69}$ & $0.57$ & $0.59$ & \xmarkg \\
 & Gem-F & \cellcolor{passgreen}$0.73$ & \cellcolor{passgreen}$0.79$ & \cellcolor{passgreen}$0.79$ & \cellcolor{bestgreen}$0.79$ & \xmarkg & $0.56$ & \cellcolor{gray!18}$0.65$ & $0.61$ & $0.60$ & \xmarkg \\
 & Gem-L & \cellcolor{passgreen}$0.79$ & \cellcolor{bestgreen}$\mathbf{0.80}$ & \cellcolor{passgreen}$0.79$ & \cellcolor{passgreen}$\mathbf{0.80}$ & \xmarkg & $0.42$ & \cellcolor{gray!18}$0.58$ & $0.43$ & $0.47$ & \xmarkg \\
\textit{GPT} & GPT-T & \cellcolor{bestgreen}$\mathbf{0.80}$ & \cellcolor{passgreen}$0.79$ & \cellcolor{passgreen}$0.79$ & \cellcolor{passgreen}$0.79$ & \xmarkg & $0.32$ & \cellcolor{gray!18}$0.59$ & $0.38$ & $0.39$ & \xmarkg \\
 & GPT-L & \cellcolor{bestgreen}$0.79$ & \cellcolor{passgreen}$0.79$ & \cellcolor{passgreen}$0.79$ & \cellcolor{passgreen}$0.79$ & \xmarkg & $0.41$ & \cellcolor{gray!18}$0.61$ & $0.44$ & $0.46$ & \xmarkg \\
\textit{Qwen} & Q3-8b & \cellcolor{bestgreen}$0.78$ & \cellcolor{passgreen}$0.77$ & \cellcolor{passgreen}$0.78$ & \cellcolor{passgreen}$0.78$ & \cellcolor{passgreen}$0.78$ & $0.28$ & \cellcolor{gray!18}$0.46$ & $0.18$ & $0.38$ & $0.37$ \\
 & Q3-14b & \cellcolor{passgreen}$0.78$ & \cellcolor{passgreen}$0.78$ & \cellcolor{passgreen}$0.77$ & \cellcolor{passgreen}$0.78$ & \cellcolor{bestgreen}$0.79$ & $0.34$ & \cellcolor{gray!18}$0.47$ & $0.39$ & $0.43$ & $0.34$ \\
\textit{Gemma} & G3-4b & \cellcolor{passgreen}$0.78$ & \cellcolor{passgreen}$0.77$ & \cellcolor{bestgreen}$0.78$ & \cellcolor{passgreen}$0.78$ & \cellcolor{passgreen}$0.78$ & $0.17$ & $0.32$ & $0.22$ & $0.23$ & \cellcolor{gray!18}$0.52$ \\
 & G3-12b & \cellcolor{passgreen}$0.78$ & \cellcolor{passgreen}$0.79$ & \cellcolor{passgreen}$0.79$ & \cellcolor{bestgreen}$0.79$ & \cellcolor{passgreen}$0.78$ & $0.34$ & \cellcolor{gray!18}$0.50$ & $0.38$ & $0.48$ & $0.50$ \\
\textit{Llama} & L3.2-3b & \cellcolor{bestgreen}$0.79$ & \cellcolor{passgreen}$0.78$ & \cellcolor{passgreen}$0.78$ & \cellcolor{passgreen}$0.79$ & \cellcolor{passgreen}$0.79$ & $0.32$ & $0.39$ & $0.33$ & \cellcolor{gray!18}$0.39$ & $0.32$ \\
 & L3.1-8b & \cellcolor{passgreen}$0.78$ & \cellcolor{passgreen}$0.78$ & \cellcolor{bestgreen}$0.78$ & \cellcolor{passgreen}$0.78$ & \cellcolor{passgreen}$0.78$ & $0.29$ & $0.37$ & $0.22$ & $0.34$ & \cellcolor{gray!18}$0.58$ \\
\addlinespace[3pt]
\end{longtable}

\footnotesize\setlength{\tabcolsep}{3.5pt}\renewcommand{\arraystretch}{1.15}
\begin{longtable}{@{}ll!{\hspace{3pt}\color{black!25}\vrule width 0.4pt\hspace{3pt}}ccccc!{\hspace{3pt}\color{black!25}\vrule width 0.4pt\hspace{3pt}}ccccc@{}}
\caption{\textbf{\faImage\ Image.} Test outcomes and mixed-panel agreement $\bar R_m$ for every image judge and prompting technique. See the caption of Table~\ref{tab:full-text}.}\label{tab:full-image}\\
\toprule \textbf{Family} & \textbf{Judge} & \textbf{Zero} & \textbf{Few} & \textbf{CoT} & \textbf{Ens} & \textbf{GEPA} & \textbf{Zero} & \textbf{Few} & \textbf{CoT} & \textbf{Ens} & \textbf{GEPA} \\\midrule\endfirsthead
\multicolumn{12}{@{}l}{\emph{Image (continued)}}\\
\toprule \textbf{Family} & \textbf{Judge} & \textbf{Zero} & \textbf{Few} & \textbf{CoT} & \textbf{Ens} & \textbf{GEPA} & \textbf{Zero} & \textbf{Few} & \textbf{CoT} & \textbf{Ens} & \textbf{GEPA} \\\midrule\endhead
\bottomrule\endfoot
 & & \multicolumn{5}{c!{\hspace{3pt}\color{black!25}\vrule width 0.4pt\hspace{3pt}}}{\textbf{T2I-DPO}\ \small($\bar R_h=0.12$)} & \multicolumn{5}{c}{\textbf{Vibe-Arena}\ \small($\bar R_h=0.12$)} \\
\cmidrule[0.4pt]{1-12}
\textit{Gemini} & Gem-P & \cellcolor{passgreen}$\mathbf{0.15}$ & \cellcolor{passgreen}$\mathbf{0.15}$ & \cellcolor{bestgreen}$0.14$ & \cellcolor{passgreen}$\mathbf{0.15}$ & \xmarkg & \cellcolor{passgreen}$0.12$ & \cellcolor{passgreen}$0.11$ & \cellcolor{bestgreen}$0.13$ & \cellcolor{passgreen}$0.12$ & \xmarkg \\
 & Gem-F & \cellcolor{bestgreen}$\mathbf{0.15}$ & \cellcolor{passgreen}$0.14$ & \cellcolor{passgreen}$\mathbf{0.15}$ & \cellcolor{passgreen}$0.14$ & \xmarkg & \cellcolor{bestgreen}$0.13$ & \cellcolor{passgreen}$0.11$ & \cellcolor{passgreen}$0.11$ & \cellcolor{passgreen}$0.12$ & \xmarkg \\
 & Gem-L & \cellcolor{passgreen}$0.13$ & \cellcolor{passgreen}$0.12$ & \cellcolor{bestgreen}$0.13$ & \cellcolor{passgreen}$0.12$ & \xmarkg & \cellcolor{passgreen}$\mathbf{0.14}$ & \cellcolor{passgreen}$0.12$ & \cellcolor{passgreen}$0.12$ & \cellcolor{bestgreen}$\mathbf{0.14}$ & \xmarkg \\
\textit{GPT} & GPT-T & \cellcolor{passgreen}$0.12$ & \cellcolor{passgreen}$0.13$ & \cellcolor{bestgreen}$0.14$ & \cellcolor{passgreen}$0.13$ & \xmarkg & \cellcolor{passgreen}$\mathbf{0.14}$ & \cellcolor{passgreen}$0.12$ & \cellcolor{bestgreen}$\mathbf{0.14}$ & \cellcolor{passgreen}$0.13$ & \xmarkg \\
 & GPT-L & \cellcolor{passgreen}$0.11$ & \cellcolor{passgreen}$0.11$ & \cellcolor{bestgreen}$0.12$ & \cellcolor{passgreen}$0.11$ & \xmarkg & \cellcolor{bestgreen}$0.11$ & $0.08$ & \cellcolor{passgreen}$0.11$ & \cellcolor{passgreen}$0.11$ & \xmarkg \\
\textit{Qwen} & Q2.5V-3b & $-0.00$ & \cellcolor{gray!18}$0.08$ & $0.00$ & $0.00$ & $-0.00$ & $-0.05$ & \cellcolor{gray!18}$0.01$ & $-0.05$ & $-0.05$ & $-0.06$ \\
 & Q2.5V-7b & $0.03$ & \cellcolor{bestgreen}$0.11$ & $0.07$ & $0.06$ & $0.04$ & \cellcolor{bestgreen}$0.11$ & $0.06$ & $0.05$ & $0.08$ & $0.08$ \\
\textit{Gemma} & G3-4b & $0.06$ & \cellcolor{gray!18}$0.08$ & $0.01$ & $0.05$ & $-0.07$ & $-0.06$ & \cellcolor{gray!18}$0.07$ & $-0.01$ & $-0.03$ & $-0.03$ \\
 & G3-12b & \cellcolor{bestgreen}$0.11$ & $0.08$ & $0.04$ & \cellcolor{passgreen}$0.10$ & \cellcolor{passgreen}$0.11$ & $0.06$ & \cellcolor{bestgreen}$0.10$ & $-0.00$ & $0.06$ & $0.05$ \\
\textit{Llama} & L3.2V-11b & $-0.07$ & $-0.06$ & $0.02$ & $-0.07$ & \cellcolor{gray!18}$0.02$ & $-0.04$ & $-0.04$ & \cellcolor{gray!18}$-0.01$ & $-0.06$ & $-0.06$ \\
\addlinespace[3pt]
\midrule
 & & \multicolumn{5}{c!{\hspace{3pt}\color{black!25}\vrule width 0.4pt\hspace{3pt}}}{\textbf{FER+}\ \small($\bar R_h=0.51$)} & \multicolumn{5}{c}{\textbf{Flickr8k-Expert}\ \small($\bar R_h=0.87$)} \\
\cmidrule[0.4pt]{1-12}
\textit{Gemini} & Gem-P & \cellcolor{passgreen}$0.54$ & \cellcolor{bestgreen}$\mathbf{0.55}$ & \cellcolor{passgreen}$0.54$ & \cellcolor{passgreen}$\mathbf{0.55}$ & \xmarkg & $0.68$ & $0.73$ & \cellcolor{gray!18}$0.82$ & $0.73$ & \xmarkg \\
 & Gem-F & \cellcolor{passgreen}$0.52$ & \cellcolor{bestgreen}$0.53$ & \cellcolor{passgreen}$0.52$ & \cellcolor{passgreen}$0.53$ & \xmarkg & $0.85$ & $0.85$ & $0.85$ & \cellcolor{gray!18}$0.85$ & \xmarkg \\
 & Gem-L & $0.48$ & \cellcolor{bestgreen}$0.53$ & $0.46$ & $0.49$ & \xmarkg & $0.81$ & \cellcolor{gray!18}$0.83$ & $0.82$ & $0.82$ & \xmarkg \\
\textit{GPT} & GPT-T & \cellcolor{gray!18}$0.50$ & $0.46$ & $0.49$ & $0.49$ & \xmarkg & $0.83$ & \cellcolor{gray!18}$\mathbf{0.86}$ & $0.83$ & $0.84$ & \xmarkg \\
 & GPT-L & \cellcolor{bestgreen}$0.51$ & $0.48$ & $0.49$ & \cellcolor{passgreen}$0.51$ & \xmarkg & $0.79$ & \cellcolor{gray!18}$0.84$ & $0.83$ & $0.82$ & \xmarkg \\
\textit{Qwen} & Q2.5V-3b & \cellcolor{gray!18}$0.32$ & $0.24$ & $0.28$ & $0.31$ & $0.32$ & $0.51$ & \cellcolor{gray!18}$0.73$ & $0.65$ & $0.72$ & $0.51$ \\
 & Q2.5V-7b & $0.37$ & $0.32$ & $0.31$ & \cellcolor{gray!18}$0.37$ & $0.37$ & $0.66$ & \cellcolor{gray!18}$0.79$ & $0.66$ & $0.76$ & $0.72$ \\
\textit{Gemma} & G3-4b & $0.38$ & $0.37$ & $0.36$ & \cellcolor{gray!18}$0.39$ & $0.38$ & \cellcolor{gray!18}$0.82$ & $0.74$ & $0.80$ & $0.77$ & $0.82$ \\
 & G3-12b & $0.42$ & $0.39$ & $0.38$ & $0.41$ & \cellcolor{gray!18}$0.43$ & $0.80$ & \cellcolor{gray!18}$0.83$ & $0.76$ & $0.81$ & $0.82$ \\
\textit{Llama} & L3.2V-11b & $0.41$ & $0.21$ & $0.42$ & $0.40$ & \cellcolor{gray!18}$0.44$ & $0.75$ & $0.71$ & $0.64$ & \cellcolor{gray!18}$0.79$ & $0.72$ \\
\addlinespace[3pt]
\midrule
 & & \multicolumn{5}{c!{\hspace{3pt}\color{black!25}\vrule width 0.4pt\hspace{3pt}}}{\textbf{CIFAR-10H}\ \small($\bar R_h=0.88$)} & \multicolumn{5}{c}{} \\
\cmidrule[0.4pt]{1-12}
\textit{Gemini} & Gem-P & \cellcolor{passgreen}$0.89$ & \cellcolor{bestgreen}$0.89$ & $0.89$ & \cellcolor{passgreen}$0.89$ & \xmarkg &  &  &  &  &  \\
 & Gem-F & \cellcolor{passgreen}$\mathbf{0.90}$ & \cellcolor{bestgreen}$\mathbf{0.90}$ & \cellcolor{passgreen}$\mathbf{0.90}$ & \cellcolor{passgreen}$\mathbf{0.90}$ & \xmarkg &  &  &  &  &  \\
 & Gem-L & $0.88$ & \cellcolor{bestgreen}$0.89$ & $0.88$ & $0.88$ & \xmarkg &  &  &  &  &  \\
\textit{GPT} & GPT-T & $0.70$ & \cellcolor{gray!18}$0.74$ & $0.74$ & $0.73$ & \xmarkg &  &  &  &  &  \\
 & GPT-L & $0.67$ & \cellcolor{gray!18}$0.74$ & $0.65$ & $0.70$ & \xmarkg &  &  &  &  &  \\
\textit{Qwen} & Q2.5V-3b & $0.80$ & $0.78$ & $0.76$ & \cellcolor{gray!18}$0.81$ & $0.80$ &  &  &  &  &  \\
 & Q2.5V-7b & $0.82$ & \cellcolor{gray!18}$0.83$ & $0.79$ & $0.83$ & $0.82$ &  &  &  &  &  \\
\textit{Gemma} & G3-4b & $0.81$ & $0.78$ & $0.81$ & $0.82$ & \cellcolor{gray!18}$0.84$ &  &  &  &  &  \\
 & G3-12b & \cellcolor{gray!18}$0.83$ & $0.77$ & $0.82$ & $0.83$ & $0.81$ &  &  &  &  &  \\
\textit{Llama} & L3.2V-11b & \cellcolor{gray!18}$0.81$ & $0.51$ & $0.80$ & $0.80$ & $0.79$ &  &  &  &  &  \\
\addlinespace[3pt]
\end{longtable}

\footnotesize\setlength{\tabcolsep}{3.5pt}\renewcommand{\arraystretch}{1.15}
\begin{longtable}{@{}ll!{\hspace{3pt}\color{black!25}\vrule width 0.4pt\hspace{3pt}}ccccc!{\hspace{3pt}\color{black!25}\vrule width 0.4pt\hspace{3pt}}ccccc@{}}
\caption{\textbf{\faVolumeUp\ Audio.} Test outcomes and mixed-panel agreement $\bar R_m$ for every audio judge and prompting technique. See the caption of Table~\ref{tab:full-text}.}\label{tab:full-audio}\\
\toprule \textbf{Family} & \textbf{Judge} & \textbf{Zero} & \textbf{Few} & \textbf{CoT} & \textbf{Ens} & \textbf{GEPA} & \textbf{Zero} & \textbf{Few} & \textbf{CoT} & \textbf{Ens} & \textbf{GEPA} \\\midrule\endfirsthead
\multicolumn{12}{@{}l}{\emph{Audio (continued)}}\\
\toprule \textbf{Family} & \textbf{Judge} & \textbf{Zero} & \textbf{Few} & \textbf{CoT} & \textbf{Ens} & \textbf{GEPA} & \textbf{Zero} & \textbf{Few} & \textbf{CoT} & \textbf{Ens} & \textbf{GEPA} \\\midrule\endhead
\bottomrule\endfoot
 & & \multicolumn{5}{c!{\hspace{3pt}\color{black!25}\vrule width 0.4pt\hspace{3pt}}}{\textbf{TTS-Preference}\ \small($\bar R_h=0.01$)} & \multicolumn{5}{c}{\textbf{SOMOS}\ \small($\bar R_h=0.05$)} \\
\cmidrule[0.4pt]{1-12}
\textit{Gemini} & Gem-P & \cellcolor{passgreen}$0.01$ & \cellcolor{passgreen}$0.00$ & \cellcolor{passgreen}$0.02$ & \cellcolor{bestgreen}$\mathbf{0.03}$ & \xmarkg & $-0.14$ & \cellcolor{bestgreen}$0.03$ & $-0.12$ & $-0.10$ & \xmarkg \\
 & Gem-F & $-0.02$ & $-0.03$ & \cellcolor{bestgreen}$0.01$ & \cellcolor{passgreen}$0.01$ & \xmarkg & \cellcolor{bestgreen}$\mathbf{0.06}$ & \cellcolor{passgreen}$0.04$ & \cellcolor{passgreen}$0.04$ & \cellcolor{passgreen}$0.03$ & \xmarkg \\
 & Gem-L & $-0.09$ & $-0.03$ & \cellcolor{passgreen}$-0.00$ & \cellcolor{bestgreen}$0.02$ & \xmarkg & $-0.10$ & $-0.04$ & \cellcolor{bestgreen}$0.05$ & $-0.06$ & \xmarkg \\
\textit{GPT} & GPT-A & \cellcolor{bestgreen}$0.02$ & $-0.04$ & $-0.03$ & $-0.05$ & \xmarkg & \cellcolor{passgreen}$0.04$ & \cellcolor{passgreen}$0.04$ & \cellcolor{bestgreen}$0.05$ & \cellcolor{passgreen}$0.04$ & \xmarkg \\
 & GPT-A-M & \cellcolor{passgreen}$-0.01$ & \cellcolor{passgreen}$-0.00$ & \cellcolor{passgreen}$-0.00$ & \cellcolor{bestgreen}$-0.00$ & \xmarkg & \cellcolor{passgreen}$0.01$ & \cellcolor{bestgreen}$0.05$ & \cellcolor{passgreen}$0.04$ & \cellcolor{passgreen}$0.04$ & \xmarkg \\
\textit{Qwen} & Q2.5-O-3b & $-0.09$ & $-0.07$ & \cellcolor{bestgreen}$0.01$ & \cellcolor{passgreen}$0.01$ & $-0.09$ & \cellcolor{passgreen}$0.03$ & \cellcolor{passgreen}$0.04$ & \cellcolor{passgreen}$0.03$ & \cellcolor{bestgreen}$0.05$ & \cellcolor{passgreen}$0.03$ \\
 & Q2-A-7b & $-0.12$ & \cellcolor{gray!18}$-0.04$ & $-0.12$ & $-0.12$ & $-0.12$ & $-0.00$ & $0.01$ & $-0.00$ & \cellcolor{bestgreen}$0.02$ & $-0.00$ \\
\addlinespace[3pt]
\end{longtable}

\footnotesize\setlength{\tabcolsep}{3.5pt}\renewcommand{\arraystretch}{1.15}
\begin{longtable}{@{}ll!{\hspace{3pt}\color{black!25}\vrule width 0.4pt\hspace{3pt}}cccc@{}}
\caption{\textbf{\faVideo\ Video.} Test outcomes and mixed-panel agreement $\bar R_m$ for every video judge and prompting technique. See the caption of Table~\ref{tab:full-text}.}\label{tab:full-video}\\
\toprule \textbf{Family} & \textbf{Judge} & \textbf{Zero} & \textbf{Few} & \textbf{CoT} & \textbf{Ens} \\\midrule\endfirsthead
\multicolumn{6}{@{}l}{\emph{Video (continued)}}\\
\toprule \textbf{Family} & \textbf{Judge} & \textbf{Zero} & \textbf{Few} & \textbf{CoT} & \textbf{Ens} \\\midrule\endhead
\bottomrule\endfoot
 & & \multicolumn{4}{c}{\textbf{T2V-Preference}\ \small($\bar R_h=0.04$)} \\
\cmidrule[0.4pt]{1-6}
\textit{Gemini} & Gem-P & \cellcolor{passgreen}$0.07$ & \cellcolor{passgreen}$0.07$ & \cellcolor{bestgreen}$\mathbf{0.08}$ & \cellcolor{passgreen}$\mathbf{0.08}$ \\
 & Gem-F & \cellcolor{passgreen}$\mathbf{0.08}$ & \cellcolor{passgreen}$\mathbf{0.08}$ & \cellcolor{bestgreen}$\mathbf{0.08}$ & \cellcolor{passgreen}$\mathbf{0.08}$ \\
 & Gem-L & \cellcolor{passgreen}$0.06$ & \cellcolor{passgreen}$0.06$ & \cellcolor{passgreen}$0.06$ & \cellcolor{bestgreen}$0.07$ \\
\textit{Qwen} & Q2.5V-3b & \cellcolor{passgreen}$0.03$ & $-0.00$ & $-0.01$ & \cellcolor{bestgreen}$0.03$ \\
 & Q2.5V-7b & \cellcolor{passgreen}$\mathbf{0.08}$ & \cellcolor{bestgreen}$\mathbf{0.08}$ & \cellcolor{passgreen}$0.07$ & \cellcolor{passgreen}$\mathbf{0.08}$ \\
\addlinespace[3pt]
\end{longtable}

\subsection{Comparison with Existing Tests}
\label{app:real-comparison}

In the simulations, we compare the \anlt{} with the \alttest{} and the STT. In real-world data, these tests can be applied only to datasets in which the same annotators label many items. Nine of our datasets satisfy the requirements of the \alttest{}, and only DICES-350 satisfies the requirements of the STT. Table~\ref{tab:realcomp-ni} compares the tests on these datasets. Every LLM uses its best prompting technique, and for every dataset, all tests use the same items. For this comparison, we use $B=300$ bootstrap iterations. Since real-world data have no gold labels, we cannot tell which test is correct, and we only examine whether the tests agree.

The \anlt{} and the \alttest{} (with $\varepsilon=0.1$) give the same result in $68$ of the $94$ cases, and they accept a similar number of LLMs, $44$ and $42$. For example, both tests accept every LLM on CEBaB and none on Loneliness. The disagreements do not go in one direction. The \anlt{} is stricter on four datasets and more permissive on three. The largest differences are on EPIC, in which the \anlt{} accepts $9$ LLMs and the \alttest{} accepts none, and on DICES-350, in which the \anlt{} accepts $4$ and the \alttest{} accepts $9$.

On DICES-350, the STT accepts only two LLMs. As an equivalence test, it does not accept an LLM that increases the agreement. \textsc{STT-Alt}, which tests non-inferiority, accepts one more LLM.

\begingroup\centering\tiny
\setlength{\tabcolsep}{1pt}\renewcommand{\arraystretch}{1.12}
\begin{longtable}{@{\extracolsep{\fill}}lccccccccccc@{}}
\caption{\textbf{Outcomes of the three tests on datasets with shared annotators.} For each dataset and judge, the table reports the \alttest{} winning rate $\omega$ at $\varepsilon\!=\!0.1$, the \anlt{} decision statistic $q_\alpha^*$, the STT $p_{\mathrm{TOST}}$, and the STT-Alt $p_{\mathrm{lower}}$ at $\mathrm{frac}\!=\!0.5$, using each judge's best prompting technique. For each dataset, all tests are applied to the items that satisfy the requirements of every test. Green marks a passing cell. Because each test depends on its own hyperparameter, the comparison should not be read as strictly apples-to-apples.}\label{tab:realcomp-ni}\\
\toprule
& \multicolumn{3}{c}{\textbf{Gemini}} & \multicolumn{2}{c}{\textbf{GPT}} & \multicolumn{2}{c}{\textbf{Qwen}} & \multicolumn{2}{c}{\textbf{Gemma}} & \multicolumn{2}{c}{\textbf{Llama}} \\
\midrule
\endfirsthead
\multicolumn{12}{@{}l}{\emph{continued}}\\
\toprule
& \multicolumn{3}{c}{\textbf{Gemini}} & \multicolumn{2}{c}{\textbf{GPT}} & \multicolumn{2}{c}{\textbf{Qwen}} & \multicolumn{2}{c}{\textbf{Gemma}} & \multicolumn{2}{c}{\textbf{Llama}} \\
\midrule
\endhead
\bottomrule
\endfoot
\textbf{\faFile*\ Text} & \textbf{Gem-P} & \textbf{Gem-F} & \textbf{Gem-L} & \textbf{GPT-T} & \textbf{GPT-L} & \textbf{Q3-8b} & \textbf{Q3-14b} & \textbf{G3-4b} & \textbf{G3-12b} & \textbf{L3.2-3b} & \textbf{L3.1-8b} \\
\midrule
\multicolumn{12}{@{}l}{\faFile*~\textbf{DICES-350}\ ($\bar R_h=0.20$)}\\[1pt]
\hspace{3pt}\alttest{}\ $\omega$ & \cellcolor{passgreen}$0.89$ & \cellcolor{passgreen}$0.81$ & \cellcolor{passgreen}$0.78$ & \cellcolor{passgreen}$0.88$ & \cellcolor{passgreen}$0.87$ & $0.12$ & \cellcolor{passgreen}$0.70$ & $0.49$ & \cellcolor{passgreen}$0.98$ & \cellcolor{passgreen}$0.87$ & \cellcolor{passgreen}$0.80$ \\
\hspace{3pt}\anlt{}\ $q^*_\alpha$ & \cellcolor{passgreen}$+0.03$ & $-0.07$ & $-0.06$ & \cellcolor{passgreen}$+0.01$ & \cellcolor{passgreen}$+0.01$ & $-0.09$ & $-0.02$ & $-0.06$ & \cellcolor{passgreen}$+0.04$ & $-0.03$ & $-0.02$ \\
\hspace{3pt}STT\ $p_{\mathrm{TOST}}$ & \cellcolor{passgreen}$0.01$ & $1.00$ & $1.00$ & $0.15$ & \cellcolor{passgreen}$0.02$ & $1.00$ & $0.94$ & $1.00$ & $0.08$ & $1.00$ & $1.00$ \\
\hspace{3pt}STT-Alt\ $p_{\mathrm{lower}}$ & \cellcolor{passgreen}$<0.01$ & $1.00$ & $1.00$ & $0.15$ & \cellcolor{passgreen}$0.02$ & $1.00$ & $0.94$ & $1.00$ & \cellcolor{passgreen}$<0.01$ & $1.00$ & $1.00$ \\
\addlinespace[4pt]
\multicolumn{12}{@{}l}{\faFile*~\textbf{EPIC}\ ($\bar R_h=0.20$)}\\[1pt]
\hspace{3pt}\alttest{}\ $\omega$ & $0.42$ & $0.26$ & $0.11$ & $0.42$ & $0.37$ & $0.11$ & $0.21$ & $0.00$ & $0.05$ & $0.05$ & $0.00$ \\
\hspace{3pt}\anlt{}\ $q^*_\alpha$ & \cellcolor{passgreen}$+0.05$ & \cellcolor{passgreen}$+0.01$ & \cellcolor{passgreen}$+0.06$ & \cellcolor{passgreen}$+0.05$ & \cellcolor{passgreen}$+0.10$ & \cellcolor{passgreen}$+0.08$ & \cellcolor{passgreen}$+0.07$ & $-0.09$ & \cellcolor{passgreen}$+0.06$ & \cellcolor{passgreen}$+0.01$ & $-0.10$ \\
\addlinespace[4pt]
\multicolumn{12}{@{}l}{\faFile*~\textbf{10k-Prompts}\ ($\bar R_h=0.37$)}\\[1pt]
\hspace{3pt}\alttest{}\ $\omega$ & \cellcolor{passgreen}$0.56$ & $0.11$ & \cellcolor{passgreen}$0.67$ & $0.00$ & \cellcolor{passgreen}$0.67$ & \cellcolor{passgreen}$0.78$ & $0.33$ & \cellcolor{passgreen}$0.67$ & $0.33$ & $0.44$ & $0.00$ \\
\hspace{3pt}\anlt{}\ $q^*_\alpha$ & \cellcolor{passgreen}$+0.07$ & \cellcolor{passgreen}$+0.05$ & \cellcolor{passgreen}$+0.09$ & \cellcolor{passgreen}$+0.01$ & \cellcolor{passgreen}$+0.03$ & \cellcolor{passgreen}$+0.05$ & \cellcolor{passgreen}$+0.05$ & $-0.01$ & \cellcolor{passgreen}$+0.02$ & $-0.04$ & $-0.05$ \\
\addlinespace[4pt]
\multicolumn{12}{@{}l}{\faFile*~\textbf{SummEval}\ ($\bar R_h=0.60$)}\\[1pt]
\hspace{3pt}\alttest{}\ $\omega$ & \cellcolor{passgreen}$1.00$ & \cellcolor{passgreen}$0.67$ & $0.33$ & $0.33$ & $0.33$ & \cellcolor{passgreen}$1.00$ & \cellcolor{passgreen}$1.00$ & $0.00$ & \cellcolor{passgreen}$1.00$ & $0.00$ & \cellcolor{passgreen}$0.67$ \\
\hspace{3pt}\anlt{}\ $q^*_\alpha$ & \cellcolor{passgreen}$+0.05$ & \cellcolor{passgreen}$+0.01$ & $-0.02$ & $-0.05$ & $-0.07$ & $-0.09$ & $-0.04$ & $-0.19$ & $-0.04$ & $-0.21$ & $-0.10$ \\
\addlinespace[4pt]
\multicolumn{12}{@{}l}{\faFile*~\textbf{ConvAbuse}\ ($\bar R_h=0.71$)}\\[1pt]
\hspace{3pt}\alttest{}\ $\omega$ & \cellcolor{passgreen}$0.75$ & $0.25$ & \cellcolor{passgreen}$0.50$ & $0.25$ & \cellcolor{passgreen}$0.50$ & \cellcolor{passgreen}$0.50$ & $0.12$ & $0.00$ & $0.00$ & $0.00$ & $0.00$ \\
\hspace{3pt}\anlt{}\ $q^*_\alpha$ & \cellcolor{passgreen}$+0.01$ & $-0.01$ & \cellcolor{passgreen}$+0.00$ & $-0.02$ & \cellcolor{passgreen}$+0.02$ & $-0.00$ & $-0.04$ & $-0.32$ & $-0.10$ & $-0.23$ & $-0.14$ \\
\addlinespace[4pt]
\multicolumn{12}{@{}l}{\faFile*~\textbf{CEBaB}\ ($\bar R_h=0.72$)}\\[1pt]
\hspace{3pt}\alttest{}\ $\omega$ & \cellcolor{passgreen}$0.60$ & \cellcolor{passgreen}$0.90$ & \cellcolor{passgreen}$1.00$ & \cellcolor{passgreen}$0.50$ & \cellcolor{passgreen}$0.80$ & \cellcolor{passgreen}$0.50$ & \cellcolor{passgreen}$0.60$ & \cellcolor{passgreen}$0.70$ & \cellcolor{passgreen}$0.90$ & \cellcolor{passgreen}$0.50$ & \cellcolor{passgreen}$0.50$ \\
\hspace{3pt}\anlt{}\ $q^*_\alpha$ & \cellcolor{passgreen}$+0.07$ & \cellcolor{passgreen}$+0.07$ & \cellcolor{passgreen}$+0.07$ & \cellcolor{passgreen}$+0.07$ & \cellcolor{passgreen}$+0.07$ & \cellcolor{passgreen}$+0.06$ & \cellcolor{passgreen}$+0.06$ & \cellcolor{passgreen}$+0.06$ & \cellcolor{passgreen}$+0.07$ & \cellcolor{passgreen}$+0.06$ & \cellcolor{passgreen}$+0.06$ \\
\addlinespace[4pt]
\multicolumn{12}{@{}l}{\faFile*~\textbf{Loneliness}\ ($\bar R_h=0.90$)}\\[1pt]
\hspace{3pt}\alttest{}\ $\omega$ & $0.00$ & $0.00$ & $0.00$ & $0.00$ & $0.00$ & $0.00$ & $0.00$ & $0.00$ & $0.00$ & $0.00$ & $0.00$ \\
\hspace{3pt}\anlt{}\ $q^*_\alpha$ & $-0.24$ & $-0.28$ & $-0.35$ & $-0.35$ & $-0.34$ & $-0.49$ & $-0.49$ & $-0.43$ & $-0.45$ & $-0.56$ & $-0.37$ \\
\addlinespace[4pt]
\addlinespace[5pt]
\textbf{\faImage\ Image} & \textbf{Gem-P} & \textbf{Gem-F} & \textbf{Gem-L} & \textbf{GPT-T} & \textbf{GPT-L} & \textbf{Q2.5V-3b} & \textbf{Q2.5V-7b} & \textbf{G3-4b} & \textbf{G3-12b} & \textbf{L3.2V-11b} &  \\
\midrule
\multicolumn{12}{@{}l}{\faImage~\textbf{Flickr8k-Expert}\ ($\bar R_h=0.87$)}\\[1pt]
\hspace{3pt}\alttest{}\ $\omega$ & $0.00$ & $0.33$ & $0.00$ & \cellcolor{passgreen}$0.67$ & $0.00$ & $0.00$ & $0.00$ & $0.00$ & $0.00$ & $0.00$ &  \\
\hspace{3pt}\anlt{}\ $q^*_\alpha$ & $-0.14$ & $-0.01$ & $-0.04$ & $-0.01$ & $-0.04$ & $-0.15$ & $-0.09$ & $-0.06$ & $-0.05$ & $-0.09$ &  \\
\addlinespace[4pt]
\addlinespace[5pt]
\textbf{\faVolumeUp\ Audio} & \textbf{Gem-P} & \textbf{Gem-F} & \textbf{Gem-L} & \textbf{GPT-A} & \textbf{GPT-A-M} & \textbf{Q2.5-O-3b} & \textbf{Q2-A-7b} &  &  &  &  \\
\midrule
\multicolumn{12}{@{}l}{\faVolumeUp~\textbf{SOMOS}\ ($\bar R_h=0.05$)}\\[1pt]
\hspace{3pt}\alttest{}\ $\omega$ & \cellcolor{passgreen}$0.69$ & \cellcolor{passgreen}$0.58$ & $0.12$ & \cellcolor{passgreen}$1.00$ & \cellcolor{passgreen}$0.85$ & \cellcolor{passgreen}$0.92$ & \cellcolor{passgreen}$0.96$ &  &  &  &  \\
\hspace{3pt}\anlt{}\ $q^*_\alpha$ & \cellcolor{passgreen}$+0.01$ & \cellcolor{passgreen}$+0.04$ & \cellcolor{passgreen}$+0.04$ & \cellcolor{passgreen}$+0.03$ & \cellcolor{passgreen}$+0.04$ & \cellcolor{passgreen}$+0.03$ & \cellcolor{passgreen}$+0.01$ &  &  &  &  \\
\addlinespace[4pt]
\end{longtable}
\endgroup

\subsection{Number of Items and Panel Size}
\label{app:real-items}

\begin{figure}[t]\centering
\includegraphics[width=\textwidth]{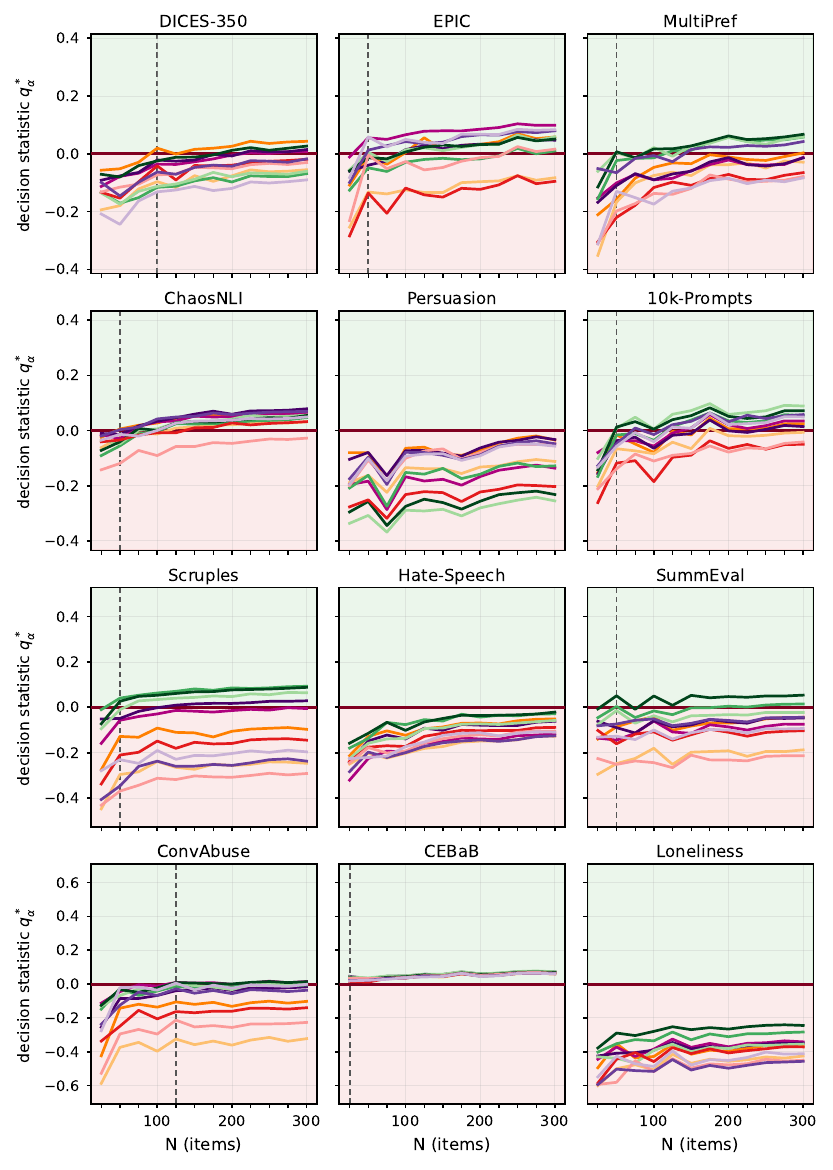}\\
\includegraphics[width=0.75\textwidth]{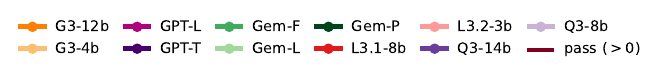}
\caption{Varying the sample size $N$ for real-world text datasets. The Y-axis of each plot presents the decision statistic $q^*_\alpha$ against the number of evaluated items $N$, with one curve per LLM, using each judge's best prompting technique. When $q^*_\alpha>0$, the LLM passes the test. The dashed line marks the smallest $N$ at which any judge first passes the test.}
\label{fig:app-nsweep}
\end{figure}

\begin{figure}[t]\centering
\includegraphics[width=\textwidth]{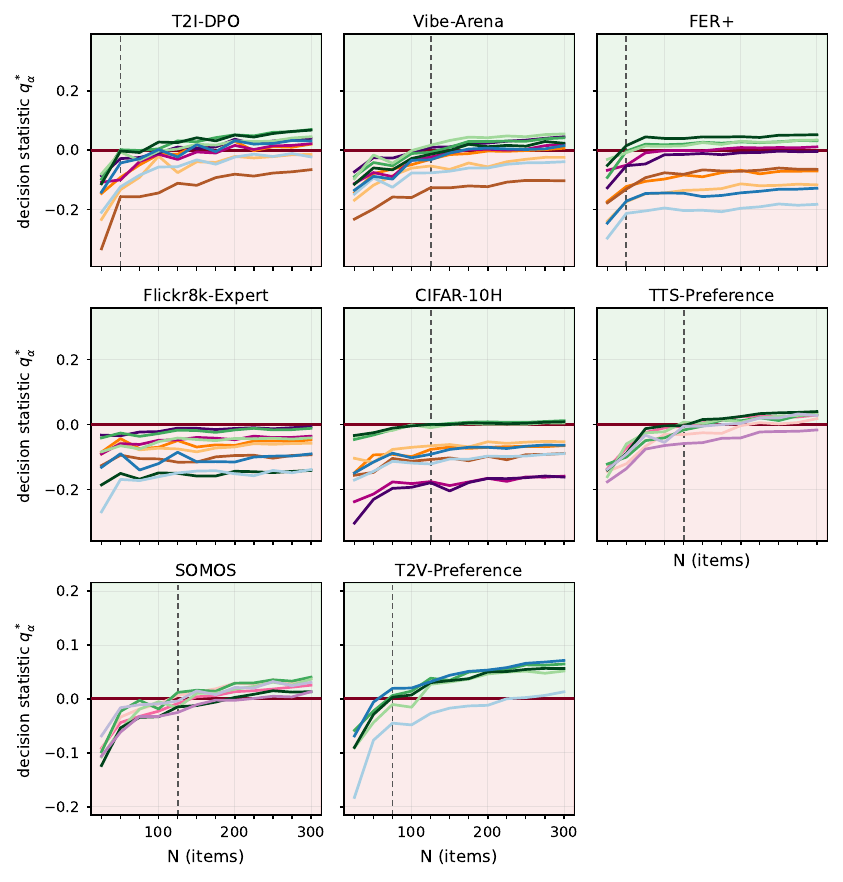}\\[1pt]
\includegraphics[width=0.85\textwidth]{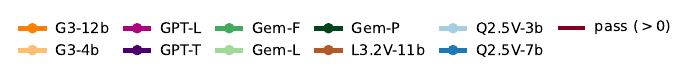}
\\[6pt]

\includegraphics[width=0.85\textwidth]{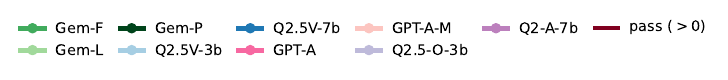}
\caption{Varying the sample size $N$ for real-world image, audio, and video datasets. See the caption of Figure~\ref{fig:app-nsweep}.}
\label{fig:app-nsweep-b}
\end{figure}

\paragraph{Number of items.}
In Section~\ref{sec:exp}, we report the minimum number of items with which the LLMs pass the test. Here, we present the full analysis. Figures~\ref{fig:app-nsweep} and~\ref{fig:app-nsweep-b} present the decision statistic $q^*_\alpha$ as a function of the number of items $N$, for all $20$ datasets and for every LLM with its best prompting technique.

The result of the test stabilizes with fewer items than the value of the statistic. LLMs that pass with $300$ items usually already pass with about $100$ items (Table~\ref{tab:method-passrate}). The value of $q^*_\alpha$ continues to change with more items, and the number of items it requires depends on the dataset. It stabilizes earlier in datasets with high human agreement and later in datasets with low agreement. Between $N=200$ and $N=300$, it changes by at most $0.05$ for every LLM. We therefore recommend using at least $100$ items, in line with the simulations in Section~\ref{sec:sim}. More items mainly make the value of $q^*_\alpha$ more precise, which is useful for comparing LLMs.

\paragraph{Panel size.}
In all the experiments, we use panels of $k=3$ annotations. A larger panel includes more annotations per item, which may be important in datasets with low human agreement. We therefore examine whether the results change with the panel size, and vary $k\in\{3,5,7,10\}$ on six such datasets, which are DICES-350, ChaosNLI, Scruples, Vibe-Arena, T2V-Preference, and TTS-Preference. Every LLM uses its best prompting technique, and the margin is scaled by $k_c/k$. Of the $55$ results, $49$ are identical with $k=10$ and with $k=3$. In the other six, the LLM fails with $k=3$ and passes with $k=10$, and in all of them the decision statistic is close to zero (at most $0.01$ with $k=10$). Three of these cases are on DICES-350, and the others are on Scruples, Vibe-Arena, and TTS-Preference. The results are therefore stable across panel sizes, and $k=3$, which requires the fewest annotations, is sufficient in practice.


\subsection{Prompting}
\label{app:real-prompting}

\begin{figure}[t]\centering
\includegraphics[width=\textwidth]{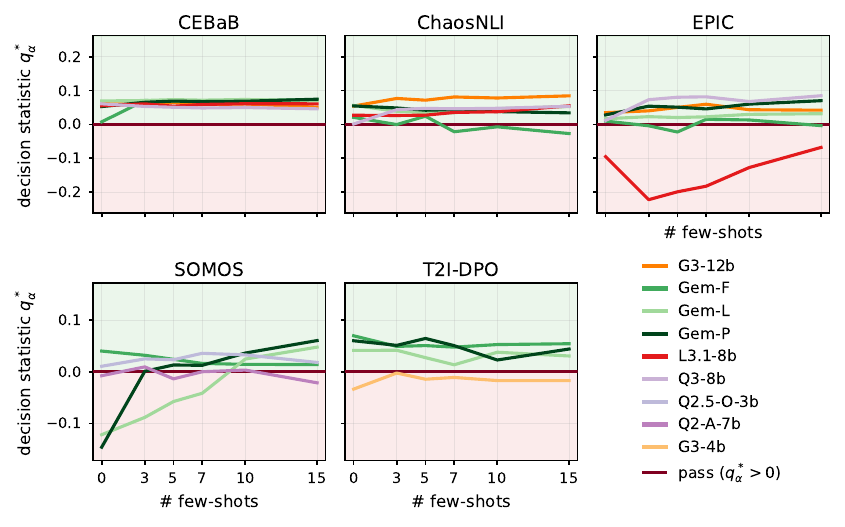}
\caption{Evaluating the impact of the number of random few-shot examples (X-axis) on the decision statistic $q^*_\alpha$. As shown, the effect is mixed and depends on both the model and the dataset.}
\label{fig:app-ksweep}
\end{figure}

\paragraph{Number of few-shot examples.}
In the main experiments, the few-shot prompt includes five examples. Figure~\ref{fig:app-ksweep} examines this choice. It presents $q^*_\alpha$ when the number of examples varies from zero to fifteen, on five datasets. In most cases, the statistic changes little beyond three to five examples. There are exceptions. EPIC is more sensitive to the number of examples, and on SOMOS, two of the Gemini models continue to improve up to fifteen examples. Overall, only four results change beyond five examples. We therefore use five examples, which gives most of the benefit with a shorter prompt.

\begin{figure}[t]\centering
\includegraphics[width=\textwidth]{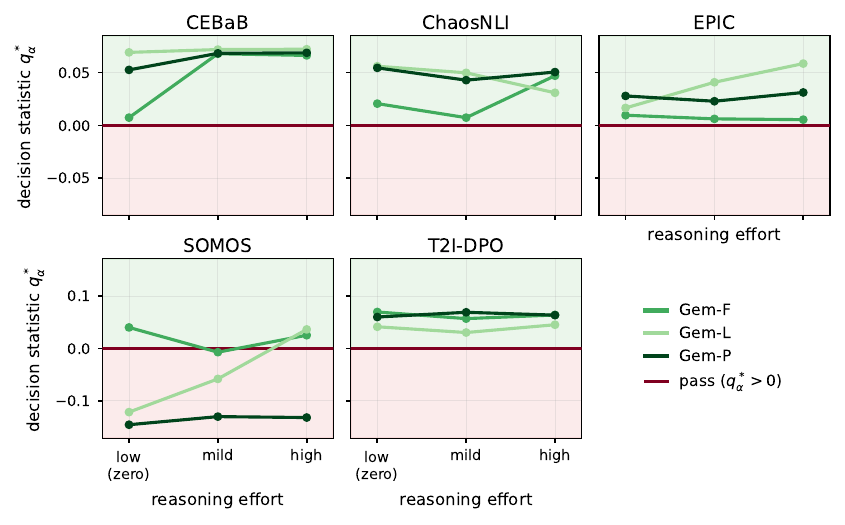}
\caption{Evaluating the impact of reasoning effort in CoT prompting (X-axis) on the decision statistic $q^*_\alpha$. Reasoning effort is controlled by the Gemini API thinking-effort parameter. As shown, the effect is mixed, but higher effort generally leads to better judges.}
\label{fig:app-cotsweep}
\end{figure}

\paragraph{Reasoning effort.}
In the main experiments, CoT prompting uses a mild reasoning effort. Figure~\ref{fig:app-cotsweep} examines this choice for the three Gemini models, for which the API controls the reasoning effort. On average, more reasoning helps, and the mean of $q^*_\alpha$ increases from $0.02$ with low effort to $0.04$ with high effort. However, the effect is mixed. High effort is better than low effort in ten of the fifteen cases, and in several cases the statistic decreases. We therefore use mild effort, which costs much less than high effort. These results also suggest that the low pass rate of CoT in Section~\ref{sec:results} is not only a result of a limited reasoning budget.














\end{document}